# Linguistic and Structural Basis of Engineering Design Knowledge

L. Siddharth[1✉], Jianxi Luo[2]

[1] *Engineering Product Development, Singapore University of Technology and Design, Singapore*

[2] *Department of Systems Engineering, City University of Hong Kong, Hong Kong*

## Abstract

Natural language artefact descriptions are primary carriers of engineering design knowledge, whose retrieval, representation, and reuse are fundamental to supporting knowledge-intensive tasks in the design process. In this paper, we explicate design knowledge from patented artefact descriptions as knowledge graphs and examine these to understand the linguistic and structural basis. The purpose of our work is to advance the traditional and ontological perspectives of design knowledge and to guide Large-Language Models (LLMs) on how to articulate natural language responses that reflect knowledge that is valuable in a design environment. We populate 33,881 knowledge graphs from a sample of patents stratified according to technology classes. For linguistic basis, we conduct Zipf distribution analyses on the frequencies of unique entities and relationships to identify 64 and 37 generalisable linguistic syntaxes respectively. The relationships largely represent attributes ("of"), structure ("in", "with"), purpose ("to", "for"), hierarchy ("include"), exemplification ("such as"), and behaviour ("to", "from"). For structural basis, we draw inspiration from various studies on biological/ecological networks and discover motifs from patent knowledge graphs. We identify four 3-node and four 4-node subgraph patterns that could be converged and simplified into sequence [→ ⋯ →], aggregation [→ ⋯ ←], and hierarchy [← ⋯ →]. Based on these results, we suggest concretisation strategies for entities and relationships and explicating hierarchical structures, potentially aiding the construction and modularisation of design knowledge.

✉ Corresponding Author. siddharth_l@mymail.sutd.edu.sg, siddharthl.iitrpr.sutd@gmail.com.

# 1. Background

## 1.1. Natural Language & Design Knowledge

An engineering design process typically involves knowledge retrieval and representation enabled by various natural language documents such as consumer opinions, textbooks, technical publications, design rationale and other information (Siddharth et al., 2022). These documents comprise artefact descriptions or an aspect thereof such as usage context, requirements, functions, processes, mechanisms, embodiments, specifications, issues etc. An artefact description, as illustrated in Figure 1.1, includes explicit facts of the form *head entity :: relationship :: tail entity* which could combine into a knowledge graph, as shown in Figure 1.1.

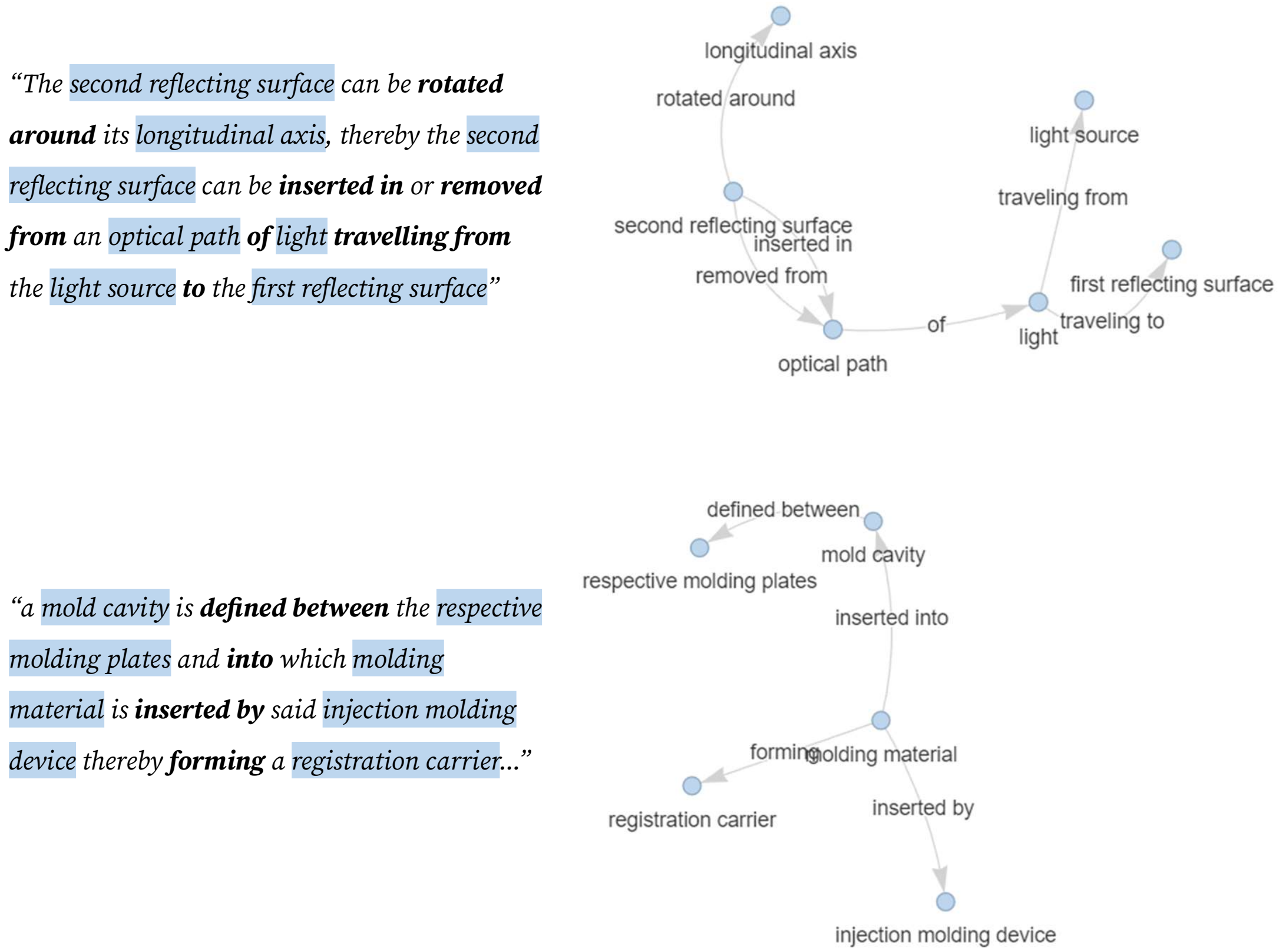

**Figure 1.1**: Illustrating explication of design knowledge from patented artefact descriptions on a) vehicle light[1] and b) injection molding[2].

The artefact descriptions are composed of domain-specific, engineering design entities like "second reflecting surface" and "optical path". The descriptions capture specific structural or behavioural relationships such as "rotated around", "inserted in", and "removed from". These relationships are often also compounded by multiple words that need not be placed adjacent to each other, e.g., "travelling from" and "travelling to". The relationships could be placed more intricately as shown in Figure 1.1b, wherein, "inserted into" is identified from "into" and "inserted" that are placed tokens apart due to the sentence structure.

---

[1] Vehicle light with movable reflector portion and shutter portion for selectively switching an illuminated area of light incident on a predetermined portion of the vehicle light during driving - https://patents.google.com/patent/US6796696/

[2] Installation for manufacturing registration carriers - https://patents.google.com/patent/US5451155/

### 1.2. Perspectives of Design Knowledge

While engineering design knowledge could explicitly be identified as shown in Figure 1.1, traditional views of design knowledge offer different perspectives as explained below.

*Function Structure Perspective*. In a design environment, Rodenacker (1976) suggests simplifying a function (overall requirement) into sub-functions to form a function structure, which could guide the retrieval of ready-made solutions from design catalogues. Stemming from this approach and improvising on Pahl and Beitz (1996, pp. 31–33), the functional models (Stone and Wood, 2000, pp. 359, 360) systematically connect various sub-functions (verb-object) using flows (nouns). Flows primarily represent material, energy, or signal transfer, which could be further categorised at lower levels of abstraction, e.g., material → "human", "gas", "liquid", and "solid" (2000, p. 361). Likewise, functions are also formalised into classes, e.g., "channel" → "import", "export", "transfer", and "guide" (2000, p. 362).

*Qualitative Physics Perspective*. De Kleer and Brown (1984) suggest that the representation of an artefact (system) should include "what it is?" (structure), "what it does?" (behaviour), and "what is it for?" (function). Building upon this premise, Umeda et al. (1996, p. 276) consider behaviour as a sequential change of states from which functions could be abstracted and which could be explained by constraints and physical laws derived from the structure. Chandrasekaran and Josephson (2000, pp. 163, 164) posit that objects are represented in a view and could have variables that change due to causal interactions, explained by causal relations. In addition, behaviour includes (2000, p. 169) the values of state variables measured at an instance or over a period and/or the relationship between these. Considering functions as effects of behaviours, Chandrasekaran (2005, pp. 68, 69) differentiates environment-centric and device-centric views.

Integrating qualitative physics and function structure perspectives, Yildirim et al. (2017, p. 425) propose a state-flow diagram wherein, states are connected by functions (Stone and Wood, 2000) and could be represented at multiple levels of abstraction. Fenves et al. (2008, pp. 2–4) propose a scheme for structuring design knowledge, where abstract classes include entity, object, relationship, and property. Among these, object classes primarily include artefacts and features, followed by function, form (geometry and material), behaviour, requirements etc. Relationship classes include constraints (like a shared property), entity association ("part of"), usage, and trace ("is based on", "version of").

*Systems Engineering Perspective*. Simon (1962, p. 468) suggests that complex systems constitute various parts that interact in different ways such that the whole is more than the sum of parts. Dori and Sillitto (2017, p. 211) posit that systems, conceptual or real, could be explained at multiple levels of abstraction, have constituents that have relationships among these and exhibit an emergence from these relationships, which involve material, energy, and information exchange that may be targeted at a specific goal. Design Structure Matrices have been popular mediums for capturing these relationships within a system that could be a product or a process (Siddharth and Sarkar, 2018, 2017).

Among the above-stated perspectives,

- Function structures are considered to oversimplify design knowledge into a set of abstract functions or flows. Umeda et al. (1990, p. 179) point out that functions need not always represent a transformation between inputs and outputs. Yildirim et al. (2017, pp. 414, 415) note that verb-noun representations could lead to imprecise articulation of functions that may not accommodate measurable attributes. In artefact descriptions, an abstract function like "transfer" could be expressed as "moved from... via... to" making it difficult to interpret functions as defined in literature.
- The schemes in qualitative physics were introduced with simple artefacts (e.g., electric buzzer, glue gun) in which, Umeda et al. (1990, p. 179) posit that the system hierarchy is not captured. In addition, within complex systems, the function of one sub-system could be the behaviour of another, causing ambiguity in the definitions of functions and behaviours (Siddharth et al., 2018). In artefact descriptions, the relationships could become increasingly complex and it is difficult to categorise these into functions, behaviours, and structures.

- The systems perspective requires design knowledge to be bounded by the system - product, process, organisation etc., where entities could be hierarchically organised and the relationships among these could be either categorical (e.g., spatial, energy), quantitative (e.g., +2, 0), or both. Such strict boundedness may not be observed in artefact descriptions, where several aspects are mentioned in different levels of abstraction.

Although existing perspectives may provide guidelines for newly representing a system, the constructs of design knowledge within these perspectives do not reflect the explicit design knowledge that is most commonly populated though artefact descriptions. The primary motivation of our work is therefore to advance design theories in terms of the constructs of design knowledge embodied in natural language artefact descriptions.

### 1.3. LLMs & Design Knowledge

LLMs like GPT, Llama, Falcon etc., are being entrusted for the retrieval and generation of knowledge in various fields and are becoming popular in engineering design research, education, and practice. Since LLMs abstract language from a variety of sources, the content and characteristics of the text generated are yet to reflect design knowledge that is communicated in typical artefact descriptions. As follows, we compare the entities and **relationships** in a patent abstract[3] and a GPT-generated technical description (of similar length) of a coordinate measuring device.

*[Patent Abstract] Coordinate measuring device. A method* ***of inspecting*** *a series* ***of*** *workpieces* ***using*** *a coordinate measuring apparatus,* ***comprising*** *the steps of:* ***measuring*** *a calibrated artefact* ***on*** *a coordinate measuring apparatus* ***at*** *a fast speed 28;* ***generating*** *an error map* ***corresponding*** *to the difference* ***between*** *the calibrated artefact and the measured artefact 30;* ***measuring*** *subsequent workpieces* ***at*** *the same fast speed 34 and* ***correcting*** *the measurements* ***of*** *the subsequent workpieces* ***using*** *the error map 36. The artefact may* ***be*** *one* ***of*** *the workpieces.*

*[GPT-generated description] A coordinate measuring device (CMM)* ***is*** *a precision instrument* ***used to measure*** *the physical geometrical characteristics of an object. It* ***utilizes*** *a probe* ***to determine*** *the coordinates* ***of*** *points* ***on*** *the object's surface,* ***translating*** *them* ***into*** *digital data. CMMs can* ***be*** *manual or automated,* ***employing*** *various probe types* ***such as*** *touch-trigger, optical, or laser. They are* ***essential in*** *quality control and reverse engineering,* ***ensuring*** *accurate dimensional analysis and conformance* ***to*** *specifications* ***in*** *manufacturing processes.*

The patent abstract describes a method and its underlying steps, with concrete entities (e.g., “error map 36”) and specific relationships (e.g., “of inspecting”). The entities are associated intricately by the relationships whose tokens are placed far apart (e.g., “of correcting”) in the description, as depicted in the knowledge graph in Figure 1.2. The GPT-generated description, on the other hand, only gives an abstract view of the device and states its common applications, which is less valuable for a design environment. Since tools that serve knowledge-based applications are expected to built upon LLMs, we recognise the need to inform such tools on the content and characteristics of design knowledge embodied in natural language artefact descriptions.

[3] Coordinate Measuring Machine - https://patents.google.com/patent/US9797706B2/

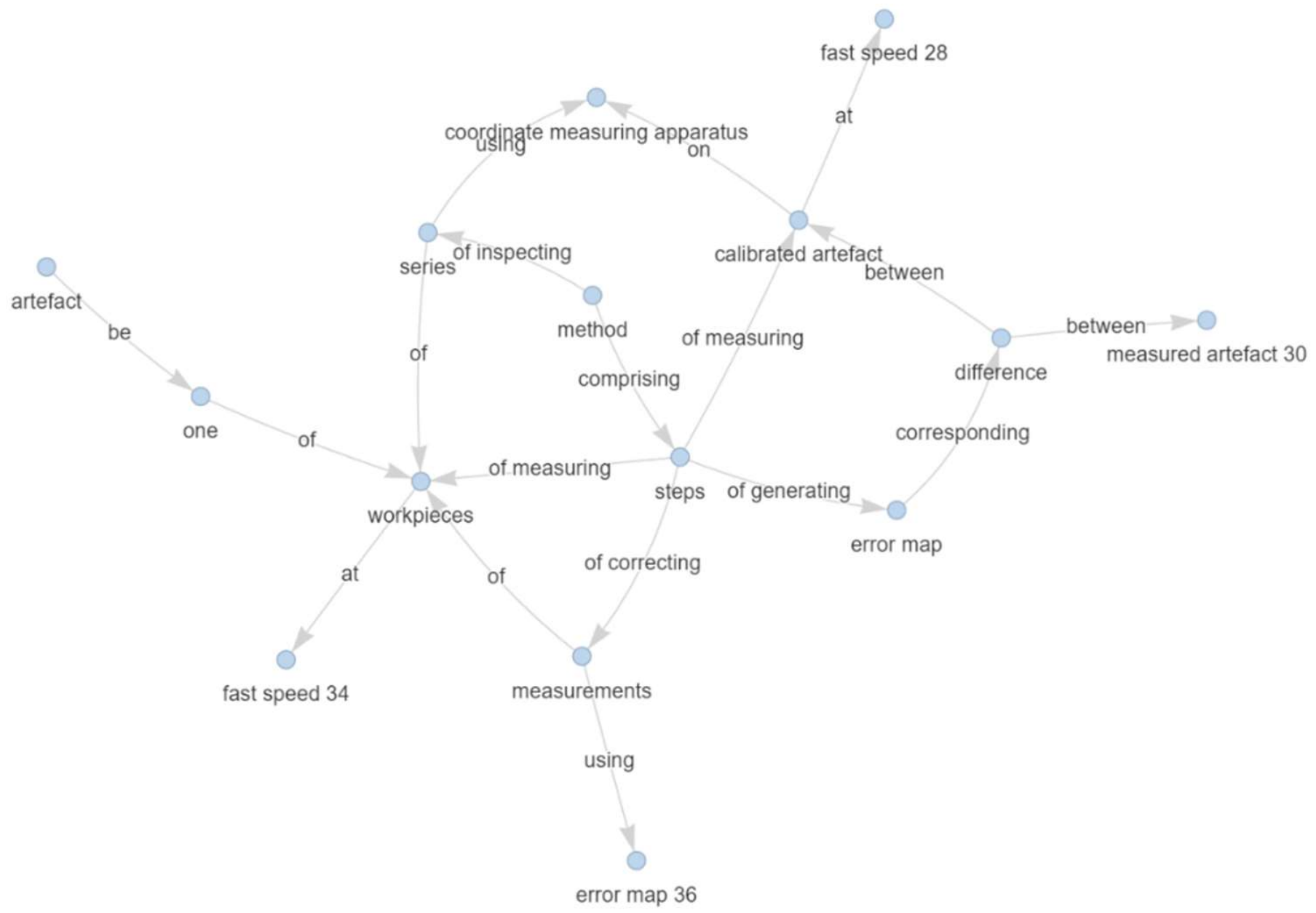

**Figure 1.2**: Knowledge graph extracted from the patent abstract.

## 1.4. Research Approach

The perspectives of design knowledge in literature capture only a part of actual design knowledge that is communicated in artefact descriptions. Resorting to LLMs for knowledge retrieval leads to responses that do not reflect design knowledge as communicated in typical artefact descriptions. These broad issues raise fundamental questions regarding the constructs and characteristics of design knowledge as follows.

*What is the basis of engineering design knowledge in artefact descriptions?*

- *What is the linguistic basis of engineering design knowledge in artefact descriptions?*
- *What is the structural basis of engineering design knowledge in artefact descriptions?*

The objective of our work, as reported in this paper, is to provide empirical insights based on explicit design knowledge, as identified from artefact descriptions. To this end, our approach is to explicate design knowledge from a large sample of artefact descriptions as knowledge graphs (as illustrated in Figure 1.3). The method to explicate design knowledge is detailed in our prior work (Siddharth and Luo, 2024a) and is available as a public resource[4]. We examine the knowledge graphs populated using the method to understand the basis. Since knowledge graphs express both linguistic (entities and relationships) and structural (network structure) aspects, we study linguistic and structural basis separately as indicated in the questions above.

Our research is expected to make the following contributions to the engineering design and knowledge engineering literature. First, we populate a large dataset of knowledge graphs from patented artefact descriptions (as illustrated in Figure 1.3) that could serve various applications in engineering design research and education. Second, we propose an innovative approach to examine these knowledge graphs, drawing inspiration from Computational Linguistics, Biology/Ecology, and Network Sciences. Third, the lingustic basis derived using our analysis could be an imperative support system for LLMs, informing how entities and relationships are constructed in artefact descriptions. Fourth, the structural basis dervied using our analysis could instigate fundamental theories in engineering design synthesis, explaining how entities and relationships are combined and organised within artefacts. Fifth, based on the learnings from our analysis, we propose concretisation strategies that could be useful for future knowledge management systems. In addition to the Supplementary Material for this paper, we provide an external resource[5] for accessing the datasets utilised and generated during this research (Siddharth and Luo, 2024b).

---

[4] Design Knowledge Extraction - https://github.com/siddharthl93/engineering-design-knowledge/tree/main

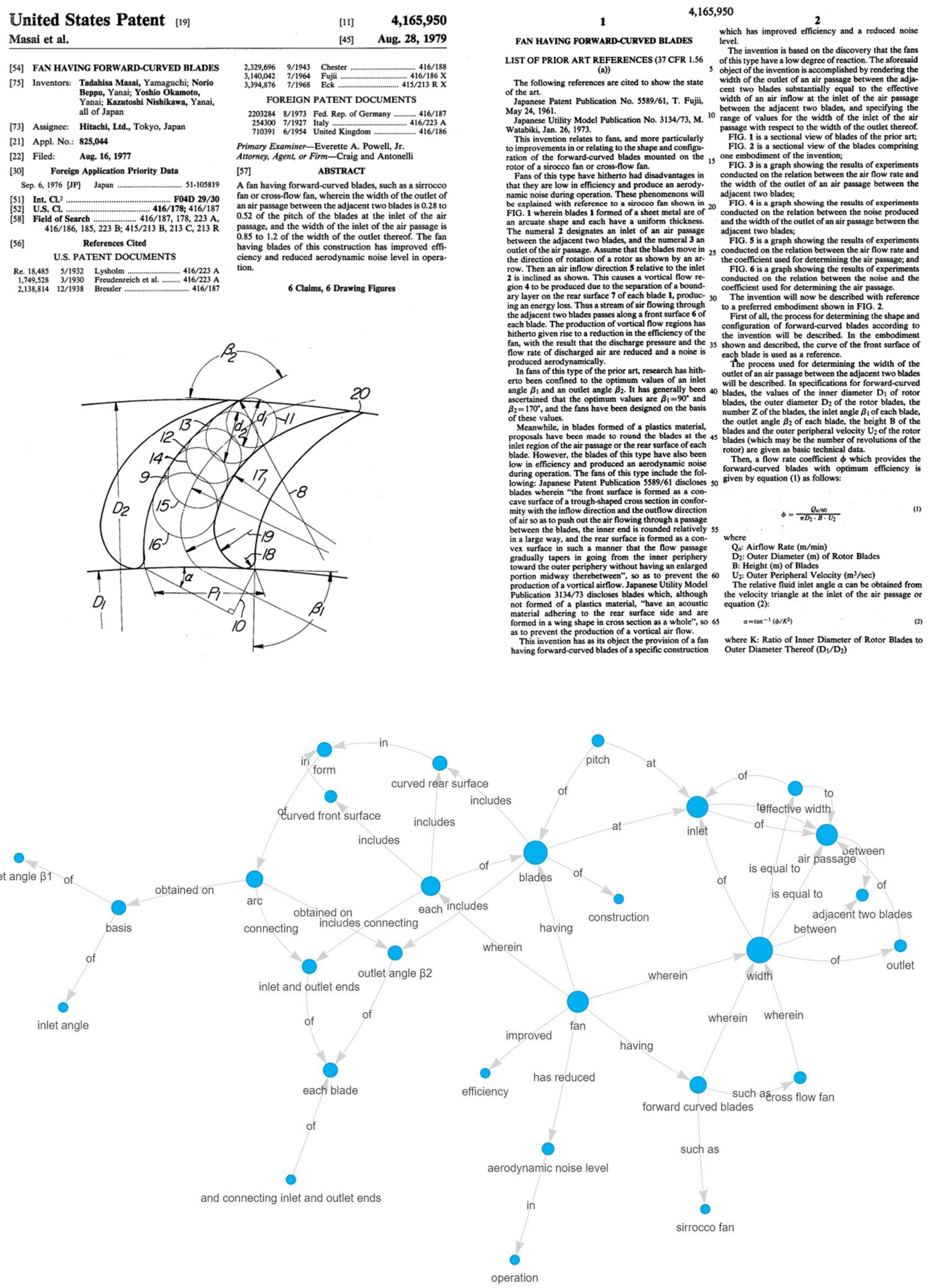

**Figure 1.3**: Example knowledge graph extracted from a patent related to fan systems[6] using a method developed in our prior work (Siddharth and Luo, 2024a).

[5] Datasets utilised and generated during this research - https://zenodo.org/records/13328258

[6] Fan having forward-curved blades - https://patents.google.com/patent/US4165950A/

## 2. Data and Methods

### 2.1. Knowledge Graph Data

We gather a large sample of patented artefact descriptions from USPTO, which granted over 8.2 million patents as of June 8, 2023[7]. As shown in Table 2.1, we filter utility patents from these and use Cochran's formula to sample patents. Accounting for a 99% confidence level and 1% margin of error, we obtain a sample of 33,945 patents that are stratified according to the CPC 4-digit classification scheme[8] (e.g., A02F). We scrape full-text from the sample of patents and then apply the method developed in our prior work (Siddharth and Luo, 2024a) to extract knowledge graphs. The patent data and the knowledge graphs are accessible in the external resource (Siddharth and Luo, 2024b).

**Table 2.1**: Sampling patents and extracting design knowledge from USPTO.

| **Sampling Patented Artefact Descriptions** | |
|---|---|
| Total Number of Granted Patents | 8,260,142 |
| - Upon filtering WIPO kind to A, B1, B2 | 7,484,623 |
| - Upon selecting utility patents | 7,484,622 |
| - Upon filtering minimum of one claim | 7,373,519 |
| - Upon sampling | 33,945 |
| - Upon scraping full text | 33,884 |
| **Extracting Design Knowledge** | |
| Number of Patents with Facts | 33,881 |
| Number of Sentences | 7,191,733 |
| Number of Sentences with Facts | 5,451,709 |
| Number of Facts | 24,537,587 |
| Number of Unique Entities | 5,015,681 |
| Number of Unique Relationships | 845,303 |

### 2.2. Linguistic Basis – Zipf Distribution Analysis

For the linguistic basis of design knowledge, we study the frequencies of 5,015,681 unique entities and 845,303 unique relationships that constitute the knowledge graphs. Term frequencies in a corpus are often studied using the Zipfian distribution (Mollica and Piantadosi, 2019, p. 6), which is also adopted by Murphy et al. (2014) in their analysis of 65,000 patent abstracts to identify the most frequent 1,700 functions (verbs). Zipf distribution follows the probability mass function (PMF) as shown in Eq. 1.

$$PMF(k; s, N) = \frac{1}{H_{N,s}} \cdot \frac{1}{k^s} \quad (1)$$

$$H_{N,s} = \sum_{k=1}^{N} \frac{1}{k^s} \quad (2)$$

In the above equations, $k$ stands for the rank of a term in the frequency distribution and $s$ is the distribution parameter. The idea behind the Zipf distribution is that the probability of a term in the corpus is inversely related to its rank – $k$ raised to the power – $s$. To ensure the probabilities of all terms sum to 1, a harmonic number ($H$) as in Eq. 2 is multiplied in the probability mass function. Based on the frequency proportions of entities and relationships, we perform a curve-fitting using SciPy[9] on Eq. 1. The fitted probability mass function could be plotted as a straight line with slope $s$.

[7] US Patent Database access facilitated by Patents View - https://patentsview.org/download/data-download-tables

[8] Cooperative Patent Classification Scheme - https://www.uspto.gov/web/patents/classification/cpc/html/cpc.html

[9] SciPy Curve Fit - https://docs.scipy.org/doc/scipy/reference/generated/scipy.optimize.curve_fit.html

In this analysis, we show the cumulative distribution plots of the fitted distribution to visualise terms at different percentiles. As the terms in entities and relationships assume different meanings in varied contexts, we transform these terms into linguistic syntaxes as follows. We convert each word in a term into its parts of speech except for the word that includes pronouns, prepositions and 30 most frequent words like 'device', 'system', 'said', 'portion', 'information', 'signal', 'invention', 'layer', 'method', 'user', 'material' etc. Some example syntaxes thus derived include "a shake" → "a NN", "the cured spar assembly" → "the JJ NNP NNP", "all three erase blocks" → "all CD NN NNS", "the main antenna signal" → "the JJ NN signal". In this analysis, we use the frequencies of these syntaxes instead of actual terms.

### 2.3. Structural Basis – Motif Analysis

For structural basis, we identify 3-node and 4-node subgraph patterns that form the building blocks of the network structures of the knowledge graphs. Such patterns are termed motifs and are identified using the statistical significance of their counts in a given network, in comparison with a randomised version of the network. To discover motifs from network structures of the knowledge graphs, we incorporate methods from earlier studies on sensory transcription networks, ecological food webs etc., (Alon, 2006; Milo et al., 2002; Strona et al., 2014). We also customise these methods to suit the sparse knowledge graphs, e.g., the patent[10] with the largest graph, having 4460 nodes and 8204 edges, indicating a mean degree = 1.839. The primary steps underlying this process include subgraph mining, pattern matching, network randomising, and analysing differences.

*Subgraph Mining*. Considered NP-hard, the subgraph mining problem is usually addressed with specialised algorithms, specific to a certain pattern (Ribeiro et al., 2021, p. 5). Since the knowledge graphs in our data are quite sparse, e.g., Supplementary Material, Figure S-1, we mine subgraphs by combining edges. For instance, a graph includes nodes 0, 1, 2… 8 and edges (0, 6), (0, 7), (1, 4), etc. We mine triples (3-node subgraphs) by combining edges with three unique nodes, e.g., ((0, 6), (0, 7)) → {0, 6, 7}. We combine triples and edges to get quadruples (4-node subgraphs), e.g., ((0, 6, 7), (0, 3)) → {0, 3, 6, 7}. This strategy is much simpler and more effective compared to several sophisticated methods in literature.

**Table 2.2**: Obtaining unique sequences for 3-node and 4-node patterns to match with isomorphs.

| | 3-Node Patterns | |
|---|---|---|
| | 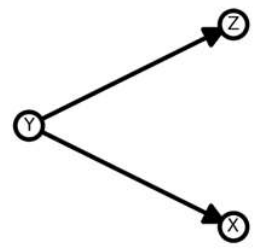 | 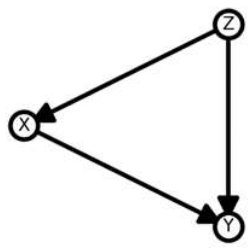 |
| In Degree (X) | 1 | 1 |
| In Degree (Y) | 0 | 2 |
| In Degree (Z) | 1 | 0 |
| Out Degree (X) | 0 | 1 |
| Out Degree (Y) | 2 | 0 |
| Out Degree (Z) | 0 | 2 |
| Edges (XY) | 1 | 1 |
| Edges (YZ) | 1 | 1 |
| Edges (XZ) | 0 | 1 |
| Unique Sequence | 011-002-011 | 012-012-111 |

| | 4-Node Patterns | |
|---|---|---|
| | 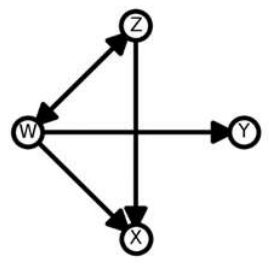 | 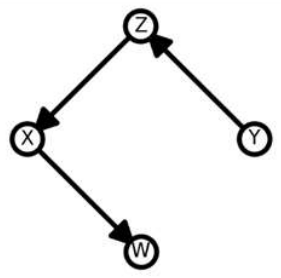 |
| In Degree (W) | 1 | 1 |
| In Degree (X) | 2 | 1 |
| In Degree (Y) | 1 | 0 |
| In Degree (Z) | 1 | 1 |
| Out Degree (W) | 3 | 0 |
| Out Degree (X) | 0 | 1 |
| Out Degree (Y) | 0 | 1 |
| Out Degree (Z) | 2 | 1 |
| Edges (WX) | 1 | 1 |
| Edges (WY) | 1 | 0 |
| Edges (WZ) | 2 | 0 |
| Edges (XY) | 0 | 0 |
| Edges (XZ) | 1 | 1 |
| Edges (YZ) | 0 | 1 |
| Unique Sequence | 1112-0023-001112 | 0111-0111-000111 |

[10] Method for treating an individual suffering from a chronic infectious disease and cancer - https://patents.google.com/patent/US11213552B2/

*Pattern Matching*. Each subgraph belongs to one of the various patterns visualised in Supplementary Material, Figure S-2. As subgraphs cannot be directly matched with the pattern graphs, it is necessary to develop a canonical form that is common for all subgraphs belonging to a pattern (Chomsky, 2014). In Table 2.2, we derive such canonical forms as unique sequences for distinct patterns by incorporating features like in-degrees, out-degrees, and number of edges between individual pairs. In the features for each type, we sort the values in ascending order so that the unique sequence is common for all subgraphs (isomorphs) for each pattern. For instance, in-degrees in the second pattern of Table 2.2 are X = 1, Y = 2, and Z = 0, which is sorted as "012-###-####". Upon following this approach, we find exactly 13 unique sequences for 3-node subgraphs, consistent with 13 patterns in the literature (Milo et al., 2002, p. 824). In total, we identify 13 patterns and 134 patterns from all possible 3-node and 4-node graphs as depicted in Supplementary Material, Figure S-2.

*Randomising Graphs*. Upon mining subgraphs and matching patterns, it is possible to get the counts of all patterns in a given graph. To measure the statistical significance of a pattern count, it is necessary to compare that against the count observed in a randomised version of the original graph. Alon (2006, p. 27) generates an ensemble of Erdos-Renyi random graphs of node count similar to that of the original graph. Watts and Strogatz (1998, pp. 440, 441) propose rewiring edges in these graphs with a probability that introduces significant amounts of disorder and randomness, measured by a decrease in average shortest path length and clustering coefficient.

In studies involving ecological networks like species-habitat, species-prey relationships etc., node degrees represent the relative differences among species, which is retained while randomising such graphs. Stone et al. (2019, p. 4) propose swapping diagonally opposite 0's and 1's in the adjacency matrices of such graphs. Strona et al. (2014, pp. 3, 4) demonstrate with a 100 x 100 matrix that over 50,000 swaps are necessary to achieve 50% perturbation, which is a general rule of thumb in randomisation. They alternatively propose a "curveball" algorithm (as illustrated in Figure 2.1) that reaches 50% perturbation after 200 iterations.

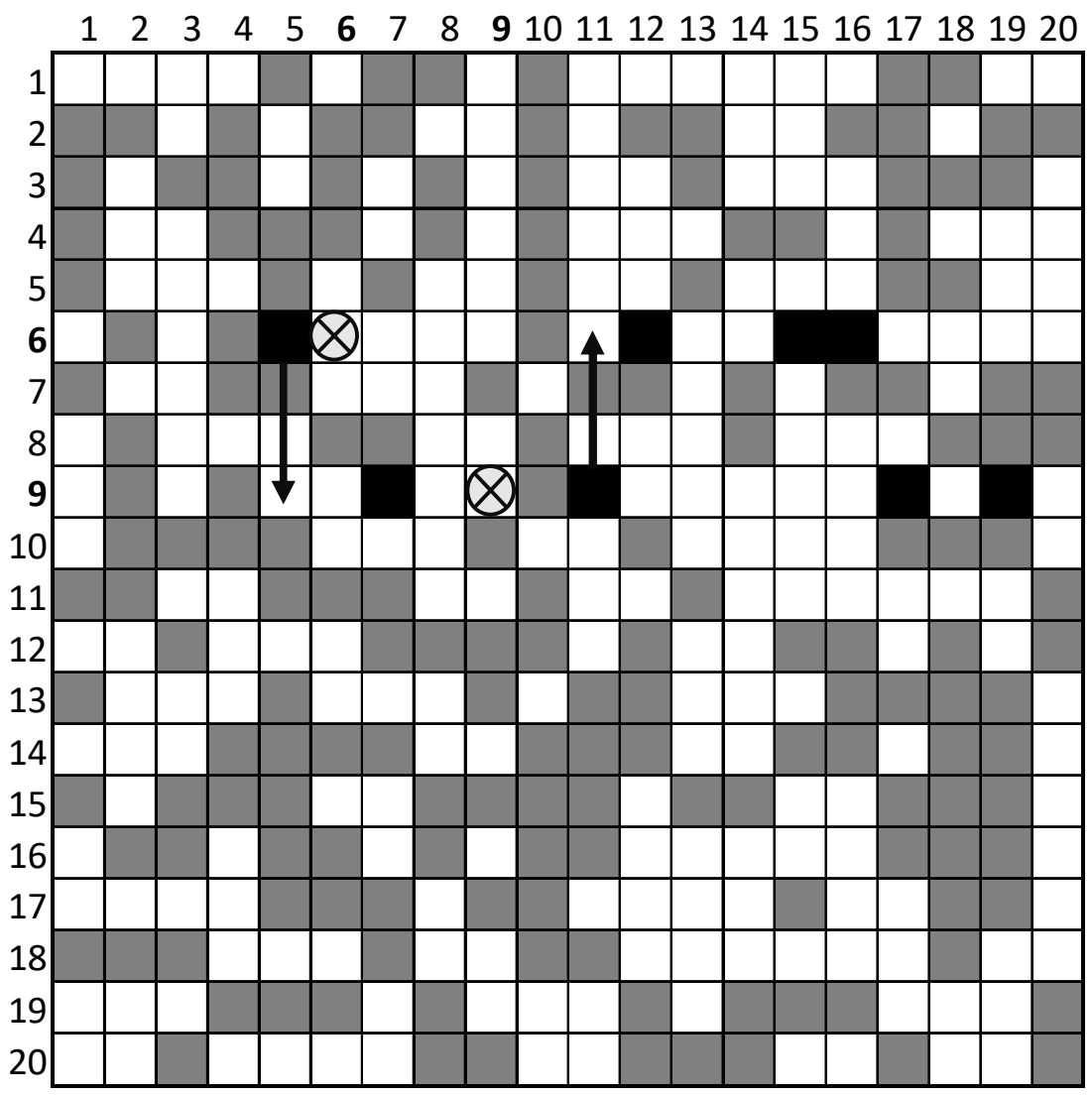

**Figure 2.1**: Illustrating the curveball algorithm.

Given a graph comprising 20 nodes and represented using a matrix as in Figure 2.1, the curveball algorithm requires choosing a random pair of nodes, e.g., 6 and 9 in Figure 2.1. Among the non-mutual connections of the chosen pair, two nodes have to be selected at random and swapped. While swapping, as shown in Figure 2.1, we pose an additional constraint that, for any node, connection to itself is prevented. While a node relating to itself is existent in other graphs, e.g., termed “auto-regulation” in sensory transcription networks (Alon, 2006, p. 37), knowledge graphs populated in our data do not include such instances. We apply the curveball method to generate a random graph for each knowledge graph in our data that includes 33,881 knowledge graphs.

*Analysing Differences*. By performing the steps thus far, subgraphs can be mined and assigned to patterns for both original and randomised graphs. Given $N$ patents in the sample, a subgraph pattern $p$ could be present $X_{ip}$ times in the $i^{th}$ patent and $Y_{ip}$ in the randomised version of the patent, where $i \in [1, N]$. We intend to measure the statistical significance of the difference $X_{ip} - Y_{ip}$ with respect to all remaining patents in the sample. Since the patents differ in terms of the sizes of knowledge graphs, we obtain the normalised difference $\Delta_{ip}$ as follows, by dividing by the number of edges $E_i$.

$$\Delta_{ip} = \frac{X_{ip} - Y_{ip}}{E_i} \tag{3}$$

Assuming the $\Delta_{ip}$ is normal, we calculate the Z-score as follows, to assess how large it is based on the mean ($\mu_p$) and standard deviation ($\sigma_p$) measured across patents in the sample.

$$Z_{ip} = \frac{\Delta_{ip} - \mu_p}{\sigma_p} \tag{4}$$

Consider $M_{ip}$ a binary variable that tells whether the pattern $p$ is a motif in patent $i$ with 95% confidence level,

$$M_{ip} = \begin{cases} 0 & Z_{ip} < 1.64 \\ 1 & Z_{ip} \geq 1.64 \end{cases} \tag{5}$$

Using the approach described above we compute $M_{ip}$ for all patterns in each patent. The pattern $p$ becomes generalisable if it is a common motif across all patents in the entire sample or within the domain. To assess this, we calculate the sum $\sum_i^N M_{ip}$ across the set of patents and observe how large this value is, compared to that of other patterns. Here, we again apply a Z-score strategy similar to Eq. 3 but with a 99% confidence level, i.e., Z-score greater than or equal to 2.32. In this paper, we only provide an overall discussion of the generalisable motifs.

## 3. Results

### 3.1. Linguistic Basis – Entity Syntaxes

Upon converting each entity into its corresponding syntax as explained in Section 2.2, we reduce the 5,015,681 unique entities into 408,323 unique entity syntaxes. When the entity syntaxes are fitted to the Zipf distribution (Eq. 1) as per their frequency proportions, we estimate the parameter $s$ = 1.093. In Figure 3.1, we plot the original frequency proportions, fitted distribution, and cumulative distribution. In Table 3.1, we showcase the top 30 syntaxes at different percentiles. In Supplementary Material, Figure S-3 and Table S-1, we provide the Zipf distribution of actual entities.

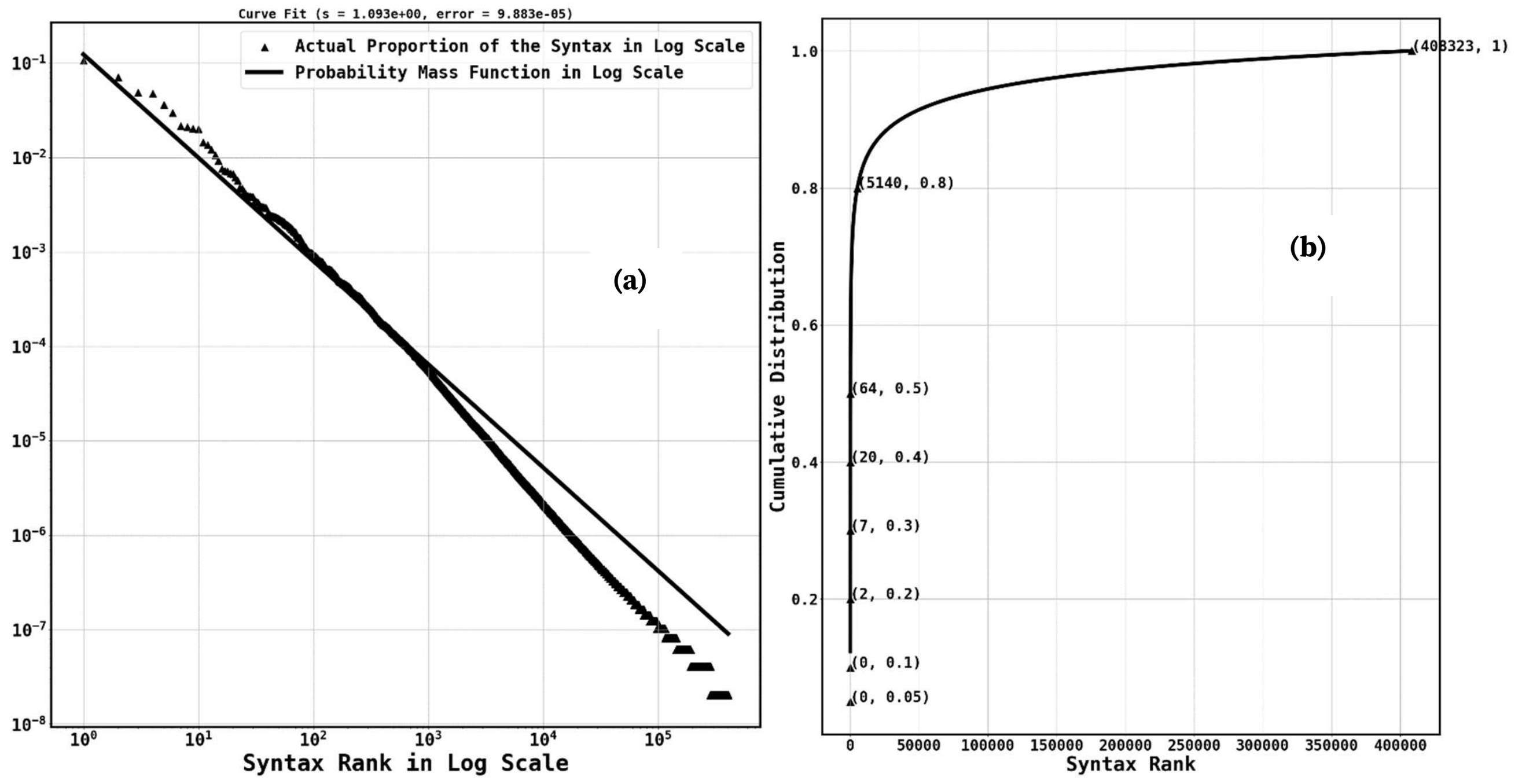

**Figure 3.1**: Zipf Distribution using the frequency of unique entity syntaxes: a) probability mass function in log scale against the actual proportion of entity syntaxes, b) cumulative distribution indicating the count at various percentiles.

**Table 3.1**: Overview of (generalisable) entity syntaxes at different percentiles as per cumulative Zipf distribution.

| **Entity Syntaxes<br>Zipf Distribution: s = 1.093, error = 9.883e-05** | | |
|---|---|---|
| **CDF** | **Count** | **Syntaxes (Top 30)** |
| **(0, 0.05]** | 0 | |
| **(0.05, 0.1]** | 0 | |
| **(0.1, 0.2]** | 2 | *the NN, NN* |
| **(0.2, 0.3]** | 5 | *NNS, a NN, the NN NN, the JJ NN, a JJ NN* |
| **(0.3, 0.4]** | 13 | *JJ NNS, the NNS, a NN NN, JJ NN, NN NNS, NN NN, an NN, VB, NNP, the JJ NN NN, an JJ NN, the JJ NNS, the NN NNS* |
| **(0.4, 0.5]** | 44 | *a JJ NN NN, the NN NN NN, VBZ, JJ NN NNS, the VBG NN, the present invention, a plurality, each, an NN NN, VBG, a NN NN NN, other NNS, NNP NN, the method, the invention, this NN, the VBN NN, the first NN, the NNP NN, VBG NNS, JJ NN NN, VBN NNS, data, NNP NNP, each NN, a VBN NN, the second NN, JJ, one or more NNS, the user* |
| **(0.5, 0.8]** | 5,076 | *an JJ NN NN, a first NN, the JJ NN NNS, these NNS, VBN NN, NN NN NN, VBG NN, the JJ, the plurality, another NN, the VBG NNS, the NN device, the NN system, the JJ surface, the number, such NNS, the VBN NNS, the first NN NN, a user, the JJ device, the surface, the second NN NN, the device, VB NN, a portion, the JJ NN NN NN, the VBN NN NN, any, a RB JJ NN, a number* |
| **(0.8, 1]** | 403,183 | *the VBN NNS NNS, all VBP, JJ or other NNS, an JJ JJ position, like NN, JJR layer VBG, JJ and NN NN NNS, the NNP user NN, any other JJ JJ NN, VBZ JJ NN, an VBG NN NN NN, the first and second end NNS, an JJ NNP, RB based NNS, one user, the first JJ image, an JJ VBG surface, RB , NN NNS, the second VBG unit, VBN NN JJ NN, each NN NN NN NN, LS -RRB- a NN NN, VB JJ information, RB RB VBN NNS, NN NNS data, FW NN NN, a JJ number NN, an JJ VBG NN NN, RB JJ JJ NN RB NN, NN VBN data* |

According to Table 3.1, over 20 percent of the entities are composed of single nouns (NN, the NN) consistent with the common notion in design literature that entities are composed of nouns (Siddharth and Chakrabarti, 2018; Song and Fu, 2019). Until 40 percent, nouns are combined with other nouns (plural and proper) and adjectives. Until 50 percent, combinations with verbs (e.g., VBG NNS), and count words (e.g., the first NN) are found. Hagoort (2019) suggests that complex terms that lie at the lower side of the distribution, e.g., "the front facing surface" are formed by hierarchically associating information like ["facing" ["front" ["surface"]]] to the abstract, root word "surface". As the syntaxes beyond 50 percent are built as a combination of abstract syntaxes, we concur that the 64 syntaxes constituting half of the distribution are generalisable (as highlighted in Table 3.1) and thus represent the linguistic basis of design knowledge.

### 3.2. Linguistic Basis - Relationship Syntaxes

We identify 73,352 unique syntaxes from 845,303 unique relationships. Upon performing the Zipf distribution, as indicated in Figure 3.2, we overview the syntaxes at different percentiles, as shown in Table 3.2.

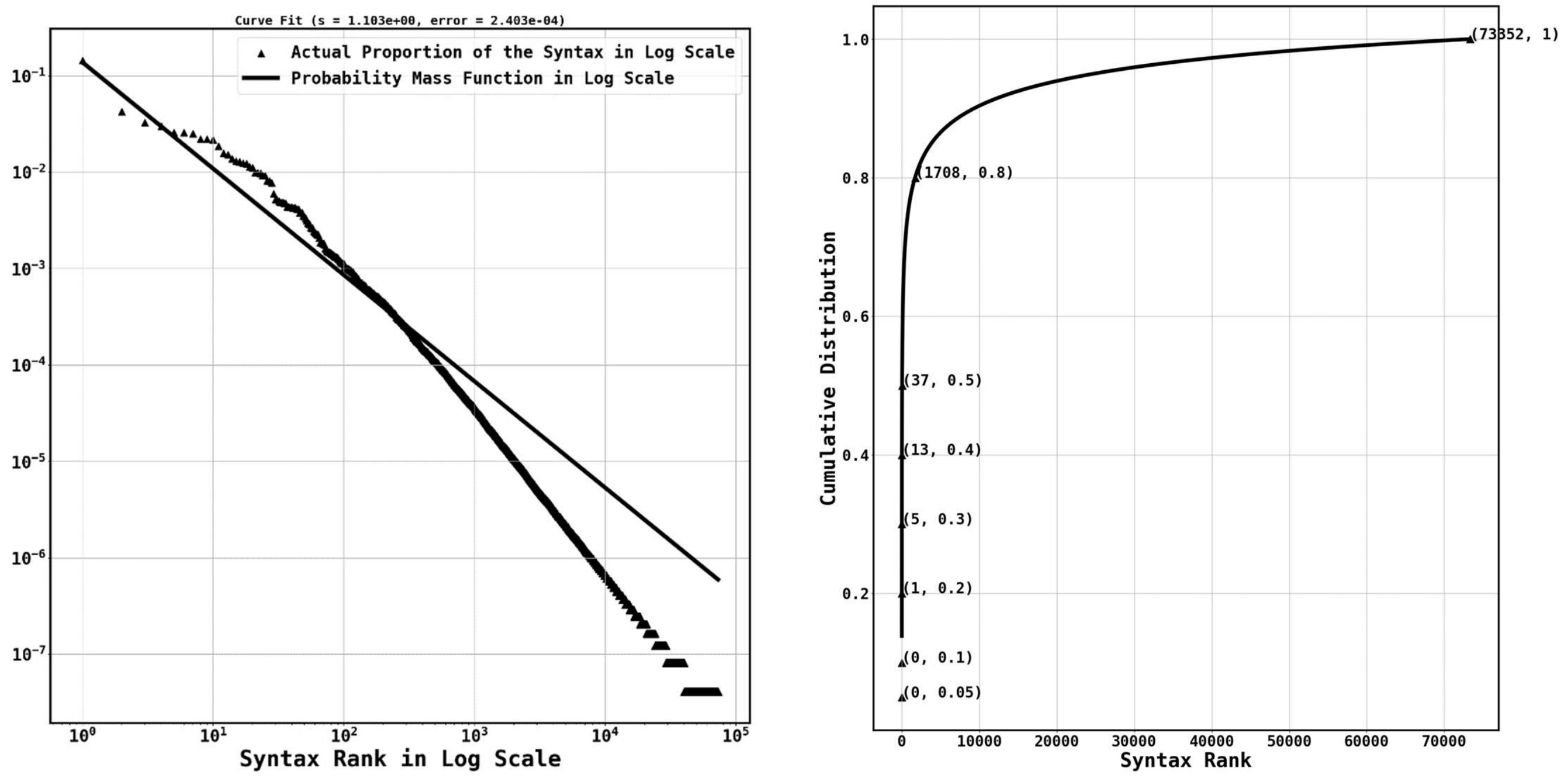

**Figure 3.2**: Zipf Distribution using frequency of unique relation syntaxes: a) probability mass function in log scale against the actual proportion of relation syntaxes, b) cumulative distribution indicating the count at various percentiles.

**Table 3.2**: Overview of (generalisable) relationship syntaxes at different percentiles as per cumulative distribution.

| **Relationship Syntaxes<br>Zipf Distribution: s = 1.103, error = 2.403e-04** | | |
|---|---|---|
| **CDF** | **Count** | **Syntaxes (Top 30)** |
| **(0, 0.05]** | 0 | |
| **(0.05, 0.1]** | 0 | |
| **(0.1, 0.2]** | 1 | *of* |
| **(0.2, 0.3]** | 4 | *in, to, include, to VB* |
| **(0.3, 0.4]** | 8 | *such as, VBZ, VB, for, with, VBG, includes, from* |
| **(0.4, 0.5]** | 24 | *VBN by, including, wherein, is, NNS, between, for VBG, on, by, having, VBN in, at, comprises, comprising, has, as, VBN to, VBN from, have, is VBN to, is VBN by, through, into, using* |
| **(0.5, 0.8]** | 1,671 | *are, is VBN in, NN, based on, by VBG, VBZ to, comprise, VBD, JJ to, of VBG, according to, within, consisting of, be, VBG to, associated with, VBD in, VBN with, are VBN in, used to VB, VBD to, configured to VB, VB to, to provide, VBN, is VBN from, provide, VBN on, are VBN by, over* |
| **(0.8, 1]** | 71,643 | *VB for VBG, are RB JJ as, are NN to, VB below, present between, used IN, RB are, is configured to have, RB VBP on, NNS based on, comprises from, are formed into, RB onto, VBG RB between, is NN in, are VBN towards, to VB based on, RBR in, NNS according to, are RB VBN via, from VBG with, JJ during, VBZ to RB VB, provided using, position, VBP RB in, VBD up to, are VBN VBN by, VBZ RB VBG, are RB VBN by VBG* |

In Supplementary Material, Figure S-4 and Table S-2, we provide the Zipf distribution of actual relationships. Over 20 percent of the relationships are given by "of" that is quite frequently expressed as "*attribute :: of :: entity*" facts in artefact descriptions. Among other frequent syntaxes, "in" and "to" capture various relationships depending on the context. For instance, "to" can be mentioned in the transition sense, e.g., "changed from... to...", in an introductory sense, e.g., "related to... and to...", or utility sense, e.g., "to deliver..." as expressed by "to VB syntax". The other common relationship syntax captures hierarchy using "include" which is quite an imperative relationship for an artefact description.

Among the syntaxes in the 30-40 percent range as shown in Table 3.2, exemplar relationship ("such as") precedes single verbs ("VBZ", "VB"). It is a common notion in literature that verbs are representative of relationships (Jamrozik and Gentner, 2020). Upon analysing 110 functional models of mechanical artefacts such as can-openers, nutcrackers, and toasters, Lucero et al. (2014, pp. 4, 6) identify the most important functions as "transfer", "convert", "store", "actuate", "separate" etc. Siddharth et al. (2021) build upon this notion to populate facts of the form "*entity :: verb :: entity*" from patent claims. Our analysis of explicit design knowledge in artefact descriptions reveals that single verbs do not constitute even the first 30 percent of relationships. When design knowledge is naturally documented in text, explicit verbs are often used in association with other verbs and prepositions as mentioned in the 40-50 percent range as shown in Table 3.2. For example, the "transfer" function identified by Lucero et al. (2014, pp. 4, 6) could be represented as "moved from" or "moved to". Altogether, the 37 syntaxes that constitute half of the relationships are generalisable (as highlighted in Table 3.2) and thus represent the linguistic basis of engineering design knowledge.

In the 40-50 percent range, different forms of hierarchical relationships are found and these are quite imperative to provide a systemic description of an artefact. To understand how such hierarchical relationships are constructed, we gather 3,968 unique relationships (not the syntaxes) as listed in Supplmentary Material, Table S-2. Upon ranking these relationships based on the similarity as measured using Sentence BERT (Reimers and Gurevych, 2019) with the term "include", we shortlist 165 terms, which we generalise to 75 syntaxes as overviewed in Table 3.3. These 75 syntaxes are generalisable representations of hierarchical relationships and are an important part of the linguistic basis of design knowledge.

**Table 3.3**: Overview of 165 hierarchical relationships and 75 syntaxes.

| Rank | Relationship | Similarity | Count | Rank | Relationship Syntax | Count |
|---|---|---|---|---|---|---|
| 1 | **includ**e | 1 | 738,101 | 1 | **includ***| 1,447,248 |
| 4 | **wherein** | 0.390 | 317,277 | 2 | **compris*** | 498,603 |
| 5 | **compris**es | 0.521 | 199,385 | 3 | **wherein** | 317,277 |
| 8 | **consist**ing of | 0.471 | 85,800 | 4 | **contain*** | 115,404 |
| 9 | **contain**ing | 0.642 | 52,438 | 5 | **consist*** of | 99,682 |
| 15 | **compos**ed of | 0.383 | 6,886 | 7 | **includ*** VBG | 12,888 |
| 18 | not **includ**e | 0.894 | 4,578 | 8 | **constitut*** | 11,621 |
| 19 | **involv**es | 0.507 | 4,508 | 11 | **involv*** | 8,527 |
| 25 | **encompass**es | 0.459 | 3,790 | 12 | **compris*** of | 7,576 |
| 37 | **compris**ing administering | 0.368 | 2,077 | 14 | **compos*** of | 6,886 |
| 41 | typically **includ**es | 0.759 | 1,671 | 15 | **compris*** VBG | 6,712 |
| 57 | **includ**es forming | 0.623 | 1,230 | 16 | **encompass*** | 6,628 |
| 61 | **consist**s essentially of | 0.457 | 1095 | 20 | not **includ*** | 4,578 |
| 80 | preferably **contain**s | 0.484 | 793 | 39 | **constitut*** by | 1,391 |
| 85 | not **compris**e | 0.504 | 700 | 55 | is **part of** | 646 |
| 96 | is **part of** | 0.454 | 646 | 60 | configured to **includ*** | 485 |
| 125 | **includ**es transmitting | 0.499 | 501 | 61 | **includ*** with | 484 |
| 151 | **part of** | 0.513 | 392 | 66 | VBN **includ*** | 407 |
| 159 | **compris**es detecting | 0.344 | 380 | 74 | to be **includ*** in | 377 |
| 165 | **includ**es obtaining | 0.660 | 367 | 75 | **compos*** | 369 |

### 3.3. Structural Basis - Motifs

We identify motifs for each patent knowledge graph in our data using the methods explained in Section 2.3. In Supplementary Material, Tables S-3, S-4, and S-5, we list all patterns, stating their raw count and the number of patents in which these are significant. Herein, we shortlist the patterns that are the most common motifs across all patents (99% confidence level) as shown in Table 3.4. We identify such common motifs across the sample as well as within the 10 most populated classes in USPTO. The most common motif is Pattern 13 that represents an aggregation [→·←]. Despite having the highest frequency, Pattern 9 – sequence [→·→] is not an overall common motif, while its combination with Pattern 13 given by Pattern 9 is the second most common motif. Among 4-node patterns, the most common ones are Patterns 130 and 141 that are formed as aggregation-sequence and aggregation-aggregation combinations respectively.

**Table 3.4**: Dominant motifs overall and within the largest classes.

| Sub Class | Patent Count | Definition |
|---|---|---|
| **Overall** | 33881 | - |
| **Y10T** | 4087 | Technical subjects covered by former US classification. |
| **G06F** | 3452 | Electric digital data processing. |
| **Y10S** | 2690 | Technical subjects covered by former USPC cross-reference art collections and digests. |
| **H04L** | 2508 | Transmission of digital information, e.g., Telegraphic communication. |
| **H01L** | 2313 | Semiconductor devices not covered by class H10. |
| **A61K** | 1739 | Preparations for medical, dental or toiletry purposes. |
| **H04N** | 1560 | Pictorial communication, e.g., Television. |
| **Y02E** | 1519 | Reduction of greenhouse gas emissions, related to energy generation, transmission or distribution. |
| **H04W** | 1460 | Wireless communication networks. |
| **A61P** | 1399 | Specific therapeutic activity of chemical compounds or medicinal preparations |

| | | **(Patent Count, Raw Count)** | | |
|---|---|---|---|---|
| **Pattern #** | **Pattern 8** | **Pattern 9** | **Pattern 11** | **Pattern 13** |
| **Motif** | 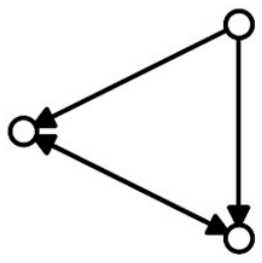 | 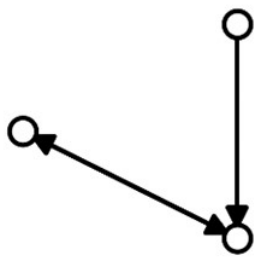 | 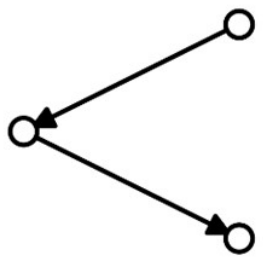 | 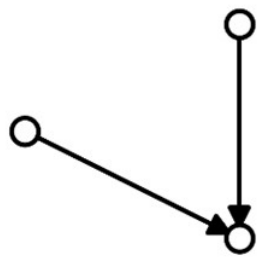 |
| **Overall** | | (2054, 3515457) | | (2132, 13708692) |
| **Y10T** | | | | (262, 1238305) |
| **G06F** | | | | |
| **Y10S** | | | | (164, 1122706) |
| **H04L** | | | | |
| **H01L** | | | | (160, 685032) |
| **A61K** | | | (108, 728048) | (94, 537996) |
| **H04N** | | | | |
| **Y02E** | | | | |
| **H04W** | | | | |
| **A61P** | (74, 2155) | (74, 75799) | (79, 620802) | |
| **Pattern #** | **Pattern 122** | **Pattern 125** | **Pattern 130** | **Pattern 141** |
| **Motif** | 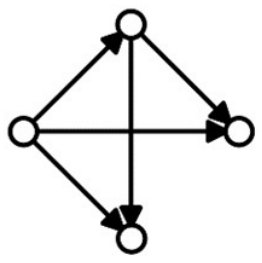 | 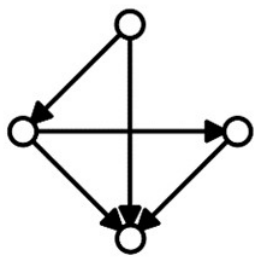 | 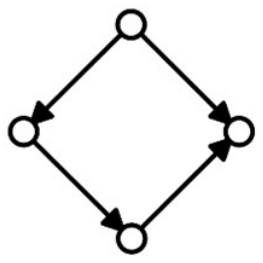 | 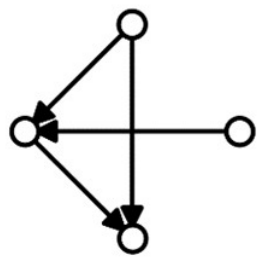 |
| **Overall** | | | (2181, 19465136) | (2050, 10909900) |
| **Y10T** | | | (255, 1476345) | |
| **G06F** | | | (258, 5484153) | |
| **Y10S** | | | (157, 1442329) | |
| **H04L** | | | (180, 4601302) | |
| **H01L** | | | (157, 816227) | |
| **A61K** | (96, 189744) | (109, 25147) | (112, 840672) | |
| **H04N** | | | (118, 2211586) | (115, 1061538) |
| **Y02E** | | (100, 14203) | (104, 827337) | |
| **H04W** | | | (126, 2804707) | (111, 1349121) |
| **A61P** | (77, 160705) | (86, 20960) | (87, 779246) | |

Surprisingly, Pattern 6 that represents a cycle [↻] is not a common motif, despite being part of various system representation schemes. It is also surprising that Pattern 12 that represents division or hierarchy [←·→] is not a common motif, despite being a fundamental part of system conceptualisations (Siddharth et al., 2020). This pattern has been referred to as "single input module" or "SIM" and is a highly significant motif in transcription networks where one gene controls a group of genes as in a hierarchical structure (Alon, 2006, pp. 75, 76). While Pattern 12 is not a common motif *per se*, it is embedded in other common motifs like Patterns 122, 125, 130, and 141.

The motifs identified in Table 3.4 suggest that knowledge explicated from artefact descriptions does not directly reflect the structure of representation schemes that are commonly used in design research and practice. The motifs that are common within domains include Patterns 8, 122, and 125 all of which combine aggregation (Pattern 13), sequence (Pattern 11) and feed-forward loop (Pattern 10), which is a highly significant motif in sensory transcription networks (Alon, 2006, p. 45). Pattern 122 in Table 3.4 has been termed "bi-fan" that is another motif in transcription networks where two genes control two other genes (Alon, 2006, p. 93).

### 3.4. Structural Basis – Qualitative Subgraphs

In transcription networks, a directed edge denotes that the transcription factor protein of one gene controls the rate of transcription of another (Alon, 2006, p. 12). As every link in such a network means the same, it is simpler to interpret motifs and generalise these, e.g., bifan being generalised to dense overlapping regulations (Alon, 2006, p. 93). The motifs shown in Table 3.4 represent a large set of subgraphs, each of them including specific relationships. It is therefore necessary to examine the most common subgraphs represented by the motifs. We first examine the top 3 most frequent subgraphs under 8 patterns as shown in Supplementary Material, Table S-6. Based on the frequency proportions, we understand that Patterns 11, 13, and 122 have generalisable subgraphs and thus examine these patterns further.

In Table 3.5, we show the top 3 subgraphs under these patterns, including nodes labelled using entity syntaxes. In Supplementary Material, Table S-7, we list subgraphs at different percentiles under these patterns. In Pattern 13 that represents aggregation [→·←], the most frequent subgraphs involve the "of" relationship between abstract entities ("the NN"). In addition, "are" is used to define specific entities ("the JJ NN JJ NNS") w.r.t., abstract entities ("NNS"). In Pattern 11 that represents sequence [→·→] involves "in" and "of" among abstract entities. Pattern 122, known as bifan, captures one attribute ("of") and two hierarchical ("include") relationships. As we explain in Section 4.1, abstract entities in these patterns could be concretised by collapsing "of" facts, leading to simplification of knowledge structures, e.g., Pattern 122 could be simplified to Pattern 12 that represents a division or hierarchy [←·→]. Overall, motif analysis led to 8 distinct patterns whose further subgraph examination helped us generalise patterns into the structural basis that includes aggregation [→·←], sequence [→·→], and hierarchy [←·→].

**Table 3.5**: Top 3 subgraphs with entity syntaxes under selected motifs.

| **Pattern 13** | | |
|---|---|---|
| Raw Count - 13,708,692, Number of Unique Graphs - 9,637,932 | | |
| 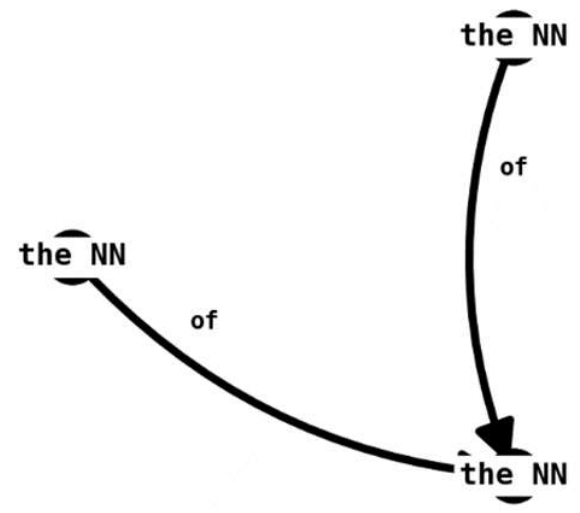 | 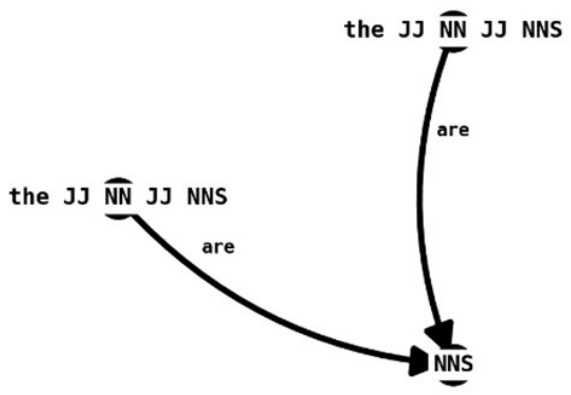 | 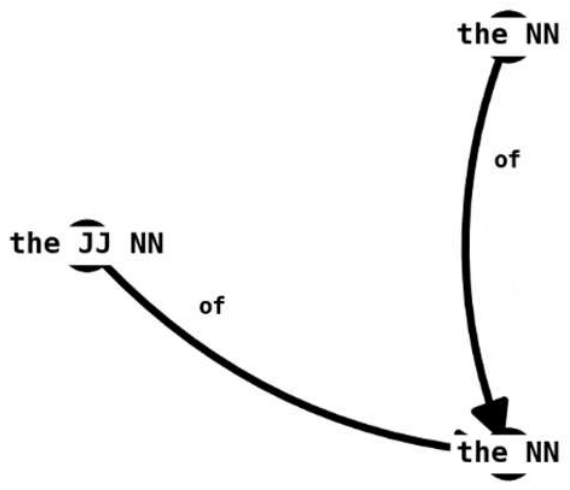 |
| 11,406 | 5,544 | 4,581 |
| **Pattern 11** | | |
| Raw Count - 19,213,782, Number of Unique Graphs - 12,750,460 | | |
| 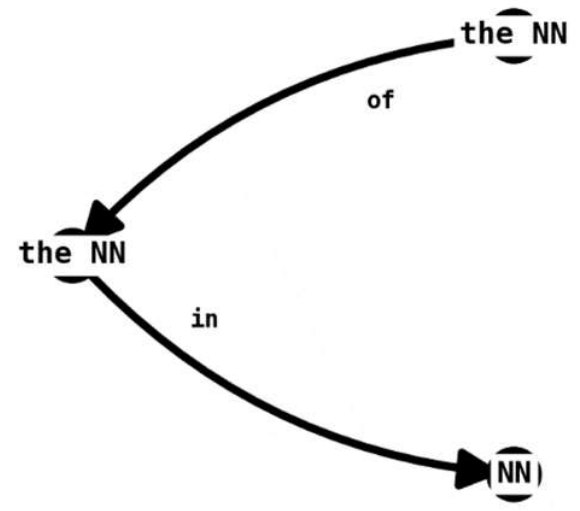 | 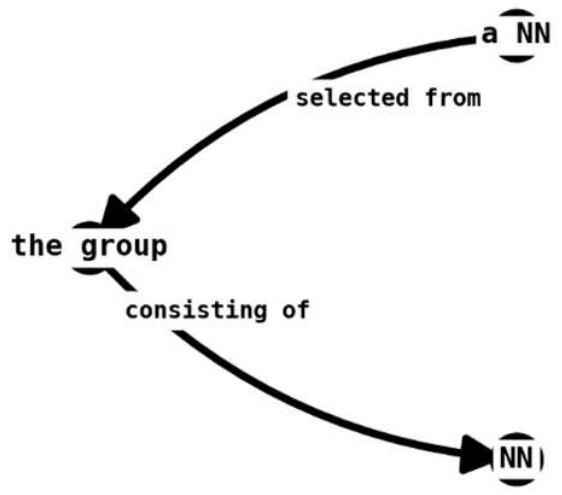 | 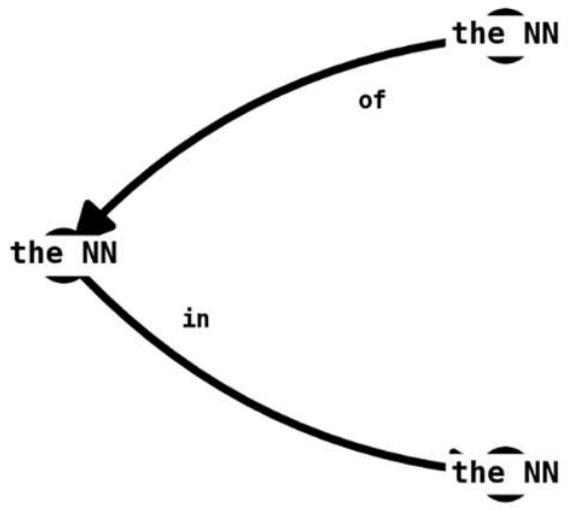 |
| 1,705 | 997 | 973 |
| **Pattern 122** | | |
| Raw Count 988,868, Number of Unique Graphs - 359,181 | | |
| 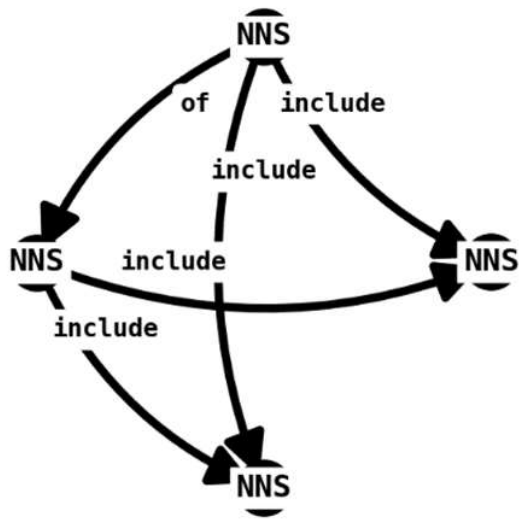 | 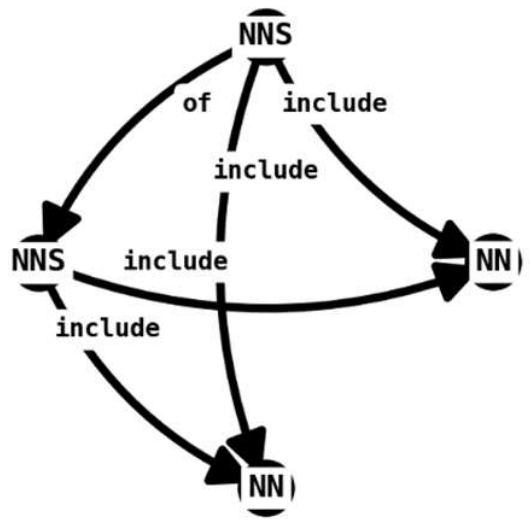 | 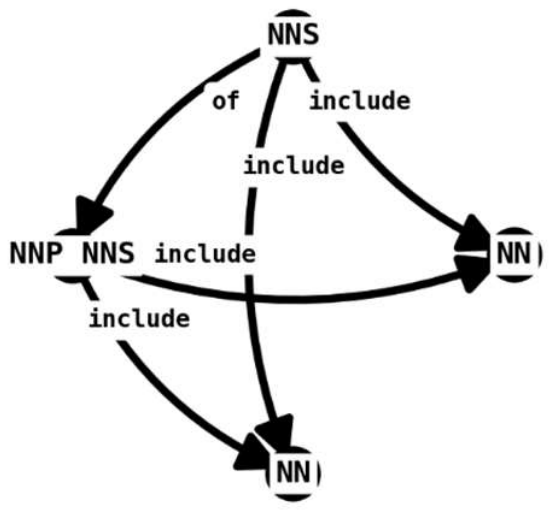 |
| 3,621 | 2,421 | 1,312 |

# 4. Discussion

## 4.1. Concretisation Strategies

Network representations could be locally collapsed or pruned to highlight meaningful portions that help better interpretability. Alon (2006, p. 91) explicates the global structure of the *E. coli* transcription network by collapsing the subgraphs into corresponding motifs. In engineering design, Caldwell and Mocko (2012, p. 3) propose eight pruning rules to be applied to functions and flows in order to explicate the design knowledge in functional models, e.g., remove all distribute functions referring to energy. Building upon the understanding of linguistic and structural basis, in this section, we suggest concretising abstract and most common elements of the knowledge graphs so that meaningful elements are better highlighted.

As illustrated in Table 4.1, we suggest three strategies for concretising entities and relationships and eliminating redundant hierarchical relationships, all of which could lead to simplified and/or modular knowledge structures. As illustrated with a Pattern 130 subgraph in Table 4.1, abstract entities (NN or a NN) associated by “of” relationship could be combined into a relatively concrete entity, e.g., “alkoxysiloxane hydrolysis”. While this strategy is applied in the Pattern 9 subgraph as well, the relationship “to” be concretised thereupon, by introducing an appropriate verb like “reduced to”. In the Pattern 122 subgraph, since “examples” and “the material” are hierarchically associated with the same entities, redundant links could be eliminated by collapsing the entities to “material examples”. These strategies help communicate design knowledge with fewer nodes/edges and with simpler knowledge structures. We provide additional examples of these strategies in Supplementary Material, Table S-8.

**Table 4.1**: Illustrating concretisation with common motifs

| | | | |
|---|---|---|---|
| Entities<br>(e.g., Pattern 130[11]) | 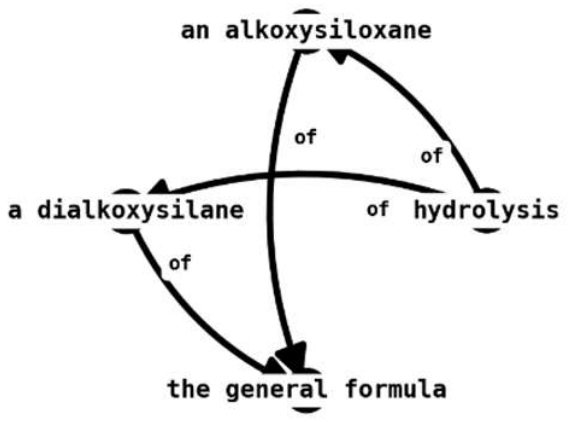 | → | 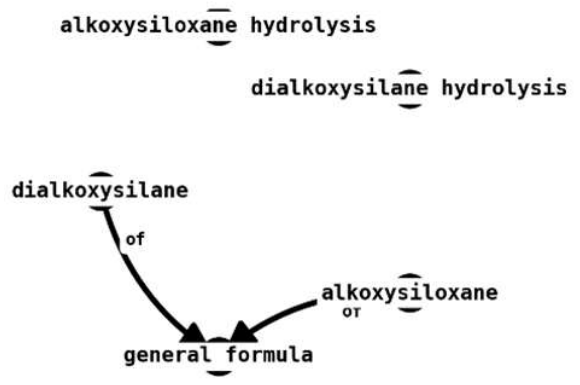 |
| Relationships<br>(e.g., Pattern 9[12]) | 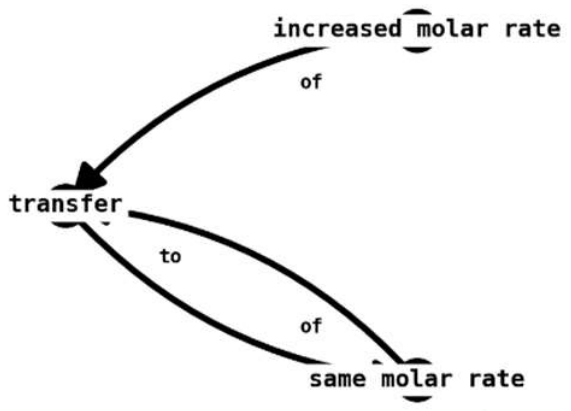 | → | 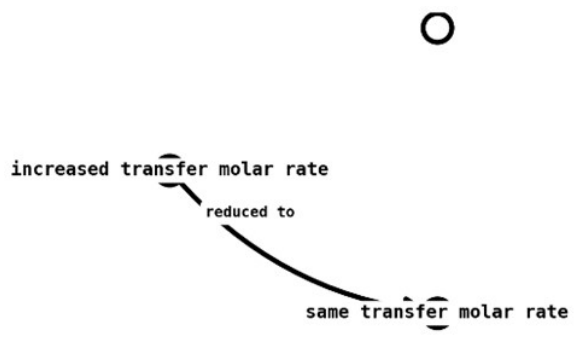 |
| Hierarchical Structures<br>(e.g., Pattern 122[13]) | 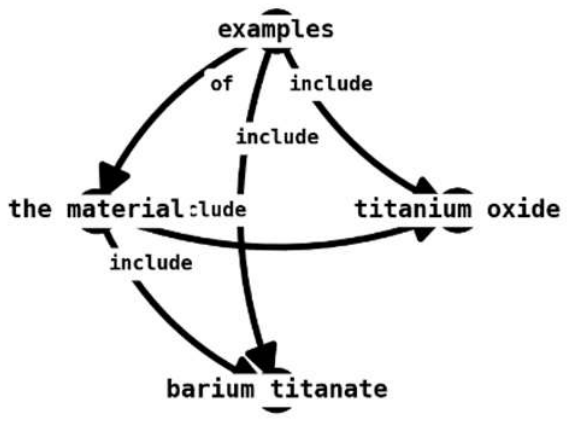 | → | 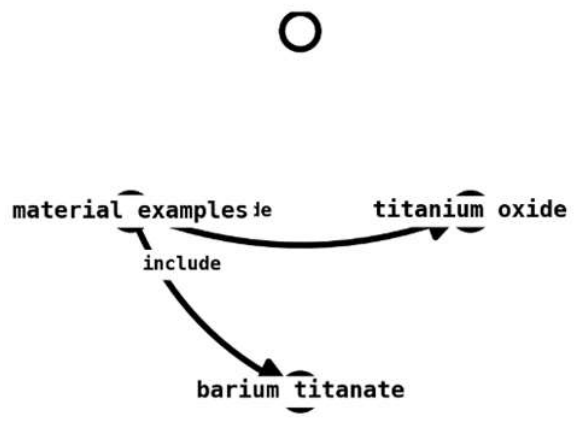 |

[11] Process for preparing low molecular weight organosiloxane terminated with silanol group - https://patents.google.com/patent/US5576408/

[12] System and method for simultaneous evaporation and condensation in connected vessels - https://patents.google.com/patent/US10876772/

[13] Dispersion liquid, composition, film, manufacturing method of film, and dispersant - https://patents.google.com/patent/US10928726/

To demonstrate these strategies with actual products, we retrieve around 3,000 facts each from patents on the coffee grinder and glue gun that were earlier used as representative examples for functional models (Mokhtarian et al., 2017, p. 495; Stone and Wood, 2000, p. 364) and state-flow diagrams (Yildirim et al., 2017, p. 425). We summarise the knowledge bases thus populated in Supplementary Material, Table S-9. Upon filtering the recurring facts for each product, we examine the knowledge surrounding the product entity as shown in Figure 4.1 for the coffee grinder. For the glue gun, which is relatively complex, we provide the knowledge in Supplementary Material, Figure S-5. In Figure 4.1-a, the facts surrounding the "frustonical burr" with "of" relationships could be collapsed into concrete entities like "internal frustonical burr bore". The abstract relationships like "on" could also be concretised to "rested on" such that the overall knowledge graph, as shown in Figure 4.1-b, becomes more modular and easier to construct, manage, and evolve.

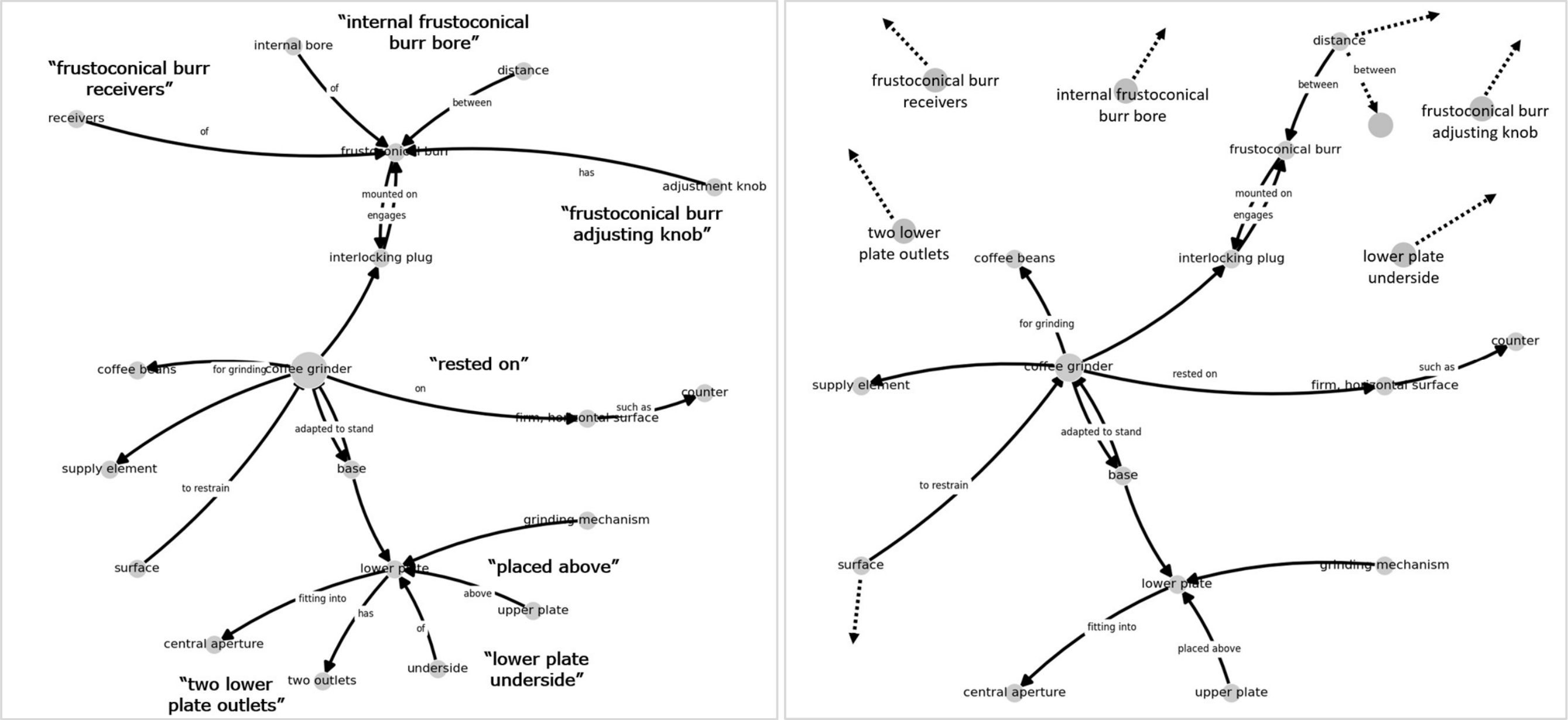

**Figure 4.1**: a) extended neighbourhood of "coffee grinder" entity; b) transformed neighbourhood thereof. The edges without a label indicate a hierarchical relationship – "include".

The necessity behind concretising and modularising design knowledge is often not emphasised in the literature, while many scholarly contributions call for the modular design of artefacts. For instance, through the simulation of product architectures with varied cyclical degrees and interaction densities, Luo (2015, p. 365) finds that those with higher interaction densities are less likely to evolve. Whether or not an artefact is modular significantly depends on how it is "perceived" and modelled into a representation scheme. In DSM literature, Rowles (2003, p. 241) supposedly decomposes the Pratt & Whitney aircraft engine into eight systems and further into a mere 54 components.

Even a hierarchical system is only nearly decomposable, as in, the temporal behaviour of a subsystem is approximately independent of others, while the outcomes of these behaviours are aggregated (Simon, 1962, p. 474). While densely connected artefacts are hard to manage, the challenges are not only associated with the actual design but also with the knowledge concerning these. For example, cricket helmets are simple, modular artefacts that used to be available in different styles and sizes to suit different players. Post the death of Phil Hughes on the pitch, the helmet design had to be revamped[14] for protection against various anticipated injuries. In such cases where performance is critical, even simple artefacts could draw extensive amounts of knowledge, managing which is crucial in a design environment.

[14] Masuri Helmets - https://www.bbc.com/sport/cricket/31421607

### 4.2. LLM Implications

The motivation of our research is to advance traditional views of design knowledge and more importantly to support LLMs in retrieving and generating knowledge that is useful for a design environment. In our prior work, we proposed and demonstrated Retrieval-Augmented Generation (RAG) using design knowledge explicated from patented artefact descriptions (Siddharth and Luo, 2024a, 2024c). In this paper, we presented an empirical understanding of design knowledge from linguistic and structural perspectives. While RAG is useful for guiding LLMs in terms of instantaneous knowledge, i.e., the content of text generated, the findings reported in this paper could gauge LLMs in terms of lexical and syntactic characteristics. We therefore envisage an advanced usage of LLM as depicted in Figure 4.2.

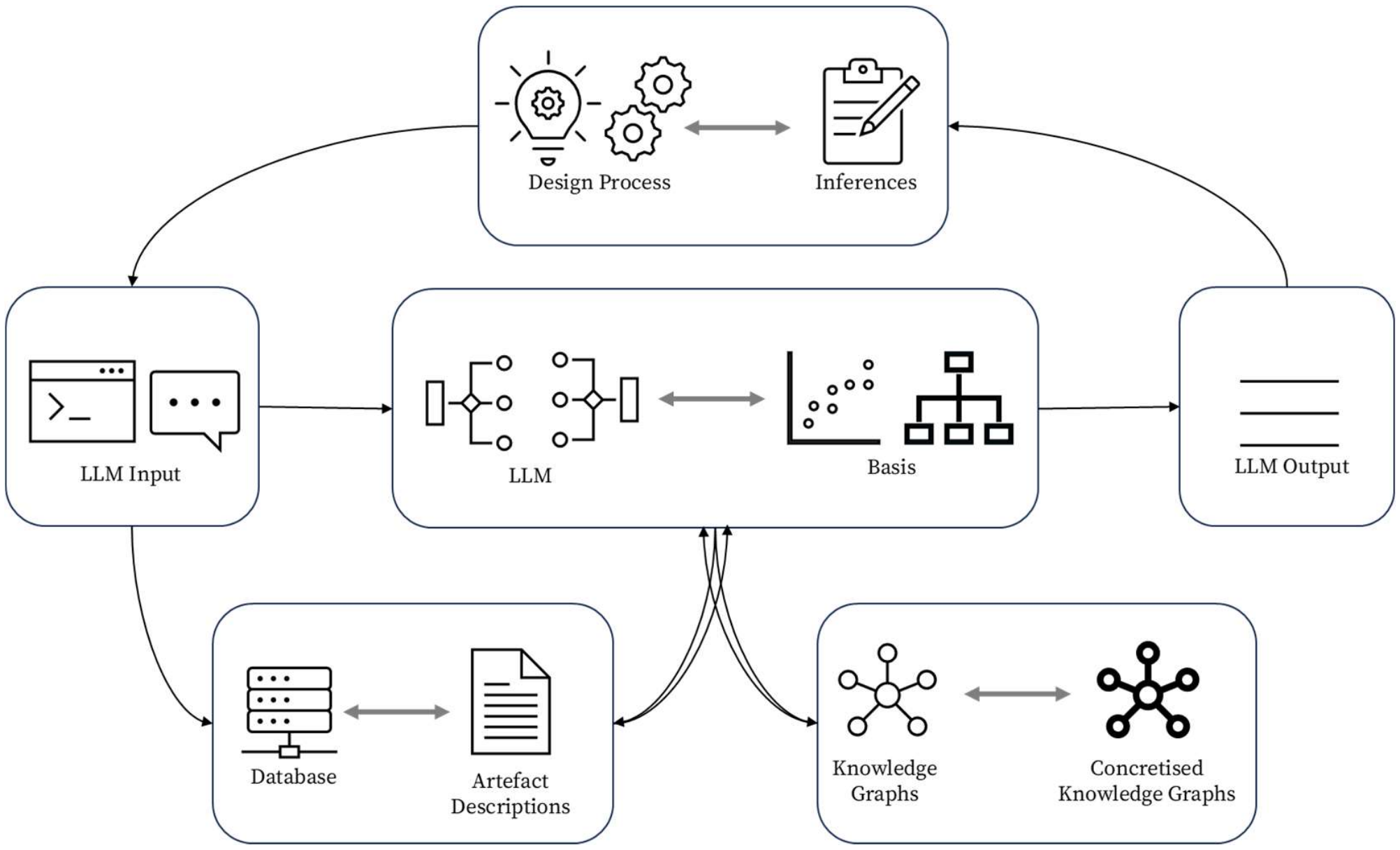

**Figure 4.2**: Envisaged usage scenario for LLMs and related applications.

In Section 1.3, we illustrated how GPT-4 is not yet suitable for direct usage in a design environment and that it should be contextualised with explicit design knowledge. RAG has been gaining popularity along with the rise of LLMs, as several domain applications require fact-based retrieval of knowledge. As indicated in Figure 4.2, it involves the usage of existing databases and text-retrieval methods to formulate a suitable context for LLMs to generate text. In our prior work (Siddharth and Luo, 2024a), we demonstrated that LLMs are not suitable to directly handle large amounts of artefact descriptions in a design environment and that design knowledge must be explicated and processed before providing context for LLMs. Since RAG may not be feasible in all design environments, the findings from our current analysis, i.e., the basis of design knowledge, could be used to gauge the LLM responses lexically and syntactically as indicated in Figure 4.2.

Our analysis reveals how design knowledge is linguistically and structurally constructed at different levels of abstraction. These findings could help build rule-based systems that gauge and dictate LLMs as to how entities are formulated, how these are associated, how relationships are specified, and how these relationships are syntactically organised in the sentences. To enable such a scenario, design knowledge needs to be concurrently explicated as knowledge graphs and simultaneously examined by the rule-based systems. Such systems could be more advanced by incorporating the concretisation strategies, wherein, LLMs could borrow cues from our findings to locate and transform abstract facts as illustrated in Table 4.1 and Figure 4.1. These systems could scalably and reliably control LLM responses to make these more suitable for a design environment, irrespective of the availability of RAG systems.

### 4.3. Limitations and Further Research

Our research as reported in this paper could address several limitations. First, the purview of our analysis is restricted to patented artefact descriptions that do not include other sources of artefact descriptions like research papers and textbooks. The analysis could also delve deeper into a domain to understand specific insights therewithin. The basis of design knowledge has been understood purely from a text standpoint, while significant portions of design knowledge are communicated through visual forms. Since patent documents explain every part of figures both structurally and functionally, the visually communicated knowledge is indirectly incorporated. Even from a text standpoint, the linguistic syntaxes do not include different senses of a term, e.g., “long chain” could be understood differently across domains.

In terms of the structural basis, the motif analysis could have involved comparing the original patent knowledge graph against an ensemble of randomised graphs. However, due to the sparsity of the knowledge graphs in our data and the additional constraints that we pose while randomisation, it is difficult to obtain several distinct randomised graphs. In some cases, we could only obtain the randomised graph with 30-50 % graph even after 30,000 iterations of the curveball algorithm. Sparsity is present in sensory transcription networks as well, i.e., Escherichia coli network has about 400 nodes and 500 edges, which are interactions selected by evolution among all mutations (Alon, 2006, p. 43). The studies that involved these networks involved constraint-free randomisation, whereas in our case, we retain not only network size but also degree distribution and non-self edges. Overall, our analysis should further delve into the meanings of individual motifs and their implications towards the future evolution of similar artefacts/domains.

The current research outcomes and limitations could instigate several directions for the future. The basis of design knowledge as reported in this paper, shall be deployed and tested for synthesis tasks in design environments, where feasible combinations of entities could be generated, particularly in embodiment design. Stemming from the primary motivation of this research, we are yet to experimentally compare existing, traditional perspectives of design knowledge against the empirical standpoint established in this paper in terms of supporting knowledge retrieval and representation. The generic concretisation strategies proposed in this paper are yet to be studied for complex artefact design and with suitable interfaces (powered by LLMs as indicated in Figure 4.2).

Since the explication of design knowledge as knowledge graphs helped us identify not only linguistic but also structural basis, the graph medium could serve various purposes in design research, education, and practice. In addition to organisational and economic factors, research into technology and innovation could leverage artefact knowledge graphs to predict technology performances such as sales, user satisfaction, etc. Subject to computational challenges, the use of knowledge graph representations for artefacts could also enable novelty assessment (Siddharth et al., 2019). The analysis and the underlying methods adopted in this paper could be conducted within a domain to reveal generalisable design knowledge to build domain ontologies and content for engineering design education. The analysis could also done with a portfolios of designers/engineers to assess their collective creativity and domain knowledge.

# Supplementary Material

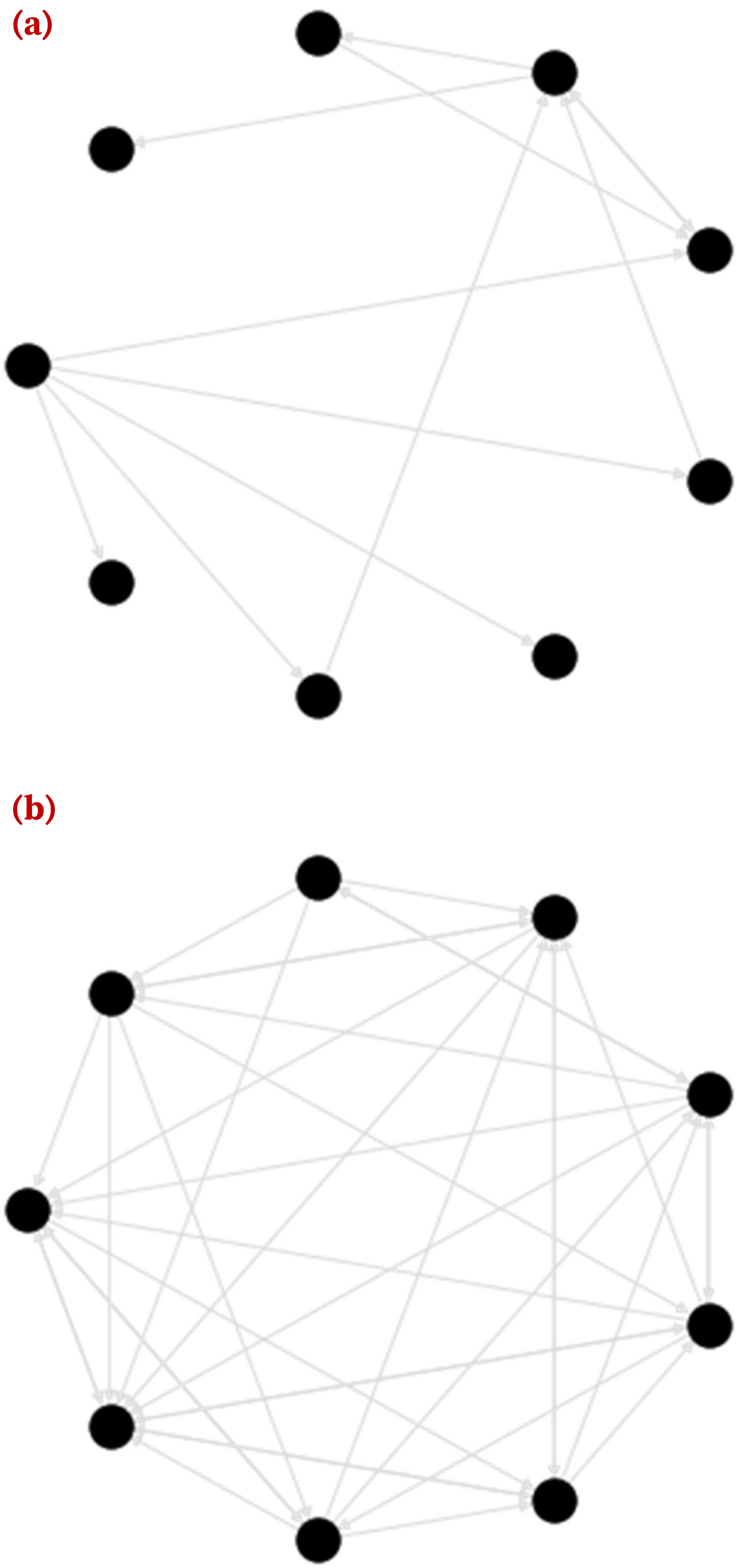

**Figure S-1**: Illustrating the structures of a) a knowledge graph in our data vs. b) a random network of similar number of nodes. The knowledge graph is based on a patent “Electrical power generating plant” - https://patents.google.com/patent/US3943374A/

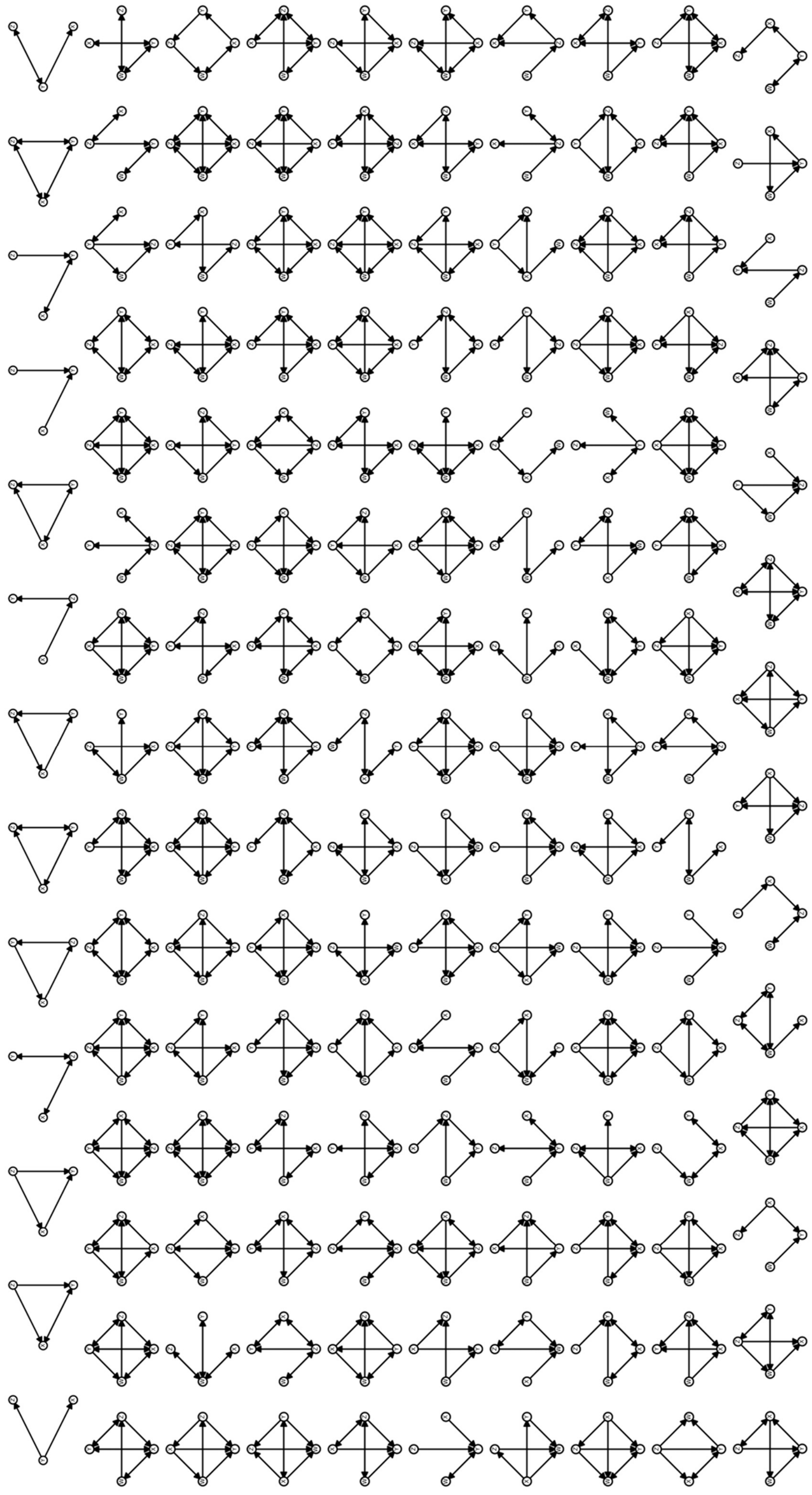

**Figure S-2**: All 3-node and 4-node subgraph patterns.

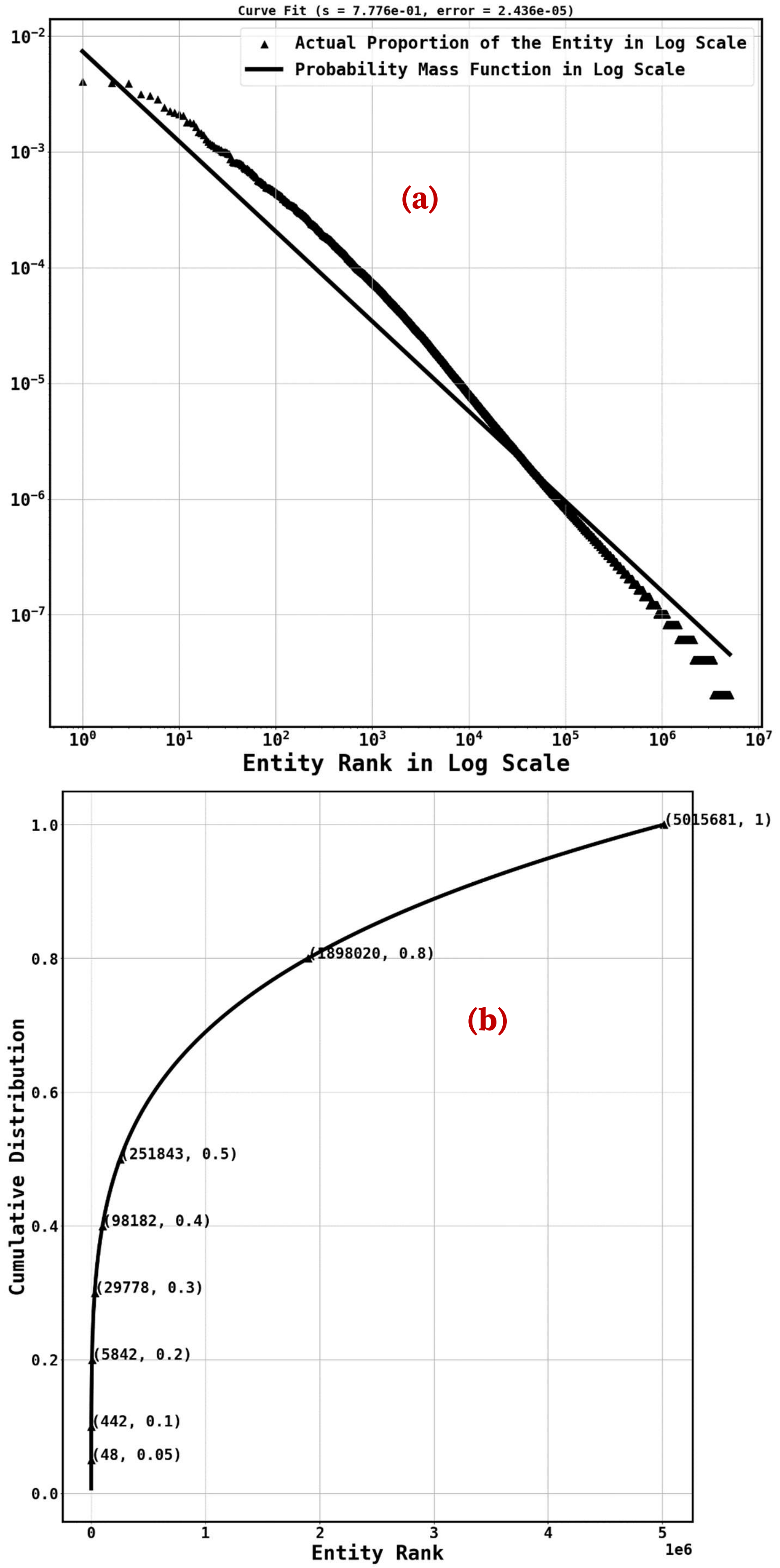

**Figure S-3**: Zipf Distribution based on the frequency of unique entities: a) probability mass function in log scale against the actual proportion of entities, b) cumulative distribution indicating the counts at various percentiles.

**Table S-1**: Overview of entities at different percentiles as per cumulative Zipf distribution.

| **Entities<br>Zipf Distribution: s = 7.776e-01, error = 2.436e-05** | | |
|---|---|---|
| **CDF** | **Count** | **Terms (Top 30)** |
| **(0, 0.05]** | 48 | the present invention, a plurality, each, the method, the invention, examples, data, the user, a method, information, the group, use, the system, order, accordance, respect, the plurality, means, the number, the amount, response, a user, the surface, the device, the use, methods, a portion, part, the form... |
| **(0.05, 0.1]** | 394 | operation, the vehicle, a set, instructions, step, the range, weight, addition, communication, this invention, the direction, an amount, the present disclosure, the position, the housing, an embodiment, a combination, the step, the case, power, the output, a system, the information, components, combination, an object, the steps, the length, place, the composition... |
| **(0.1, 0.2]** | 5,400 | network, the operations, regard, the liquid, the principles, at least some, the cells, a preferred embodiment, fuel, the head, the values, another, the light source, blocks, a level, selection, layers, the channel, a width, the event, image data, the top surface, characteristics, the processing, the tube, a sample, insertion, a pressure, the solution, the inside... |
| **(0.2, 0.3]** | 23,936 | inert gas, the computer program instructions, a word, ozone, any medium, the beacon, electrical components, rings, the photoresist, extent, the outside diameter, a gui, the stroke, other ingredients, the dielectric material, heat energy, a passenger, the virus, the selector, an organic material, ic, a dna sequence, a software program, the discovery, indium, a manifold, the driving, the defect, the reaction temperature, the base stations... |
| **(0.3, 0.4]** | 68,404 | the ups, controlling, a synthetic polymer, the organic light emitting diode, said request, the terminal devices, a preset value, standardized or proprietary technology, an alternative configuration, predetermined criteria, rinsing, colourants, a water pump, a start signal, an execution screen, a high accuracy, separate units, client computing devices, the lighting element, optional additives, the compliance, screening assays, the bolus, reproducibility, antennae, a same type, capacitor c, predetermined times, the management module, a position signal... |
| **(0.4, 0.5]** | 153,661 | the activator support, the yaw angle, the stroke length, attenuators, the labeler, optical fibres, a hydride, the welding current, the tat polypeptide, the semiconductor package structure, the usage information storage server, these axes, an ngff module, a ramp signal, the locking direction, the servo signal, a low dielectric constant film, the imaging area, other animal, suitable vehicles, further characteristics, the bottom electrodes, a local processor, at least 20 wt, the picture signal, the recorded details, a front facing surface, such diluents, the pos server, the planar light source... |
| **(0.5, 0.8]** | 1,646,177 | the measuring capacitor, rns, brownian motion, at least a first image, the candidate user, the time frequency resource, enode b, isocyanatoethyl acrylate, the solvent vapor discharge nozzle, the desired engine torque, a natural sequence, two tables, specific keys, the extruded tube, the seeding vessel, the user supplied storage device, engine noise, ftest, optical frequency, a first flange portion, a particular language, a video game console, a second flange portion, the silane containing adhesion promoter, the input fields, two effects, the methylation, this accumulation, the target memory address, predictive model generator... |
| **(0.8, 1]** | 3,117,660 | as many passes, both the internal and external surfaces, the normal rigidity, a write port, read interconnects, an external diameter c, sufficient layers, write interconnects, a select signal line, a select hierarchy i/o line, a proportionally greater quantity, the relatively thin end, proportionally more composite material, suitable ribs, wind turbine constructions, no vibration damping elements, the structural forces, suitable composite shafts, the n push lines, the generator stator, a generator coupling, the n lines, the push switching matrix portion, the aligned or registering central cone opening, one or more desired 360 degree rotational positions, the appropriate linear position, the radial data points, a passgate, rotary table motor... |

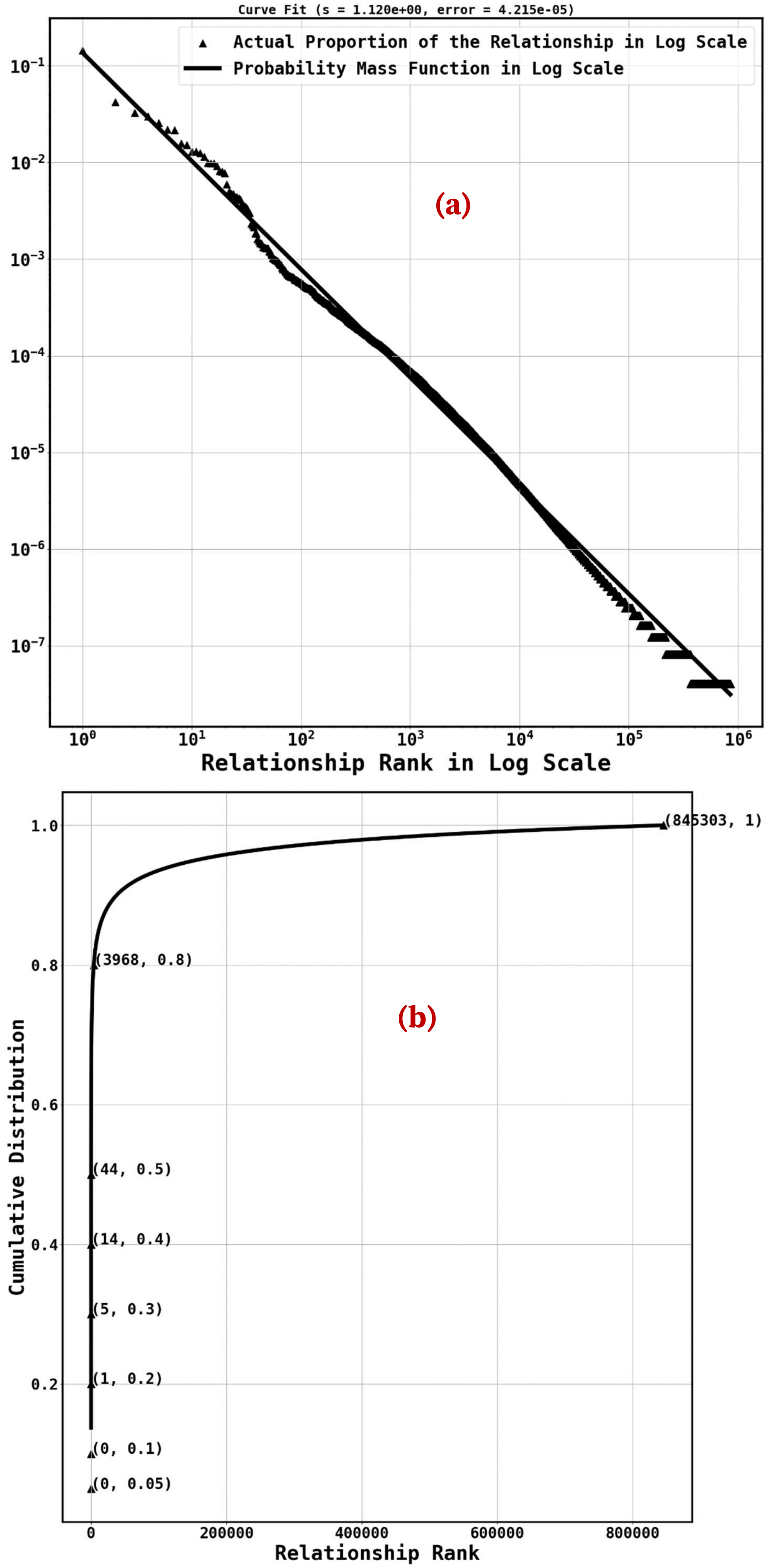

**Figure S-4**: Zipf Distribution using frequency of unique relationships: a) probability mass function in log scale against the actual proportion of relationships, b) cumulative distribution indicating the counts at various percentiles.

**Table S-2**: Overview of relationships at different percentiles according to cumulative Zipf distribution.

| **Relationships**<br>**Zipf Distribution: s = 1.120, error = 4.215e-05** | | |
|---|---|---|
| **CDF** | **Count** | **Terms (Top 30)** |
| **(0, 0.05]** | 0 | |
| **(0.05, 0.1]** | 0 | |
| **(0.1, 0.2]** | 1 | of |
| **(0.2, 0.3]** | 4 | in, to, include, such as |
| **(0.3, 0.4]** | 9 | for, with, includes, from, including, wherein, is, between, on |
| **(0.4, 0.5]** | 30 | by, having, at, comprises, comprising, has, as, have, through, into, using, are, based on, comprise, according to, within, consisting of, be, provides, associated with, to provide, provide, containing, over, corresponding to, used in, is in, connected to, to form, along |
| **(0.5, 0.8]** | 3,924 | relates to, contains, coupled to, selected from, relative to, where, is connected to, contain, via, receives, under, without, during, stored in, allows, per, about, applied to, to determine, refers to, use, used for, described in, form, included in, to produce, due to, to generate, receive, against |
| **(0.8, 1]** | 841,335 | operates with, exposed by, to distinguish, is attached via, accounts for, to express, constituted of, mounting, are both, is calculated using, is present, excludes, to restore, engageable with, to mount, located to, though, to encode, resting on, projects from, is immersed in, is providing, are integrated into, employed such as, is generated using, mixing, by accessing, mixed in, sufficient to provide, referenced throughout |

**Table S-3:** Motifs based on 3-node subgraph patterns.

| Pattern # | Motif | Unique Sequence | Patent Count ↓ | Raw Count |
|---|---|---|---|---|
| Pattern 13 | 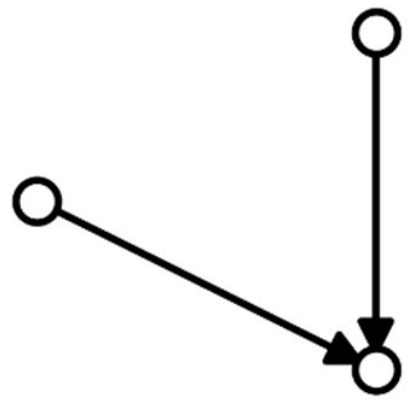 | ‘002011011’ | 2,132 | 13,708,692 |
| Pattern 9 | 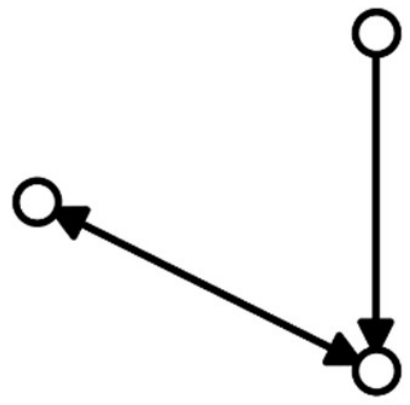 | ‘012111012’ | 2,054 | 3,515,457 |
| Pattern 8 | 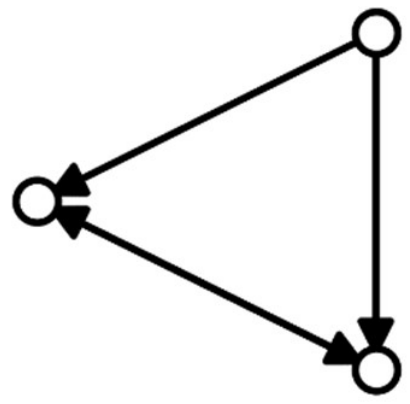 | ‘022112112’ | 1,982 | 86,509 |
| Pattern 11 | 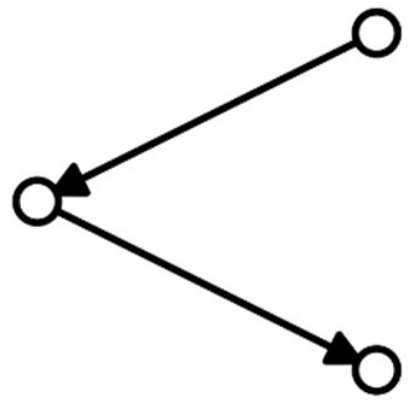 | ‘011011011’ | 1,912 | 19,213,782 |
| Pattern 3 | 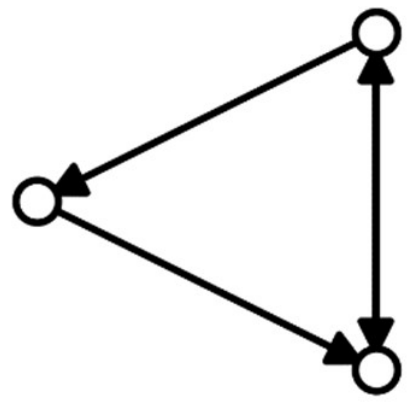 | ‘112112112’ | 1,880 | 104,548 |
| Pattern 2 | 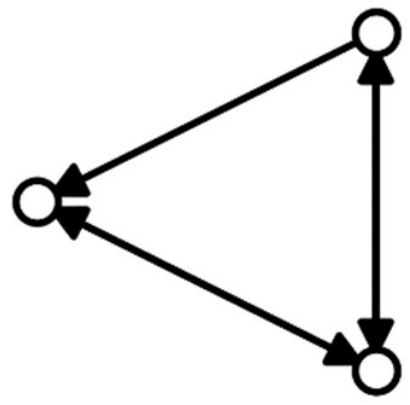 | ‘122122122’ | 1,854 | 72,898 |
| Pattern 7 | 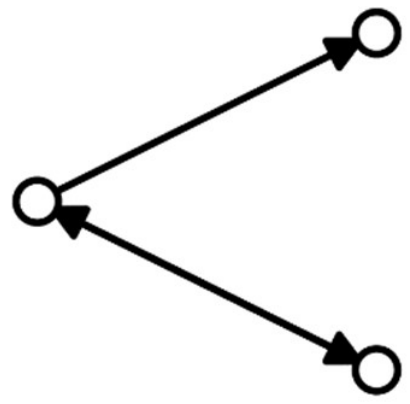 | ‘111012012’ | 1,824 | 3,375,974 |
| Pattern 10 | 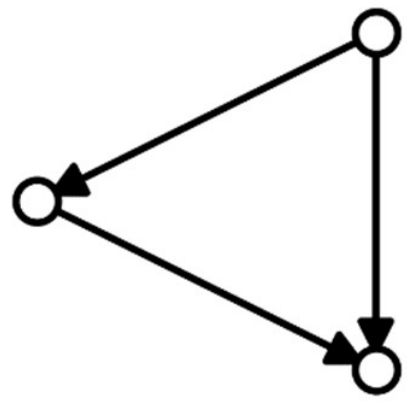 | ‘012012111’ | 1,678 | 479,010 |

| Pattern # | Motif | Unique Sequence | Patent Count ↓ | Raw Count |
|---|---|---|---|---|
| Pattern 5 | 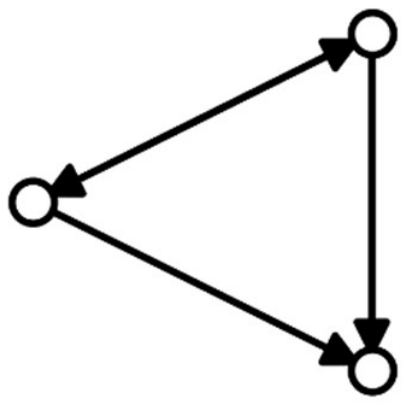 | ‘112022112’ | 1,598 | 64,558 |
| Pattern 4 | 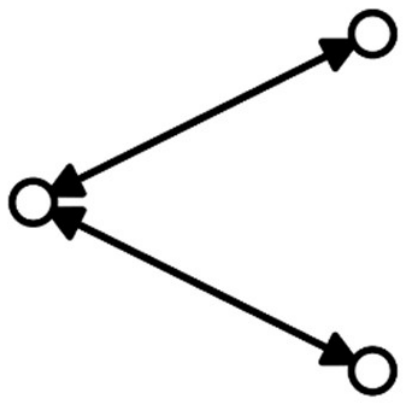 | ‘112112022’ | 1,429 | 294,341 |
| Pattern 6 | 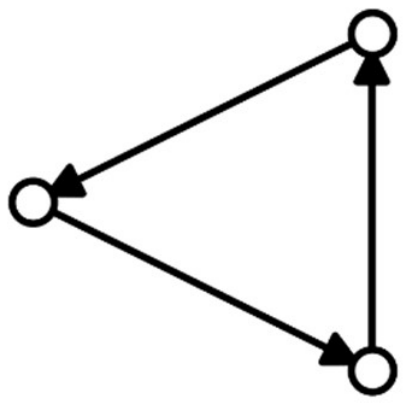 | ‘111111111’ | 1,350 | 29,421 |
| Pattern 1 | 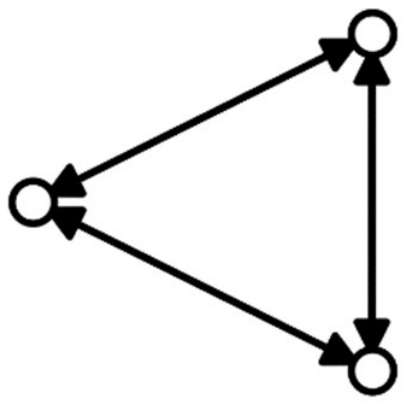 | ‘222222222’ | 1,349 | 11,329 |
| Pattern 12 | 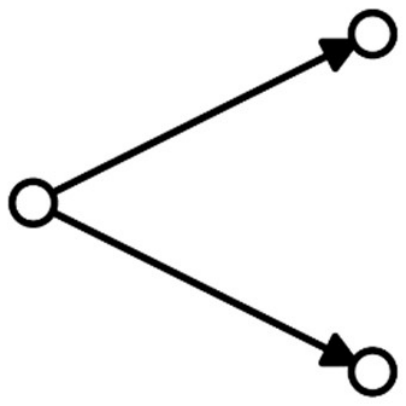 | ‘011002011’ | 1,205 | 14,522,467 |

**Table S-4**: Most common motifs based on 4-node subgraph patterns.

| Pattern # | Motif | Unique Sequence | Patent Count ↓ | Raw Count |
|---|---|---|---|---|
| Pattern 130 | 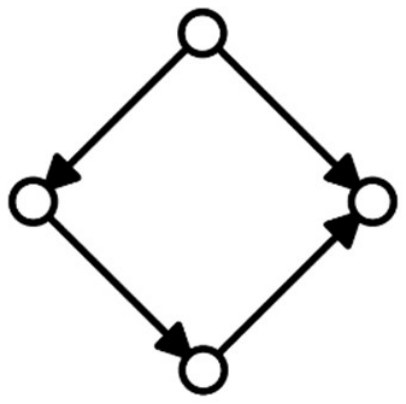 | ‘01120112001111’ | 2,181 | 19,465,136 |
| Pattern 141 | 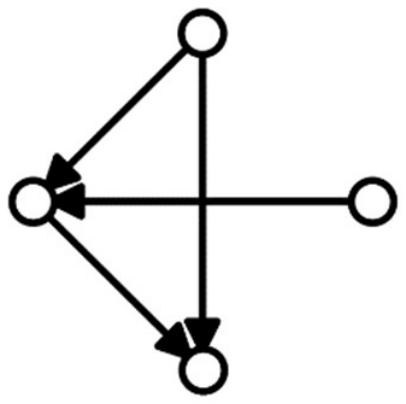 | ‘00220112001111’ | 2,050 | 10,909,900 |
| Pattern 125 | 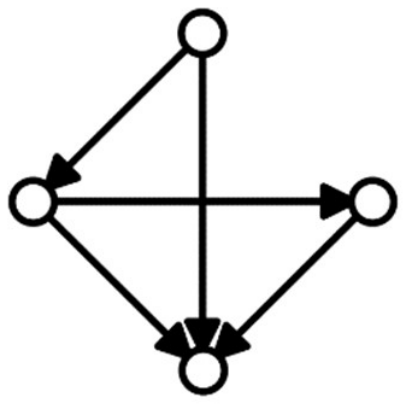 | ‘01130122011111’ | 1,873 | 345,797 |
| Pattern 144 | 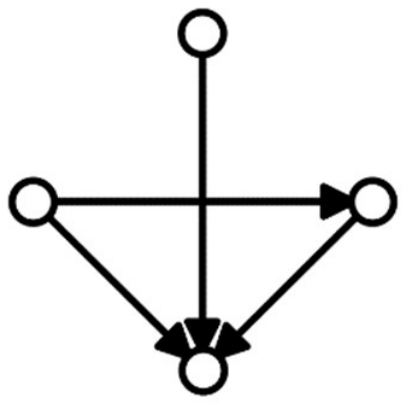 | ‘00130112001111’ | 1,771 | 12,423,358 |
| Pattern 110 | 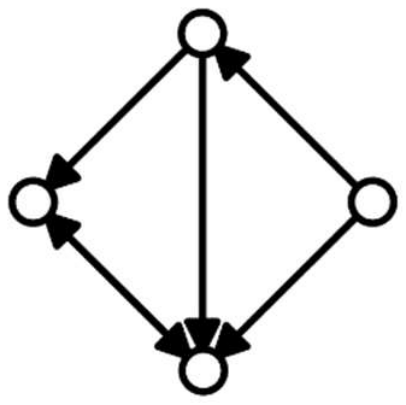 | ‘01231122011112’ | 1,768 | 486,726 |
| Pattern 119 | 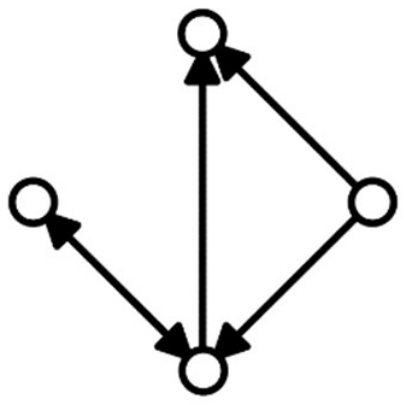 | ‘01220122001112’ | 1,682 | 11,304,395 |
| Pattern 138 | 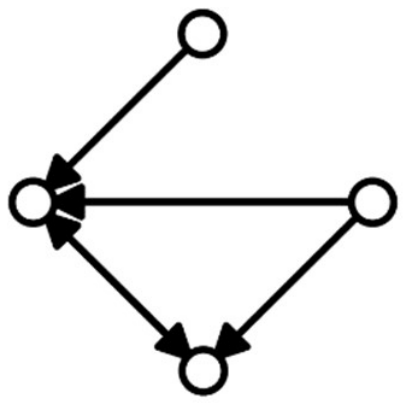 | ‘00231112001112’ | 1,663 | 5,849,555 |
| Pattern 116 | 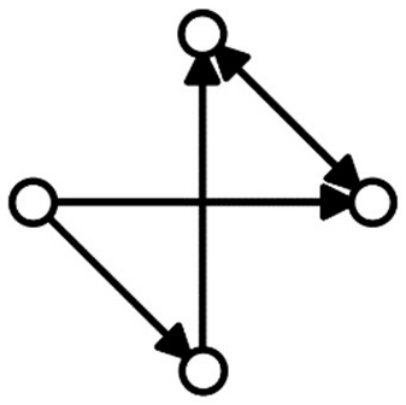 | ‘01221112001112’ | 1,644 | 5,102,086 |

**Table S-5**: Motifs based on 4-node subgraph patterns (continued).

| **Motif Pattern # Unique Sequence (Patent Count, Raw Count)** | **Motif Pattern # Unique Sequence (Patent Count, Raw Count)** | **Motif Pattern # Unique Sequence (Patent Count, Raw Count)** | **Motif Pattern # Unique Sequence (Patent Count, Raw Count)** | **Motif Pattern # Unique Sequence (Patent Count, Raw Count)** | **Motif Pattern # Unique Sequence (Patent Count, Raw Count)** |
|---|---|---|---|---|---|
| 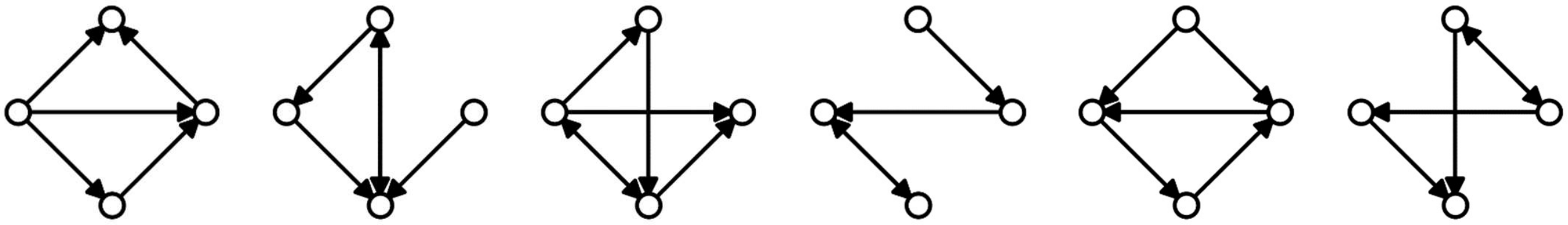 | | | | | |
| Pattern 120<br>‘01220113011111’<br>(1540, 306865) | Pattern 123<br>‘01131112001112’<br>(1512, 5434973) | Pattern 72<br>‘11220123011112’<br>(1504, 339970) | Pattern 129<br>‘01121111000112’<br>(1492, 8762201) | Pattern 115<br>‘01221112011111’<br>(1454, 112045) | Pattern 85<br>‘11120122001112’<br>(1432, 5057308) |
| 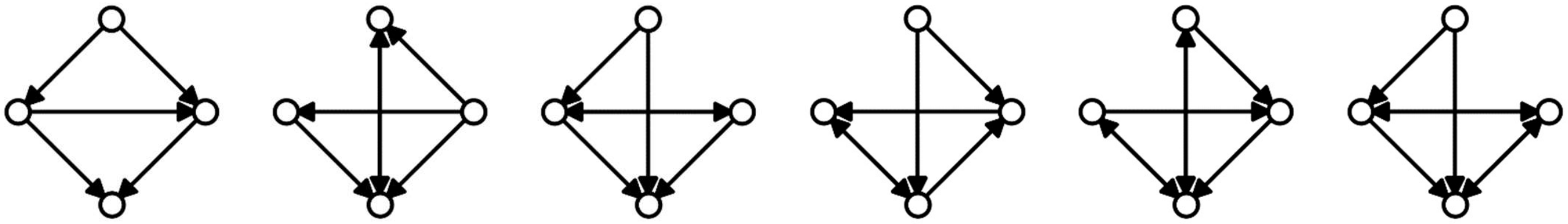 | | | | | |
| Pattern 118<br>‘01220122011111’<br>(1383, 341909) | Pattern 112<br>‘01231113011112’<br>(1365, 120540) | Pattern 113<br>‘01230222011112’<br>(1361, 112201) | Pattern 103<br>‘02221122011112’<br>(1355, 59389) | Pattern 56<br>‘11231222011122’<br>(1330, 149665) | Pattern 98<br>‘02231222011122’<br>(1329, 203003) |
| 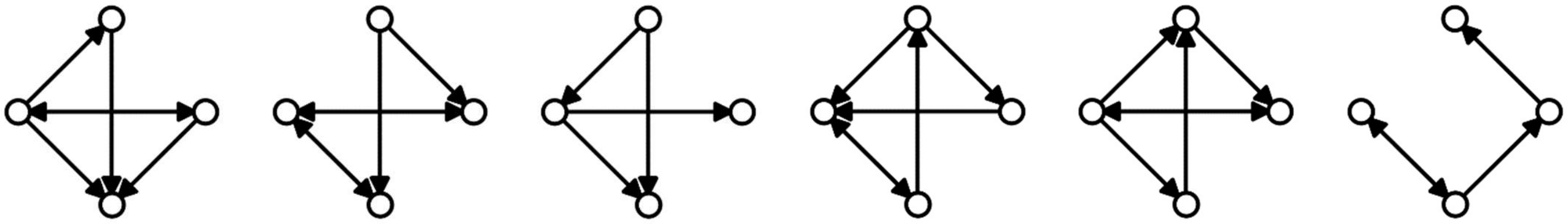 | | | | | |
| Pattern 80<br>‘11130123011112’<br>(1328, 98794) | Pattern 104<br>‘02221122001122’<br>(1326, 1175531) | Pattern 132<br>‘01120022001111’<br>(1326, 7409621) | Pattern 75<br>‘11131122011112’<br>(1324, 53409) | Pattern 68<br>‘11221113011112’<br>(1322, 53367) | Pattern 92<br>‘11110112000112’<br>(1320, 8676609) |
| 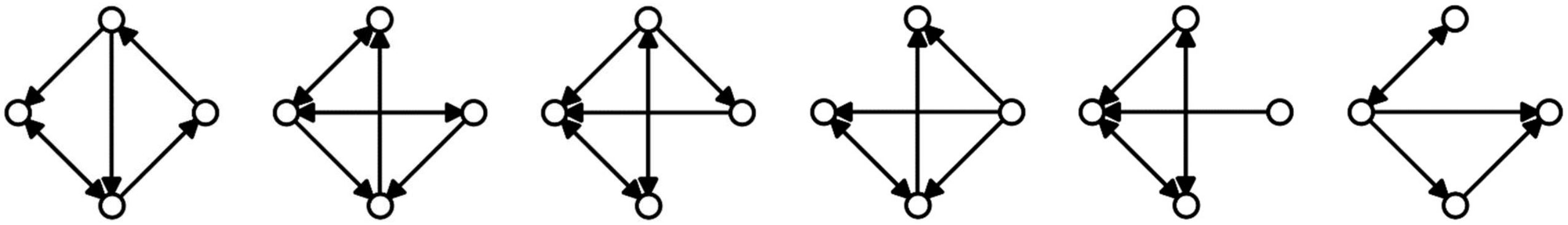 | | | | | |
| Pattern 65<br>‘11221122011112’<br>(1313, 177098) | Pattern 46<br>‘12221123011122’<br>(1300, 141085) | Pattern 59<br>‘11231123011122’<br>(1296, 303996) | Pattern 107<br>‘02220123011112’<br>(1290, 112615) | Pattern 111<br>‘01231122001122’<br>(1288, 4855701) | Pattern 87<br>‘11120113001112’<br>(1283, 5168289) |
| 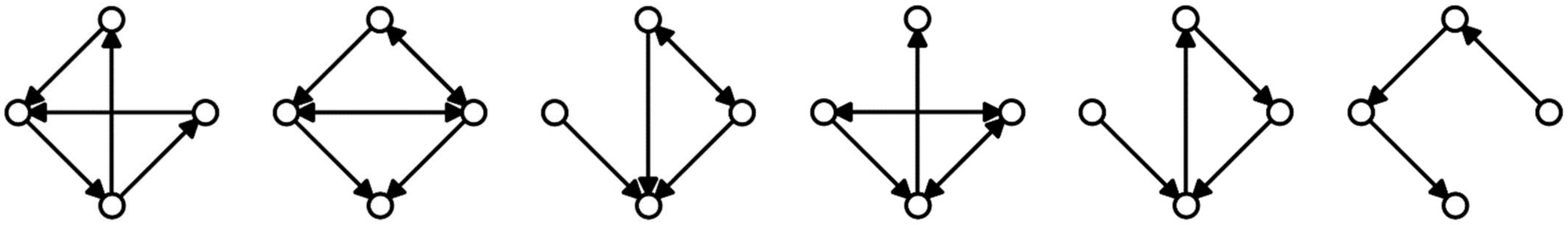 | | | | | |
| Pattern 81<br>‘11121112011111’<br>(1282, 124534) | Pattern 48<br>‘12220223011122’<br>(1274, 179823) | Pattern 126<br>‘01130122001112’<br>(1264, 715572) | Pattern 71<br>‘11220222001122’<br>(1250, 1209032) | Pattern 128<br>‘01121111001111’<br>(1243, 2136208) | Pattern 134<br>‘01110111000111’<br>(1241, 48121366) |
| 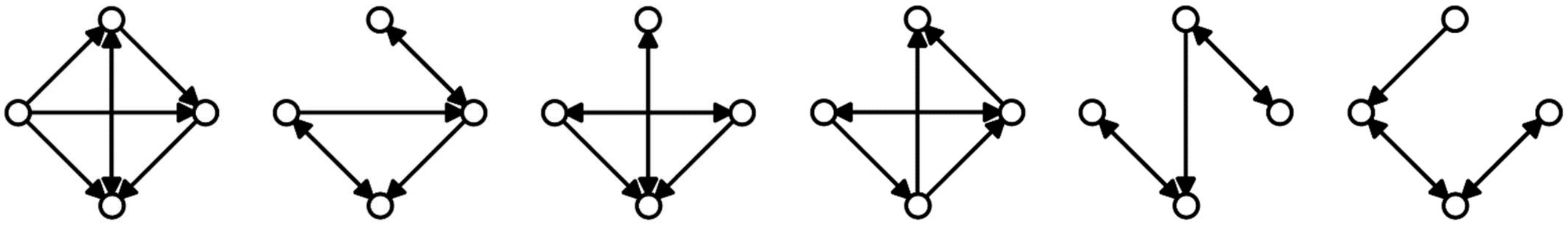 | | | | | |
| Pattern 100<br>‘02231123111112’<br>(1183, 18273) | Pattern 66<br>‘11221122001122’<br>(1173, 238042) | Pattern 76<br>‘11131122001122’<br>(1170, 750292) | Pattern 70<br>‘11220222011112’<br>(1168, 54321) | Pattern 83<br>‘11121112000122’<br>(1163, 847512) | Pattern 117<br>‘01221112000122’<br>(1140, 4965841) |

| **Motif<br>Pattern #<br>Unique Sequence<br>(Patent Count,<br>Raw Count)** | **Motif<br>Pattern #<br>Unique Sequence<br>(Patent Count,<br>Raw Count)** | **Motif<br>Pattern #<br>Unique Sequence<br>(Patent Count,<br>Raw Count)** | **Motif<br>Pattern #<br>Unique Sequence<br>(Patent Count,<br>Raw Count)** | **Motif<br>Pattern #<br>Unique Sequence<br>(Patent Count,<br>Raw Count)** | **Motif<br>Pattern #<br>Unique Sequence<br>(Patent Count,<br>Raw Count)** |
|---|---|---|---|---|---|

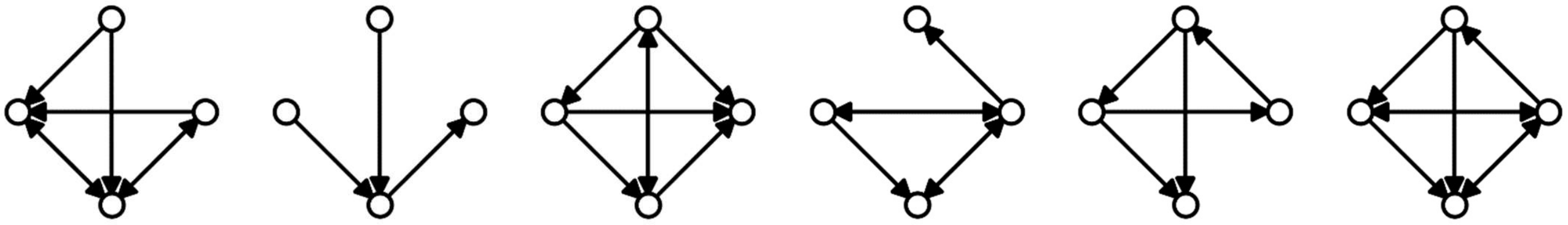

| | | | | | |
|---|---|---|---|---|---|
| Pattern 108<br>'01331222011122'<br>(1136, 169682) | Pattern 145<br>'00120111000111'<br>(1131, 438301237) | Pattern 61<br>'11230223111112'<br>(1114, 15083) | Pattern 73<br>'11220123001122'<br>(1114, 4842897) | Pattern 84<br>'11120122011111'<br>(1114, 74166) | Pattern 36<br>'12231223111122'<br>(1093, 23712) |

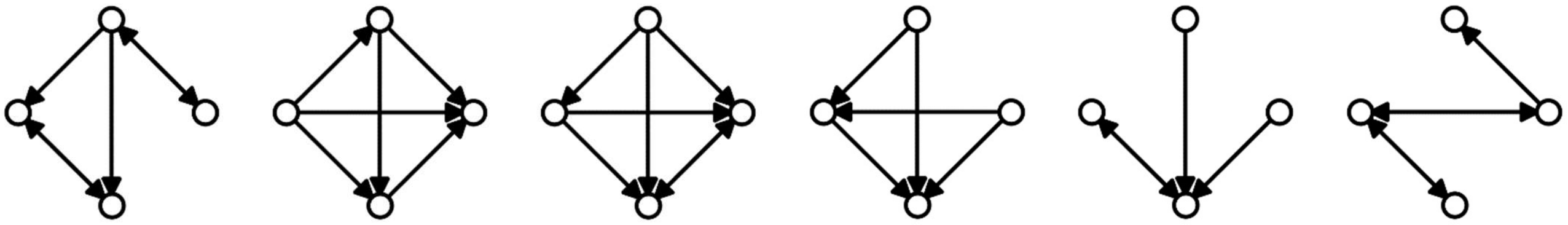

| | | | | | |
|---|---|---|---|---|---|
| Pattern 69<br>'11221113001122'<br>(1092, 722802) | Pattern 114<br>'01230123111111'<br>(1088, 138207) | Pattern 109<br>'01331123111112'<br>(1074, 21991) | Pattern 139<br>'00230122011111'<br>(1070, 335137) | Pattern 143<br>'00131111000112'<br>(1051, 100300148) | Pattern 86<br>'11120122000122'<br>(1047, 5314416) |

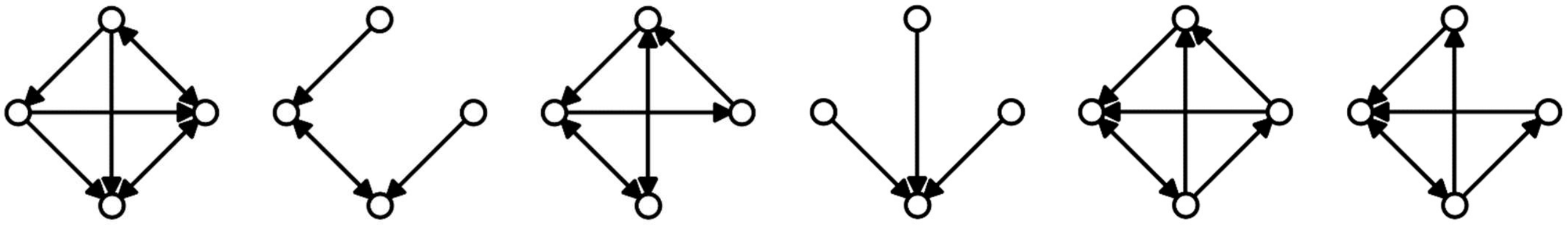

| | | | | | |
|---|---|---|---|---|---|
| Pattern 52<br>'11331223111122'<br>(1031, 12180) | Pattern 140<br>'00221111000112'<br>(1030, 15227756) | Pattern 43<br>'12221222011122'<br>(1028, 33678) | Pattern 147<br>'00030111000111'<br>(1026, 172232966) | Pattern 58<br>'11231123111112'<br>(1024, 12690) | Pattern 77<br>'11131113011112'<br>(1024, 131235) |

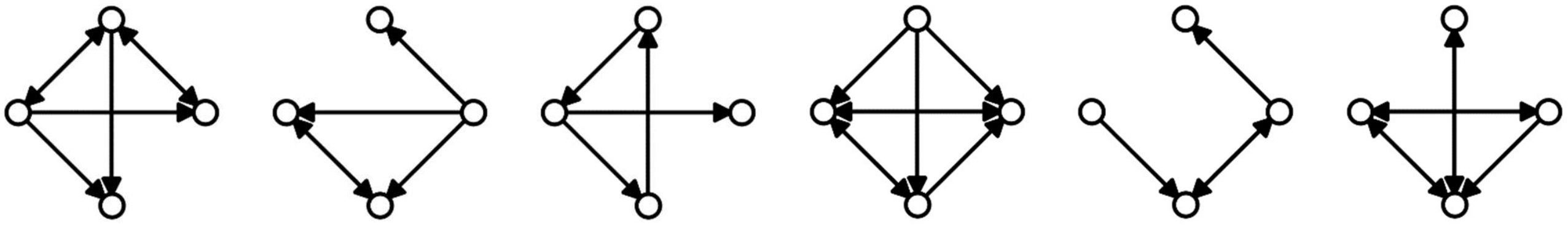

| | | | | | |
|---|---|---|---|---|---|
| Pattern 50<br>'12220133011122'<br>(1021, 134728) | Pattern 121<br>'01220113001112'<br>(1011, 668756) | Pattern 91<br>'11110112001111'<br>(1010, 1642413) | Pattern 97<br>'02331223111122'<br>(1006, 22630) | Pattern 131<br>'01120112000112'<br>(994, 241073342) | Pattern 57<br>'11231222001222'<br>(970, 171324) |

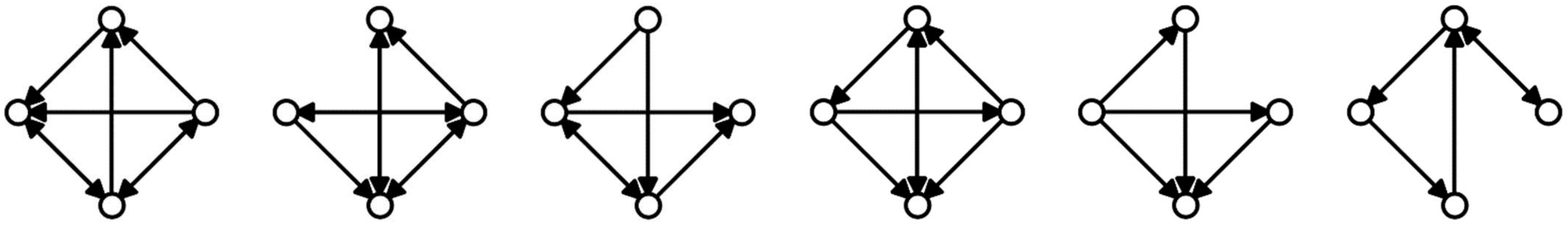

| | | | | | |
|---|---|---|---|---|---|
| Pattern 39<br>'12231133111122'<br>(967, 10656) | Pattern 37<br>'12231223011222'<br>(965, 213340) | Pattern 106<br>'02220222011112'<br>(963, 186389) | Pattern 55<br>'11231222111112'<br>(950, 6353) | Pattern 127<br>'01130113011111'<br>(949, 262028) | Pattern 82<br>'11121112001112'<br>(947, 428241) |

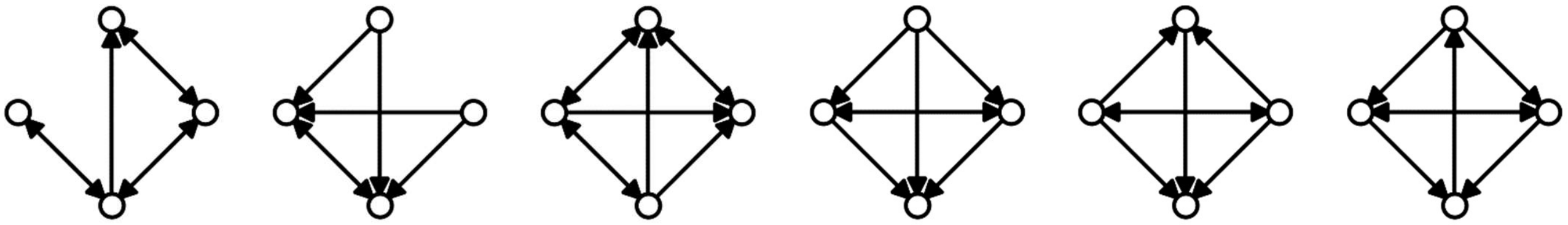

| | | | | | |
|---|---|---|---|---|---|
| Pattern 47<br>'12221123001222'<br>(935, 172463) | Pattern 137<br>'00331122011112'<br>(933, 171469) | Pattern 33<br>'12331233111222'<br>(915, 22549) | Pattern 102<br>'02230223111112'<br>(908, 16024) | Pattern 63<br>'11230133111112'<br>(905, 15150) | Pattern 42<br>'12221222111112'<br>(903, 8104) |

| Motif<br>Pattern #<br>Unique Sequence<br>(Patent Count,<br>Raw Count) | Motif<br>Pattern #<br>Unique Sequence<br>(Patent Count,<br>Raw Count) | Motif<br>Pattern #<br>Unique Sequence<br>(Patent Count,<br>Raw Count) | Motif<br>Pattern #<br>Unique Sequence<br>(Patent Count,<br>Raw Count) | Motif<br>Pattern #<br>Unique Sequence<br>(Patent Count,<br>Raw Count) | Motif<br>Pattern #<br>Unique Sequence<br>(Patent Count,<br>Raw Count) |
|---|---|---|---|---|---|
| 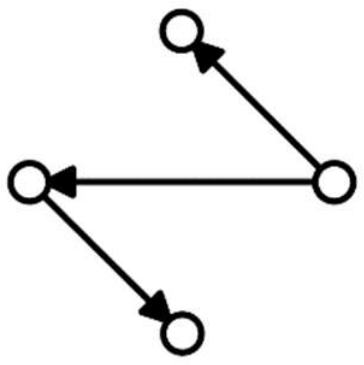<br>Pattern 135<br>‘01110012000111’<br>(902, 464168097) | 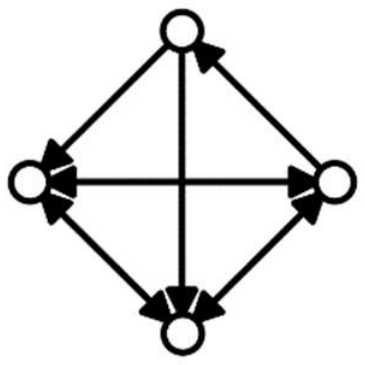<br>Pattern 31<br>‘12332223111222’<br>(881, 13486) | 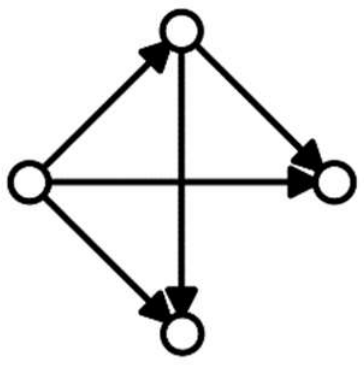<br>Pattern 122<br>‘01220023011111’<br>(869, 988868) | 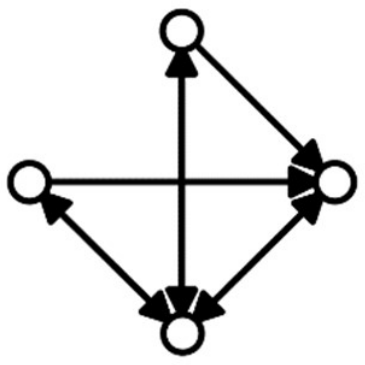<br>Pattern 53<br>‘11331223011222’<br>(864, 31812) | 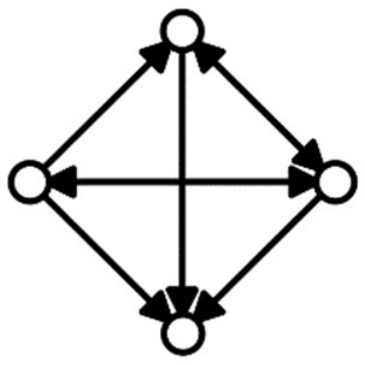<br>Pattern 41<br>‘12230233111122’<br>(862, 17168) | 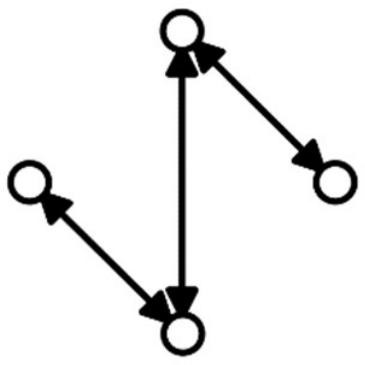<br>Pattern 67<br>‘11221122000222’<br>(848, 296972) |
| 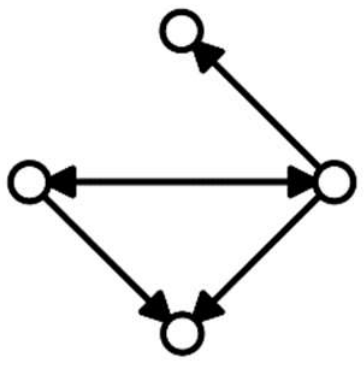<br>Pattern 89<br>‘11120023001112’<br>(847, 3931937) | 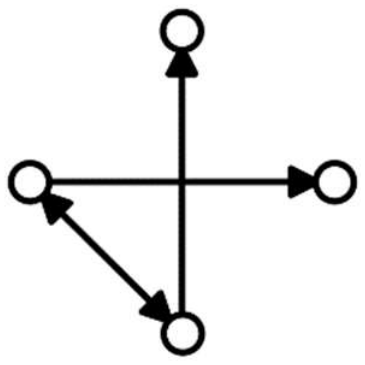<br>Pattern 93<br>‘11110022000112’<br>(843, 17800998) | 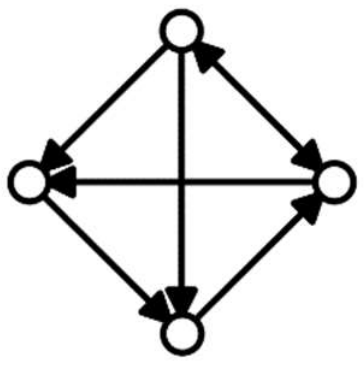<br>Pattern 45<br>‘12221123111112’<br>(835, 6106) | 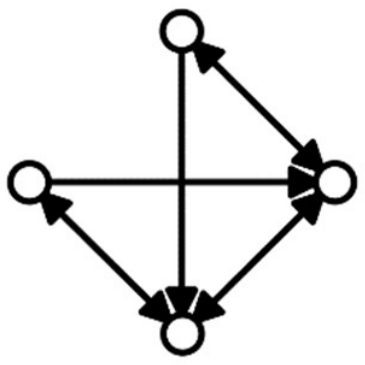<br>Pattern 51<br>‘11332222011222’<br>(831, 19661) | 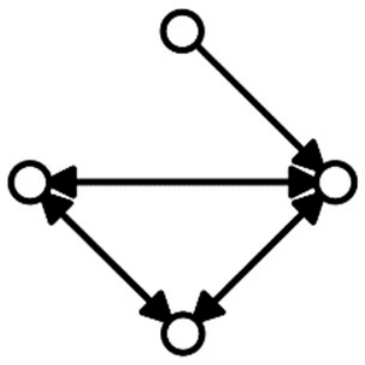<br>Pattern 99<br>‘02231222001222’<br>(823, 733747) | 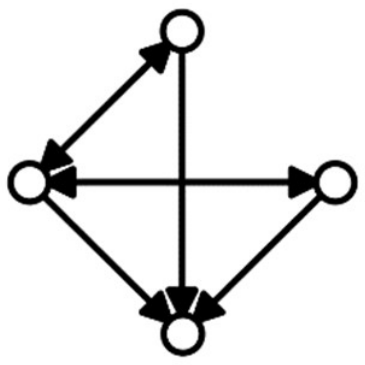<br>Pattern 62<br>‘11230223011122’<br>(821, 21828) |
| 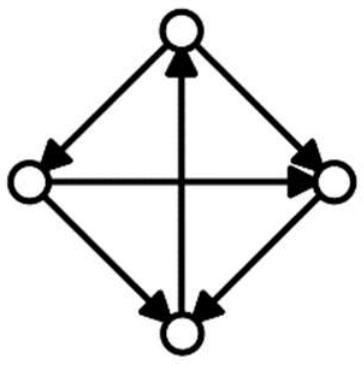<br>Pattern 64<br>‘11221122111111’<br>(816, 9119) | 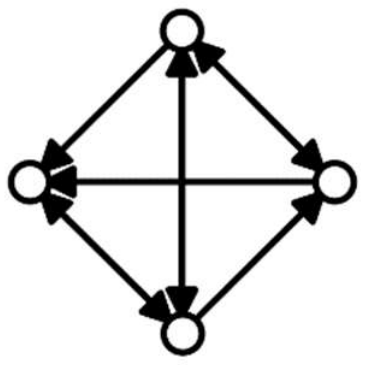<br>Pattern 21<br>‘22231233111222’<br>(810, 11790) | 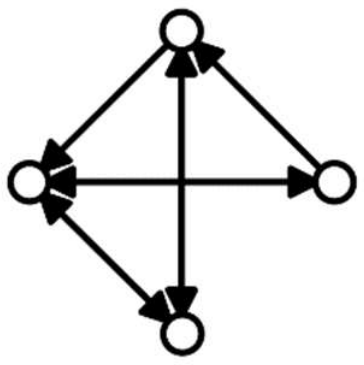<br>Pattern 35<br>‘12232222011222’<br>(810, 13144) | 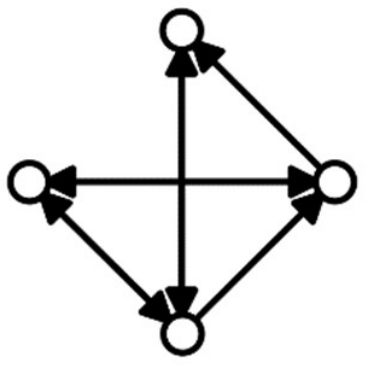<br>Pattern 27<br>‘22221223011222’<br>(808, 11931) | 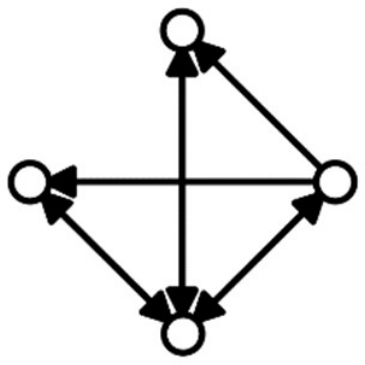<br>Pattern 40<br>‘12231133011222’<br>(801, 30853) | 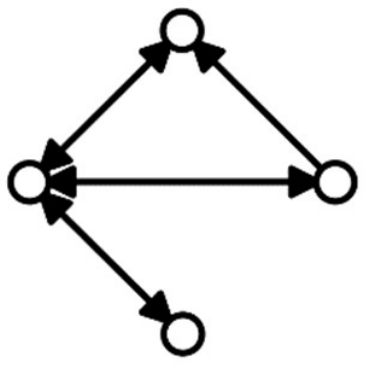<br>Pattern 60<br>‘11231123001222’<br>(793, 472485) |
| 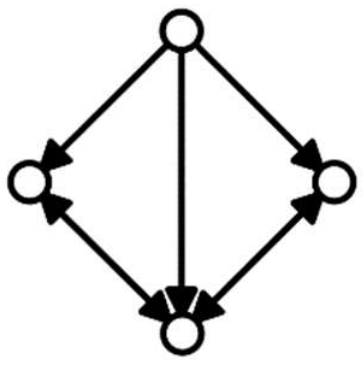<br>Pattern 101<br>‘02231123011122’<br>(778, 28560) | 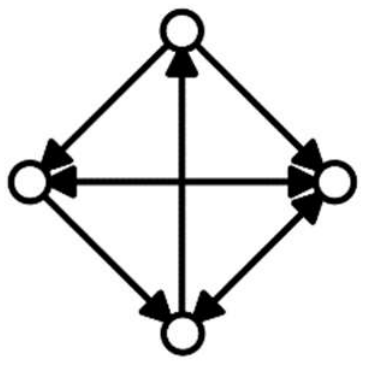<br>Pattern 34<br>‘12232222111122’<br>(777, 6327) | 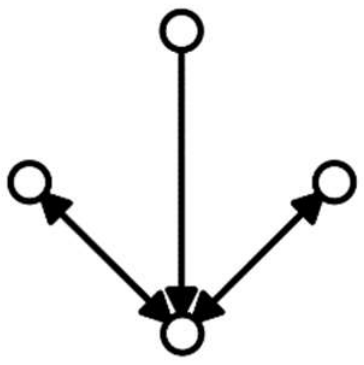<br>Pattern 124<br>‘01131112000122’<br>(777, 17965087) | 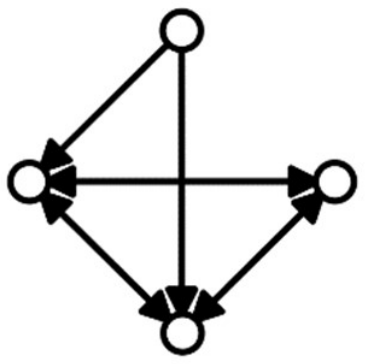<br>Pattern 96<br>‘02332222011222’<br>(723, 77036) | 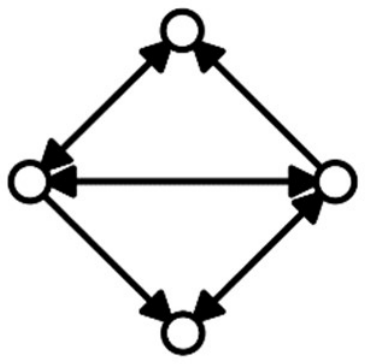<br>Pattern 28<br>‘22221133011222’<br>(713, 17103) | 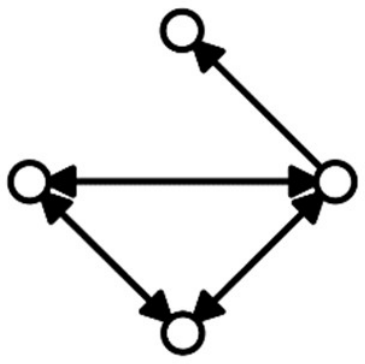<br>Pattern 49<br>‘12220223001222’<br>(708, 770373) |
| 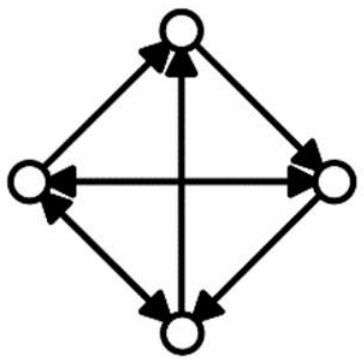<br>Pattern 26<br>‘22221223111122’<br>(706, 5387) | 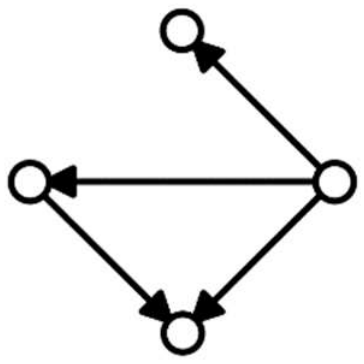<br>Pattern 133<br>‘01120013001111’<br>(705, 18986484) | 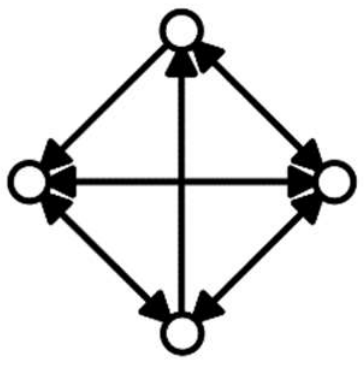<br>Pattern 16<br>‘22332233112222’<br>(704, 12245) | 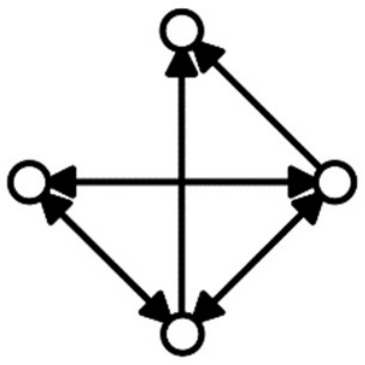<br>Pattern 29<br>‘22220233011222’<br>(693, 65357) | 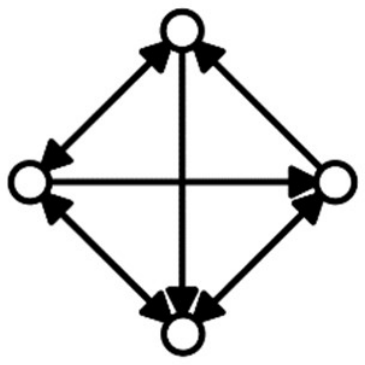<br>Pattern 19<br>‘22232223111222’<br>(664, 6826) | 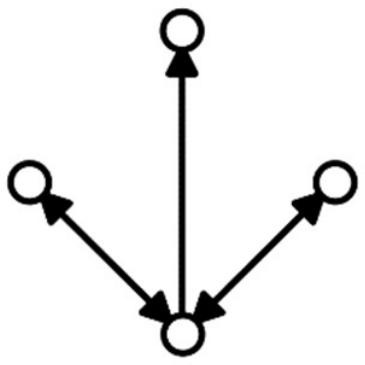<br>Pattern 88<br>‘11120113000122’<br>(661, 18630939) |
| 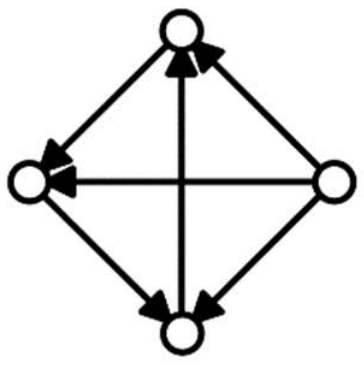<br>Pattern 105<br>‘02221113111111’<br>(658, 5200) | 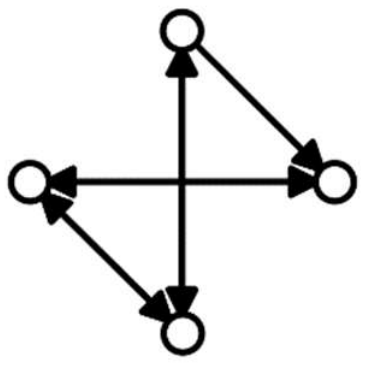<br>Pattern 44<br>‘12221222001222’<br>(655, 24423) | 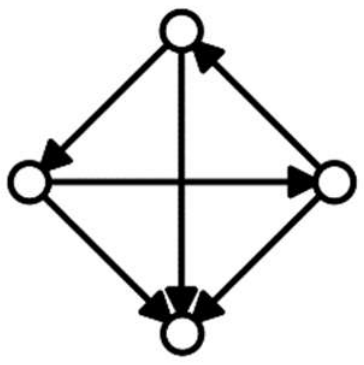<br>Pattern 79<br>‘11130222111111’<br>(643, 5211) | 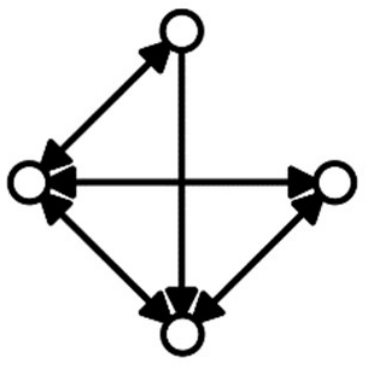<br>Pattern 32<br>‘12332223012222’<br>(635, 41198) | 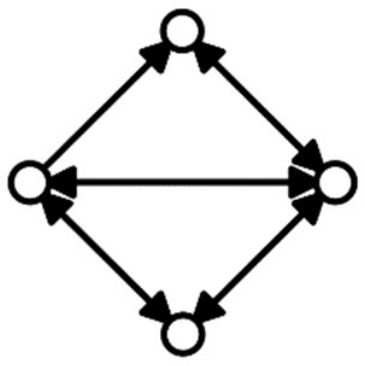<br>Pattern 22<br>‘22231233012222’<br>(600, 38433) | 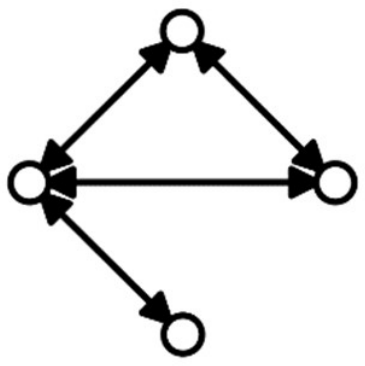<br>Pattern 38<br>‘12231223002222’<br>(600, 116119) |

| **Motif**<br>**Pattern #**<br>**Unique Sequence**<br>**(Patent Count, Raw Count)** | **Motif**<br>**Pattern #**<br>**Unique Sequence**<br>**(Patent Count, Raw Count)** | **Motif**<br>**Pattern #**<br>**Unique Sequence**<br>**(Patent Count, Raw Count)** | **Motif**<br>**Pattern #**<br>**Unique Sequence**<br>**(Patent Count, Raw Count)** | **Motif**<br>**Pattern #**<br>**Unique Sequence**<br>**(Patent Count, Raw Count)** | **Motif**<br>**Pattern #**<br>**Unique Sequence**<br>**(Patent Count, Raw Count)** |
|---|---|---|---|---|---|
| 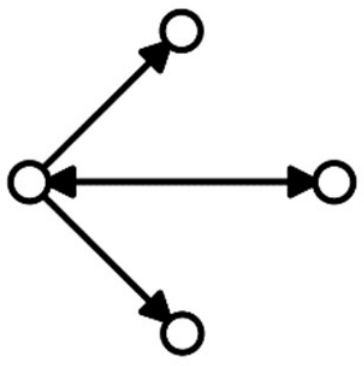<br>Pattern 94<br>'11110013000112'<br>(587, 104257858) | 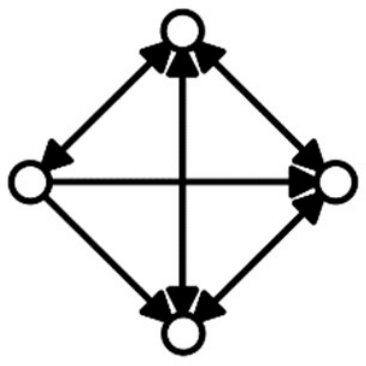<br>Pattern 30<br>'13332233112222'<br>(554, 6693) | 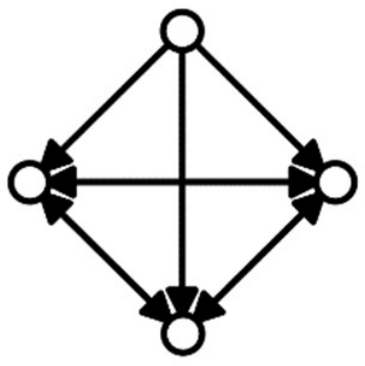<br>Pattern 95<br>'03332223111222'<br>(527, 6340) | 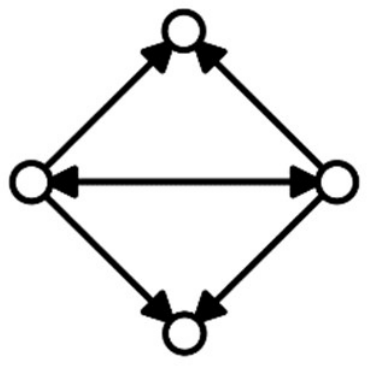<br>Pattern 74<br>'11220033011112'<br>(520, 117749) | 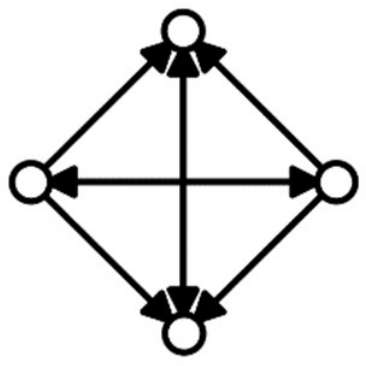<br>Pattern 54<br>'11331133111122'<br>(503, 2919) | 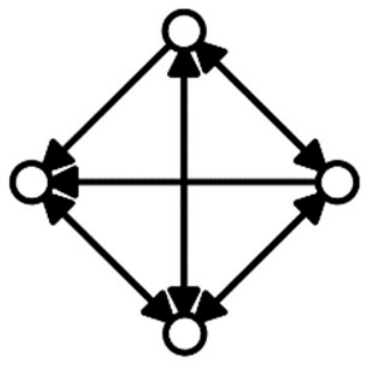<br>Pattern 18<br>'22331333112222'<br>(497, 5976) |
| 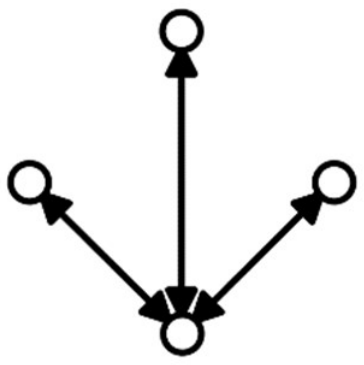<br>Pattern 78<br>'11131113000222'<br>(495, 1352655) | 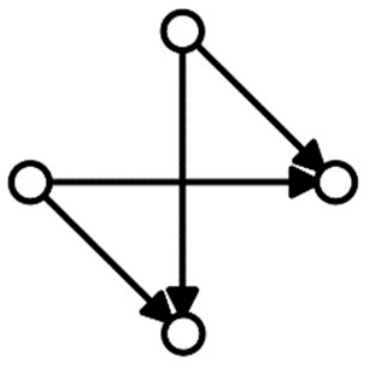<br>Pattern 142<br>'00220022001111'<br>(476, 894087) | 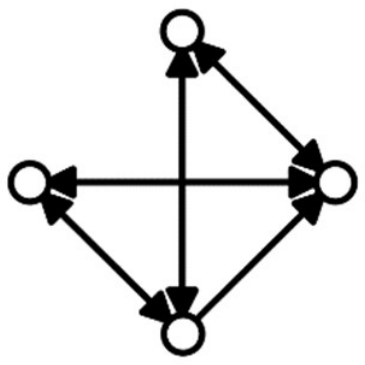<br>Pattern 20<br>'22232223012222'<br>(434, 5503) | 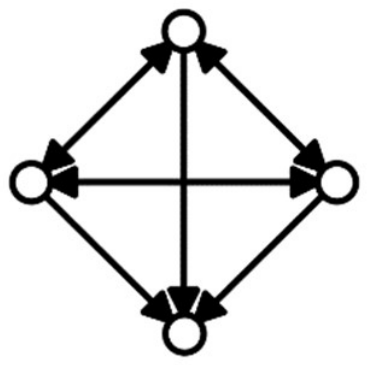<br>Pattern 23<br>'22230333111222'<br>(423, 4363) | 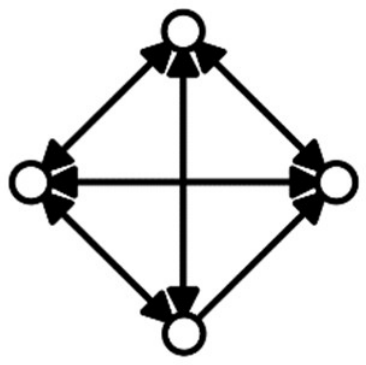<br>Pattern 15<br>'23332333122222'<br>(406, 7016) | 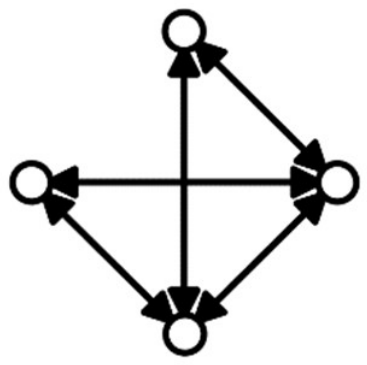<br>Pattern 17<br>'22332233022222'<br>(390, 10204) |
| 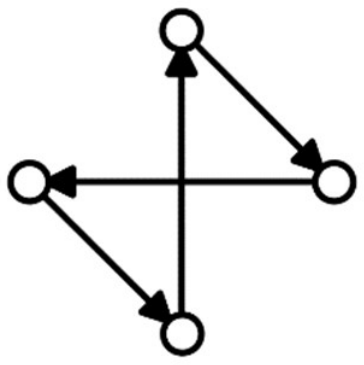<br>Pattern 90<br>'11111111001111'<br>(358, 22374) | 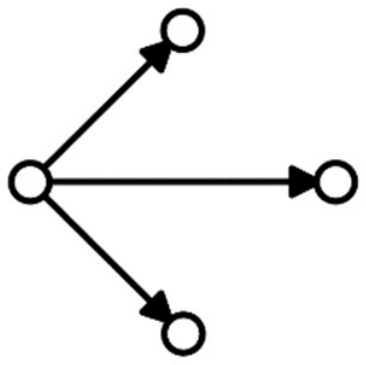<br>Pattern 136<br>'01110003000111'<br>(331, 654838961) | 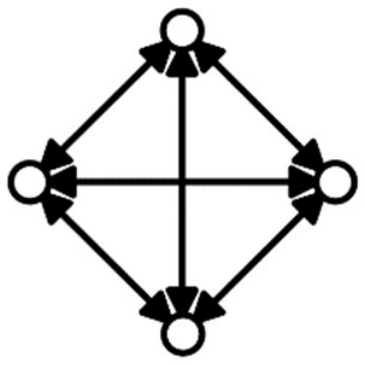<br>Pattern 14<br>'33333333222222'<br>(160, 1233) | 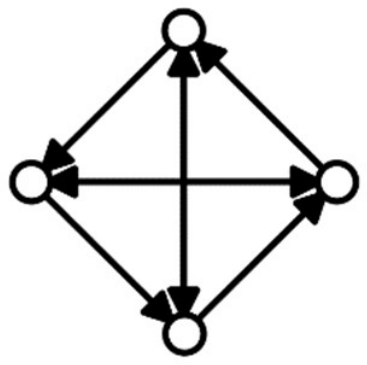<br>Pattern 24<br>'22222222111122'<br>(155, 596) | 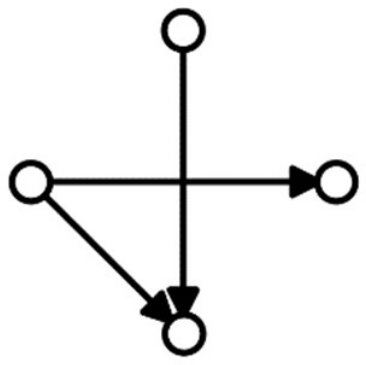<br>Pattern 146<br>'00120012000111'<br>(115, 8748376) | 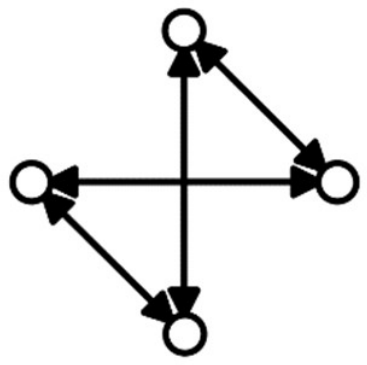<br>Pattern 25<br>'22222222002222'<br>(93, 3588) |

**Table S-6**: Top 3 subgraphs under most common motifs.

**Pattern 13**

Raw Count - 13,708,692, Number of Unique Graphs - 2,094,018

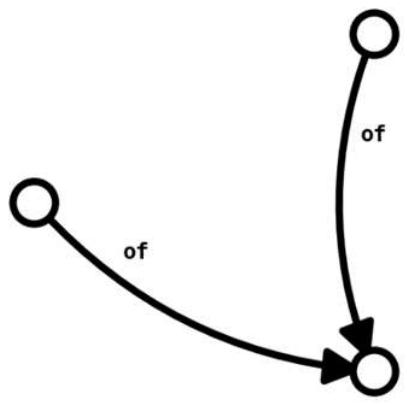

1,234,800

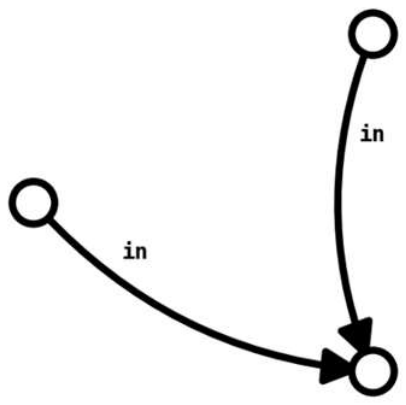

294,968

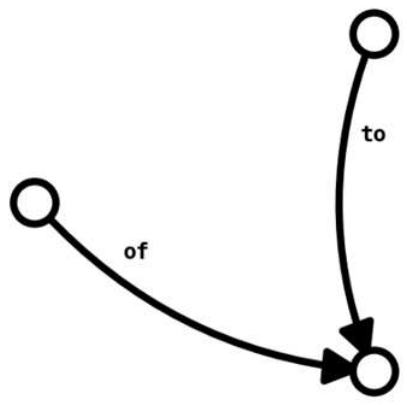

104,441

**Pattern 11**

Raw Count - 19,213,782, Number of Unique Graphs - 2,911,124

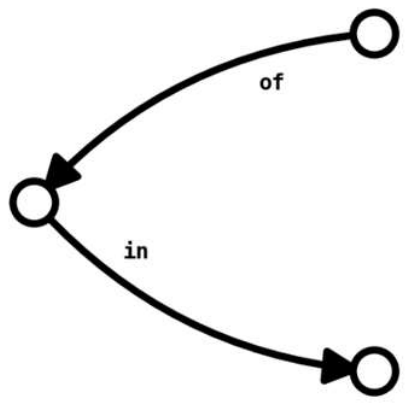

210,953

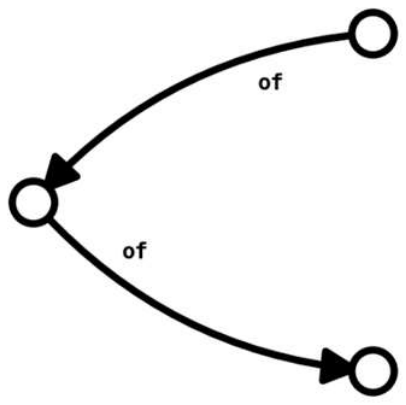

156,727

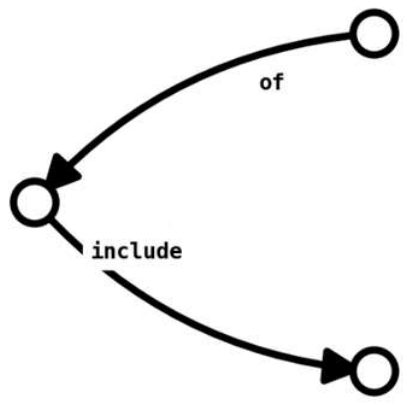

135,186

**Pattern 8**

Raw Count - 86,509, Number of Unique Graphs - 62,735

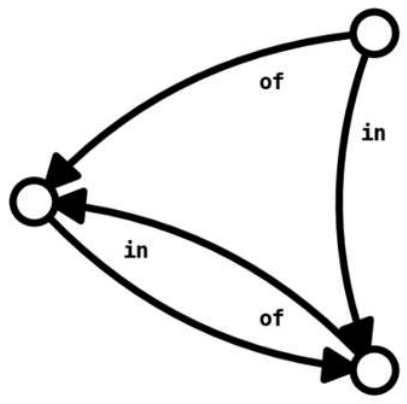

219

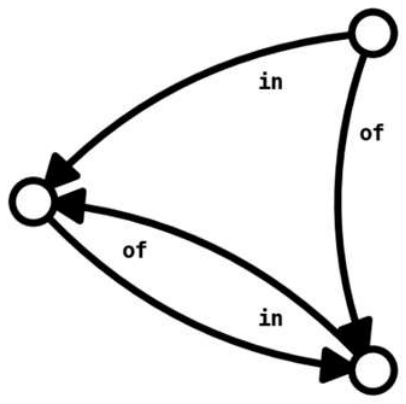

195

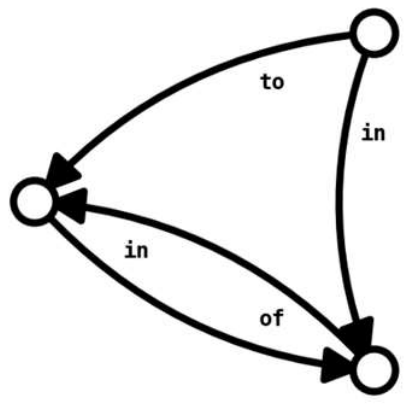

127

**Pattern 9**

Raw Count - 3,515,457, Number of Unique Graphs - 1,159,927

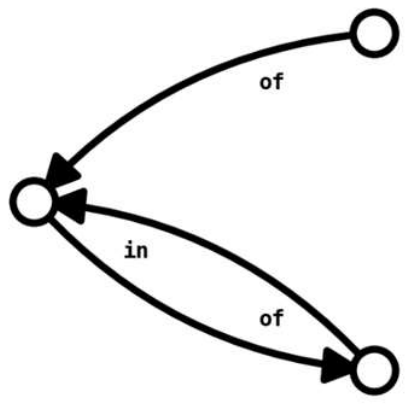

9,287

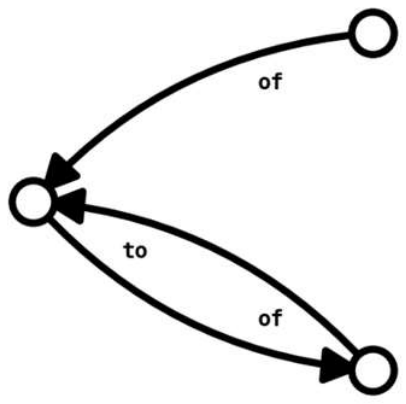

8,156

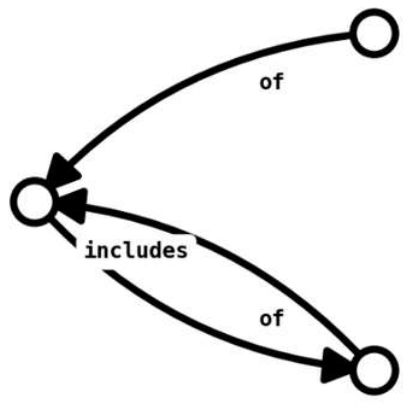

5,739

**Pattern 122**

Raw Count - 988,868, Number of Unique Graphs - 88,630

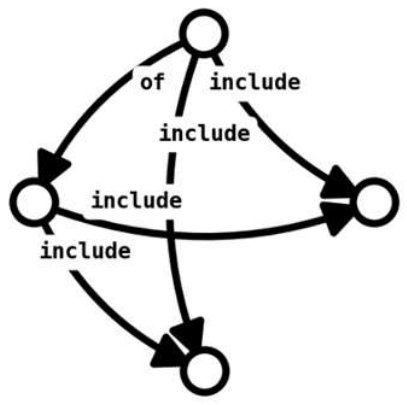

252,156

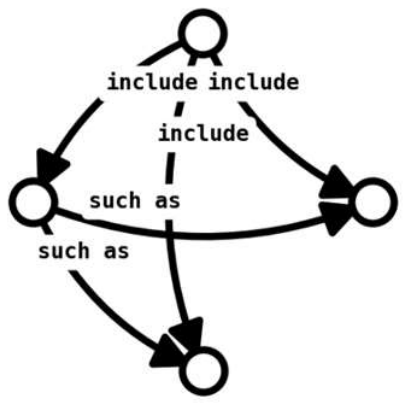

43,785

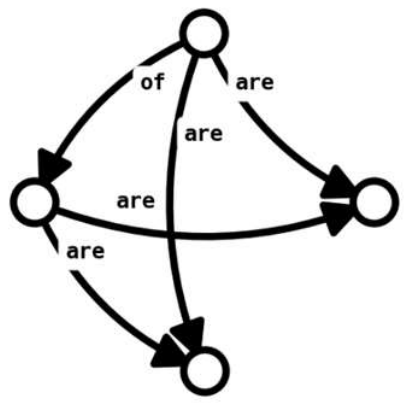

24,705

**Pattern 125**

Raw Count - 345,797, Number of Unique Graphs - 161,910

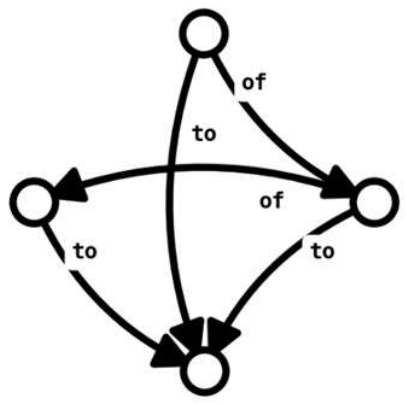

483

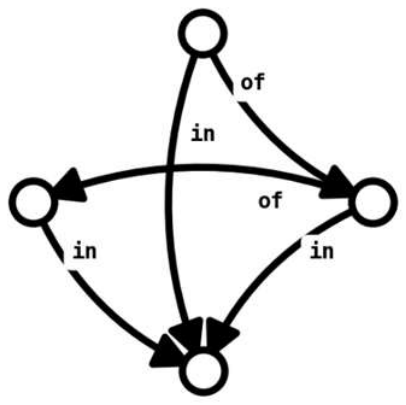

454

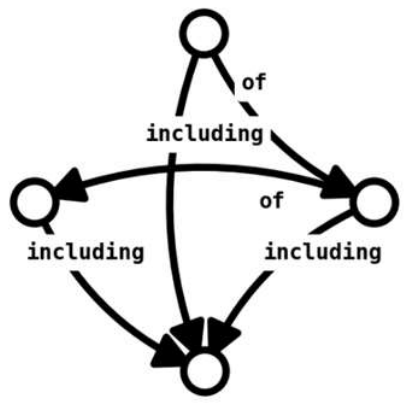

361

**Pattern 130**

Raw Count - 772,630, Number of Unique Graphs - 337,482

| | | |
|---|---|---|
| 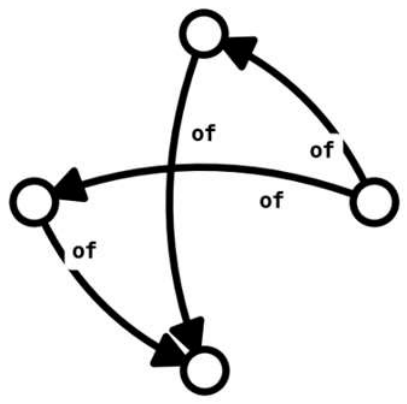 | 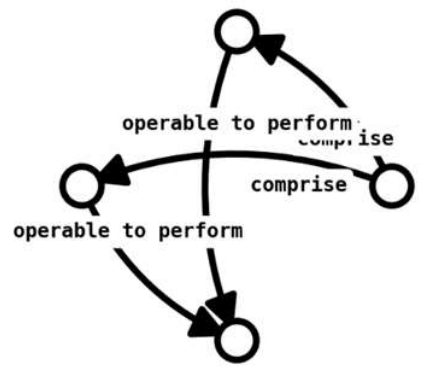 | 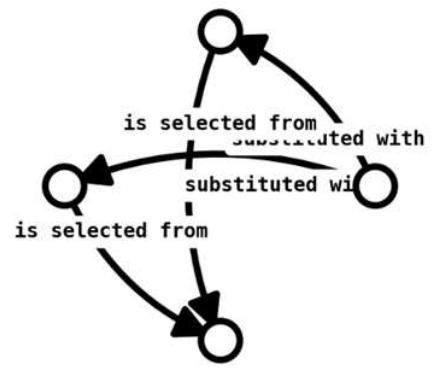 |
| 4,664 | 3,063 | 2,966 |

**Pattern 141**

Raw Count - 10,909,900, Number of Unique Graphs - 2,698,995

| | | |
|---|---|---|
| 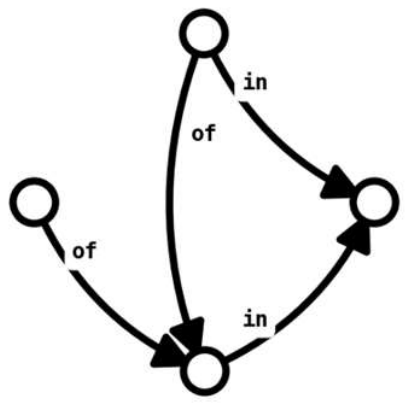 | 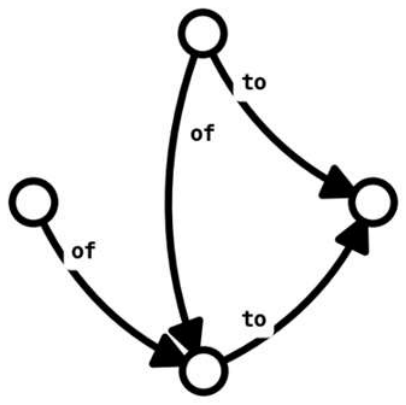 | 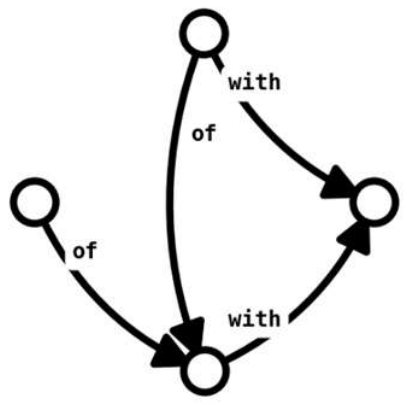 |
| 87,202 | 42,993 | 34,279 |

**Table S-7**: Top 3 subgraphs under selected patterns at different percentiles.

| Pattern 13<br>Zipf Dist. Parameter (s) = 1.016; Zipf Dist. Error (e) = 3.770e-05 | | | |
|---|---|---|---|
| (0.05, 0.2] –<br>8 Unique Graphs | 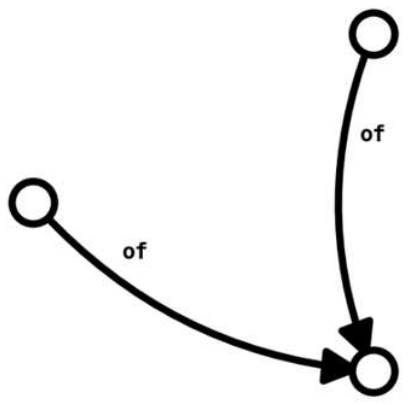 | 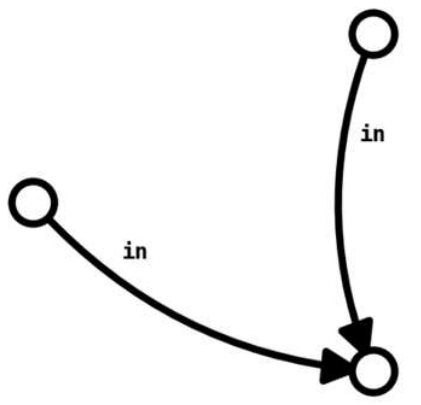 | 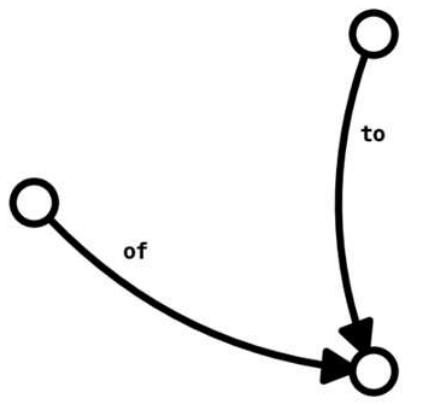 |
| (0.2, 0.5] –<br>682 Unique Graphs | 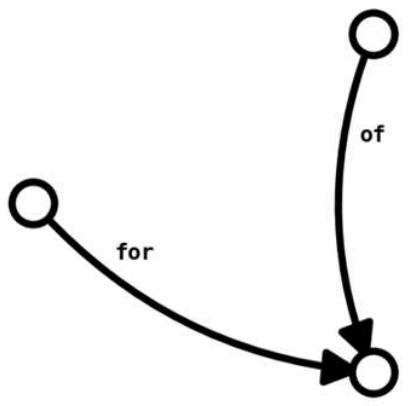 | 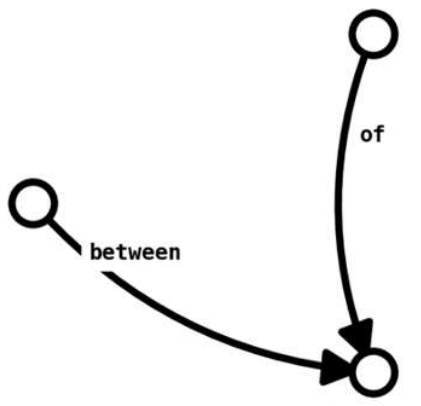 | 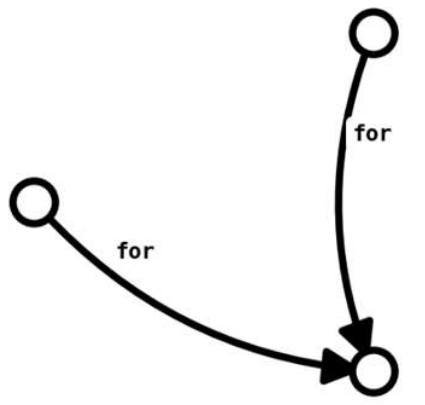 |
| (0.5, 0.8] - 74,305<br>Unique Graphs | 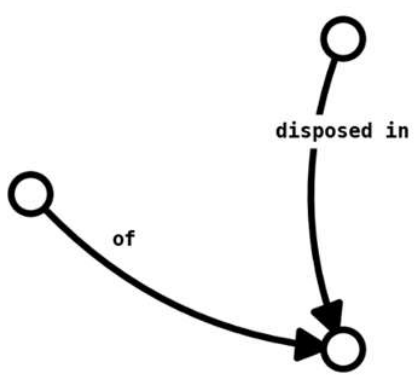 | 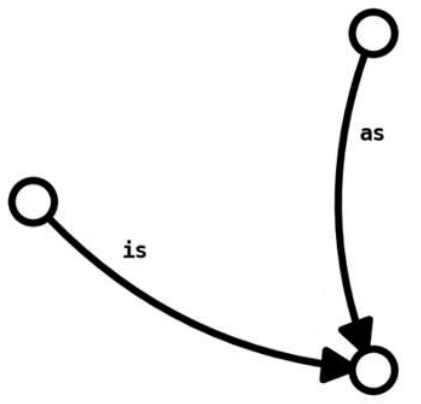 | 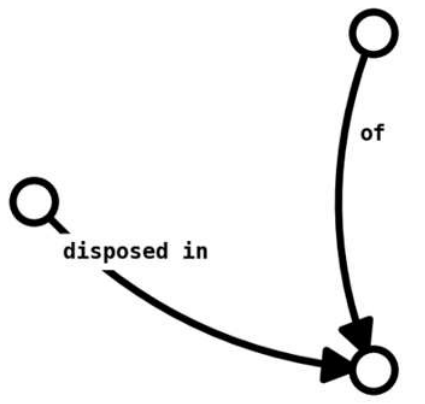 |
| (0.8, 0.1] - 2,019,023<br>Unique Graphs | 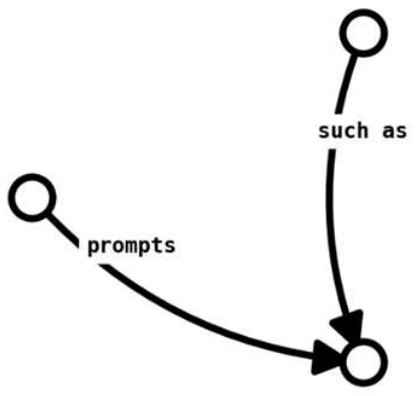 | 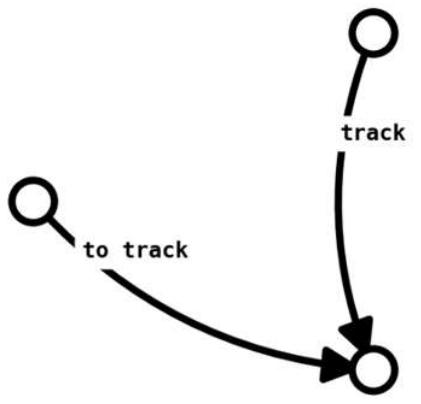 | 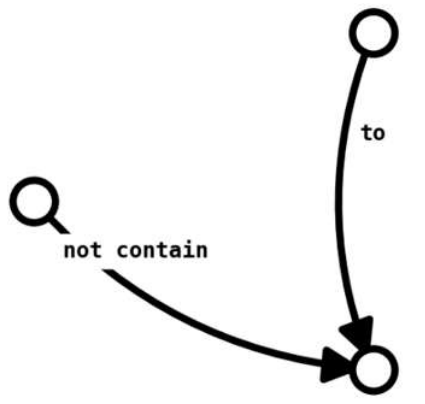 |

| Pattern 11<br>Zipf Dist. Parameter (s) = 8.334e-01; Zipf Dist. Error (e) = 2.401e-05 | | | |
|---|---|---|---|
| (0.05, 0.2] –<br>902 Unique Graphs | 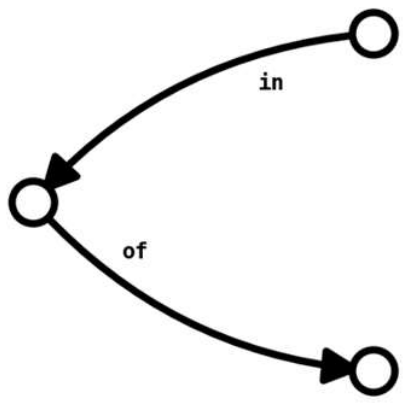 | 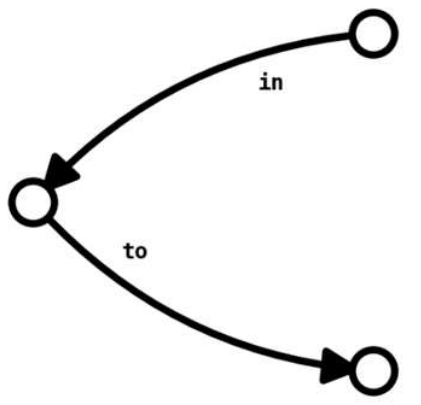 | 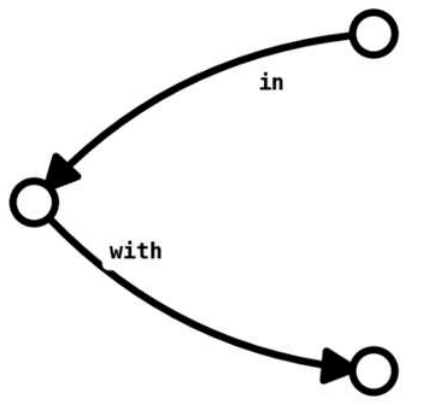 |
| (0.2, 0.5] –<br>69,563 Unique Graphs | 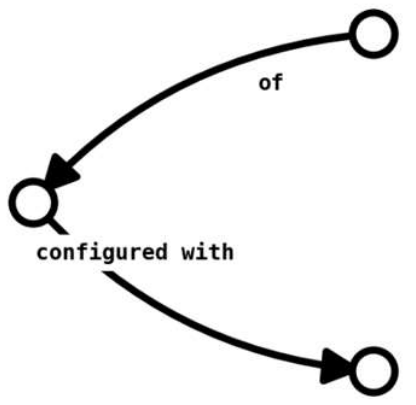 | 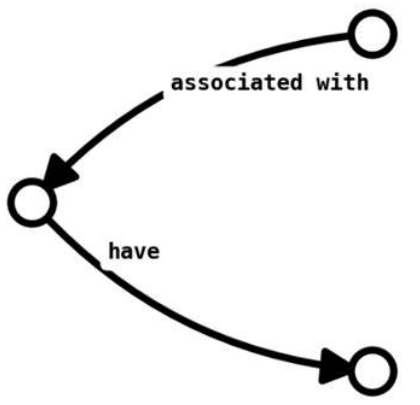 | 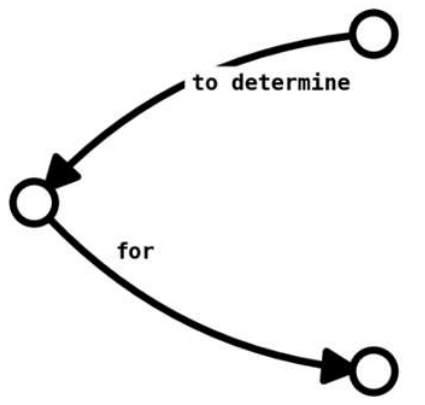 |
| (0.5, 0.8] –<br>783,475 Unique Graphs | 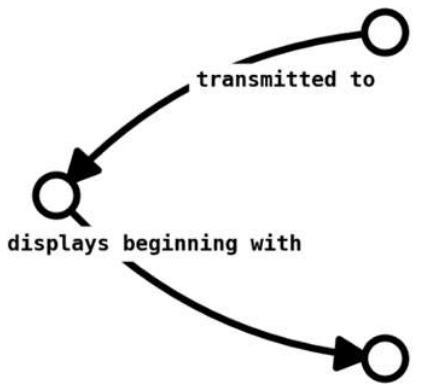 | 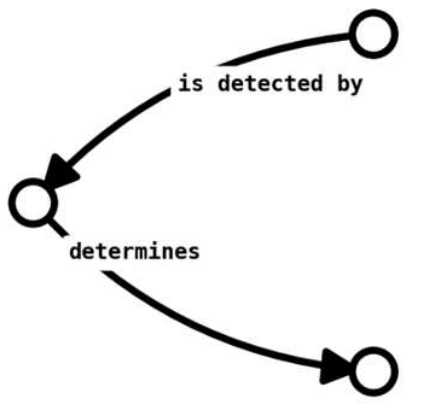 | 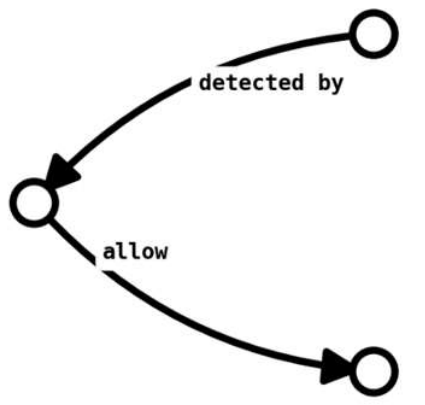 |
| (0.8, 0.1] –<br>2,057,174 Unique<br>Graphs | 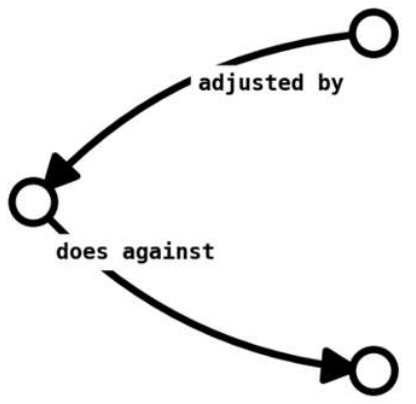 | 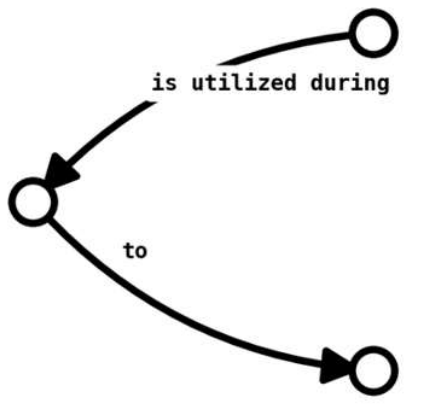 | 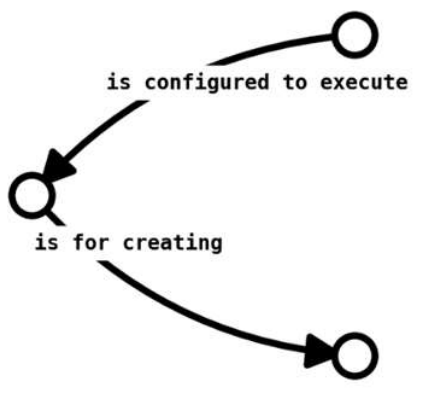 |

| Pattern 122<br>**Zipf Dist. Parameter (s) = 1.249; Zipf Dist. Error (e) = 3.932e-04** | | | |
|---|---|---|---|
| (0.05, 0.5] – 6 Unique Graphs | 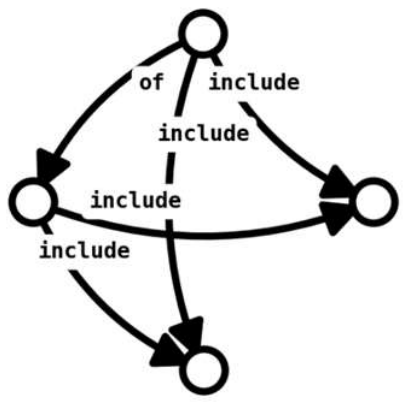  | 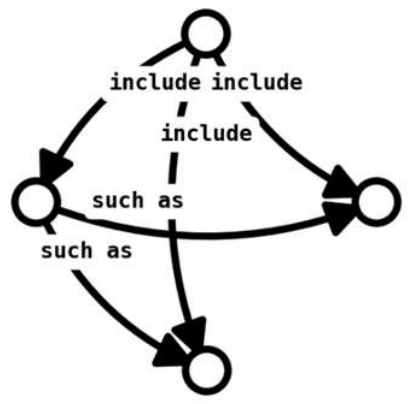  | 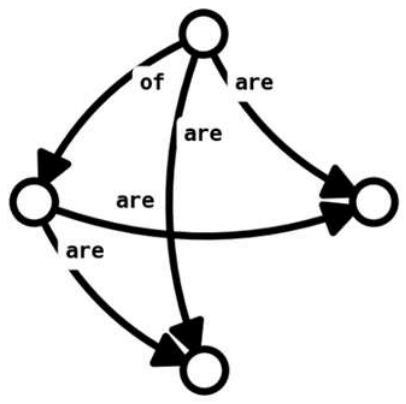  |
| (0.5, 0.8] – 168 Unique Graphs | 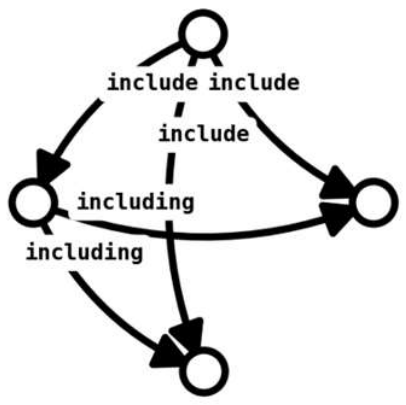  | 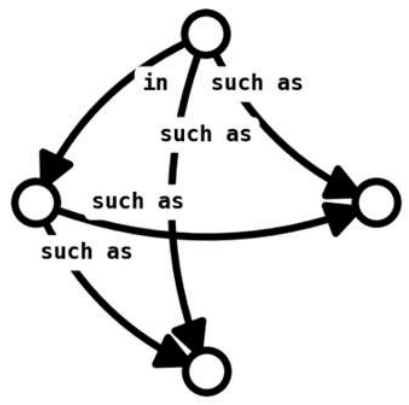  | 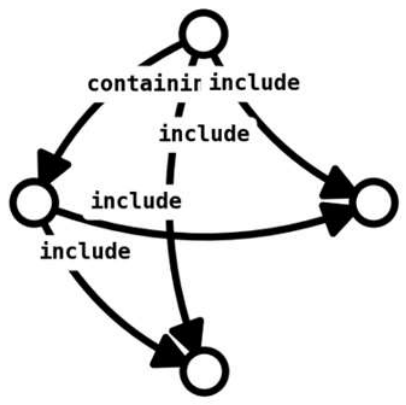  |
| (0.8, 0.1] - 88,454 Unique Graphs | 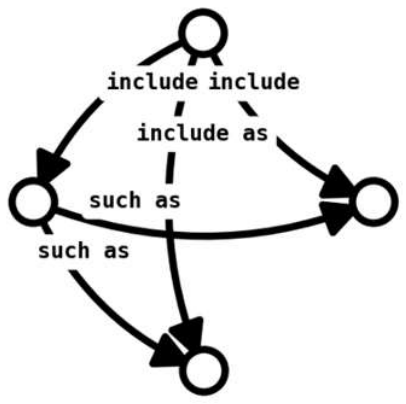  | 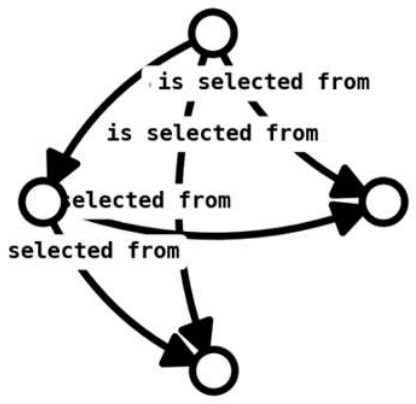  | 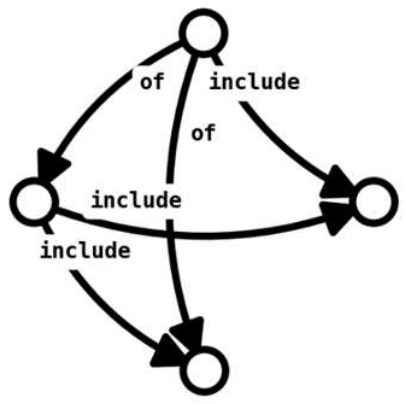  |

**Table S-8**: Examples of concretisation.

**Entities**
(Pattern 11)

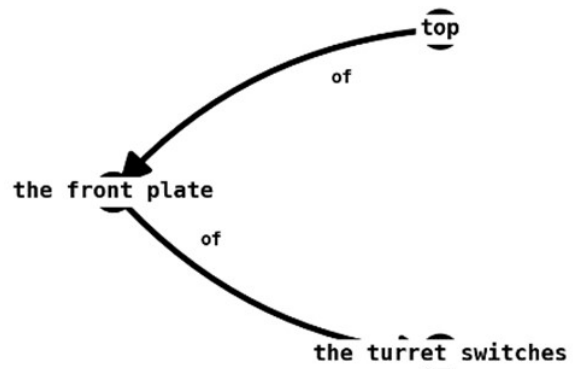

→

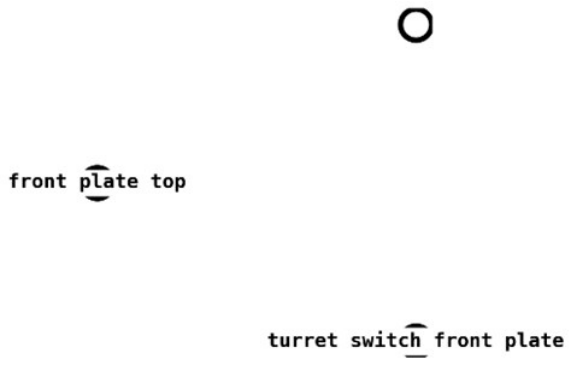

Based on “Lamp-support” - https://patents.google.com/patent/US1084362/

---

**Entities**
(Pattern 13)

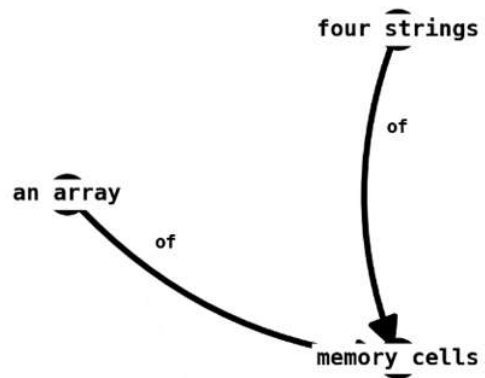

→

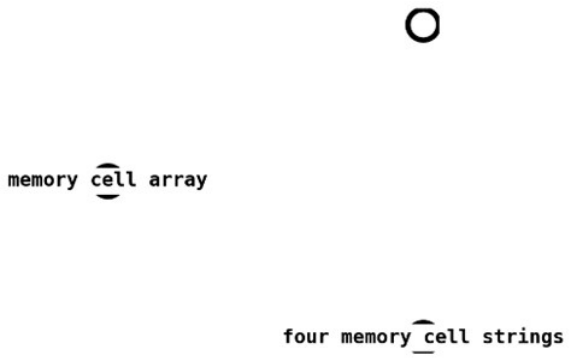

Based on “Memory configured to perform logic operations on values representative of sensed characteristics of data lines and a threshold data value” - https://patents.google.com/patent/US11074982/

---

**Entities**
(Pattern 122)

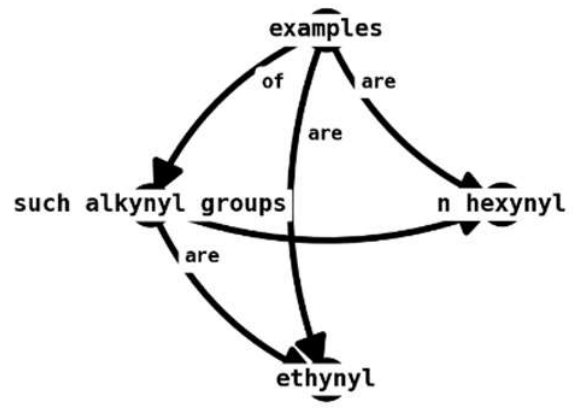

→

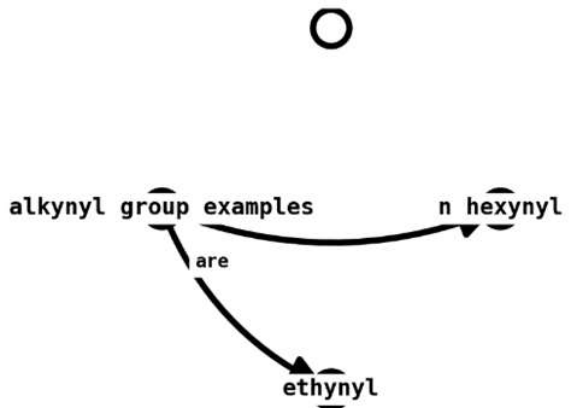

Based on “Polar-substituted hydrocarbons” - https://patents.google.com/patent/US6071895/

---

**Entities**
(Pattern 125)

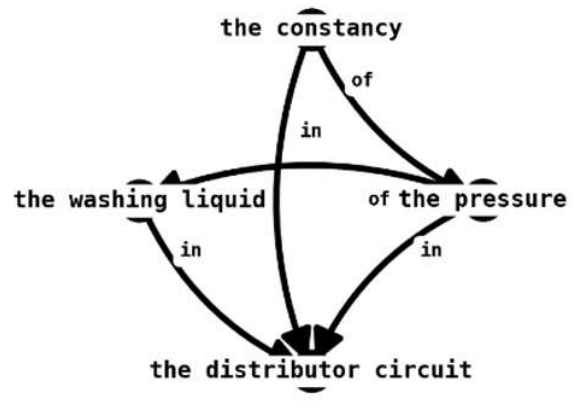

→

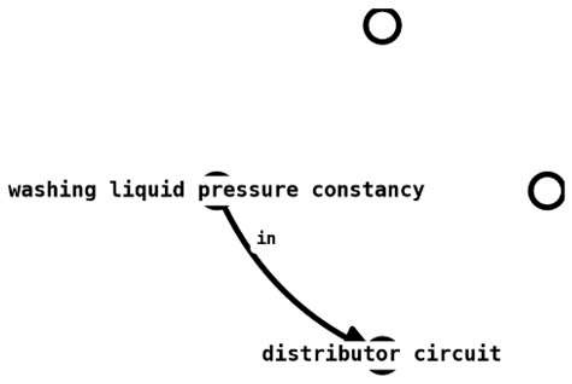

Based on “Machine for washing objects and method for the hydraulic and mechanical connection of a trolley carrying objects to be washed to a feed circuit of a washing liquid for a machine for washing objects” - https://patents.google.com/patent/US9393600/

---

**Entities**
(Pattern 141)

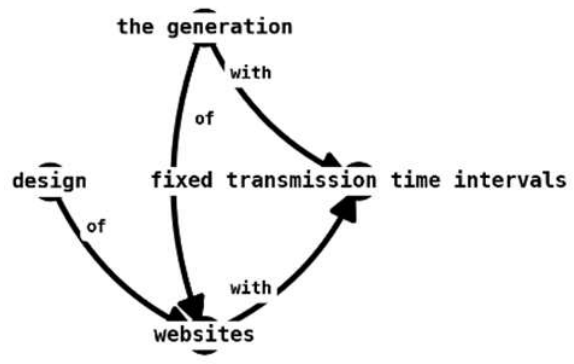

→

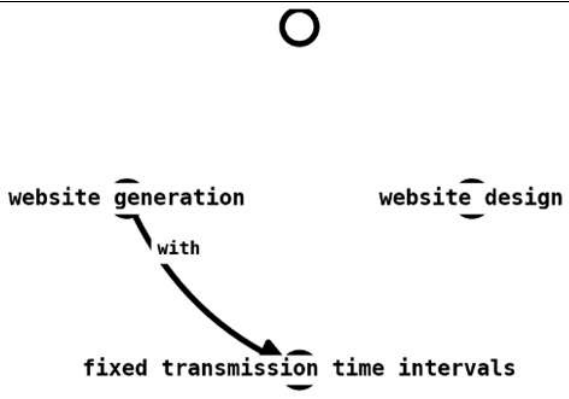

Based on “Systems and methods for loading websites with multiple items” - https://patents.google.com/patent/US11055378/

---

**Relationships**
(Pattern 8)

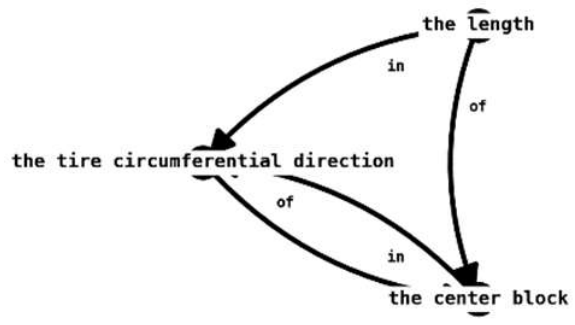

→

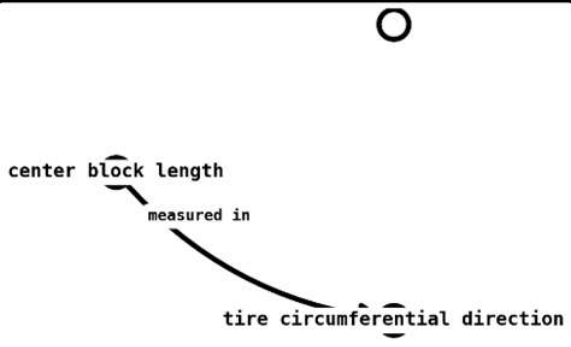

Based on “Pneumatic tire for heavy loads” - https://patents.google.com/patent/US10308079/

---

**Hierarchy**
(Pattern 122)

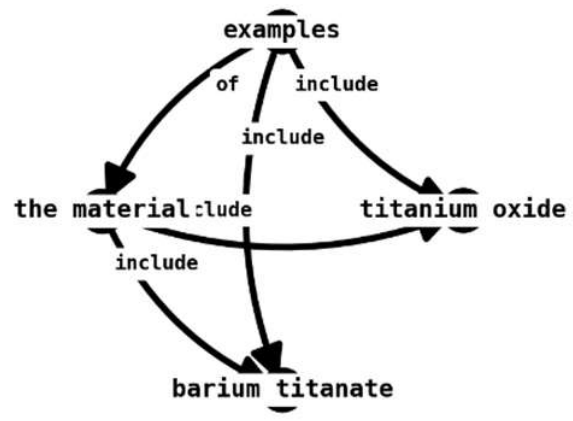

→

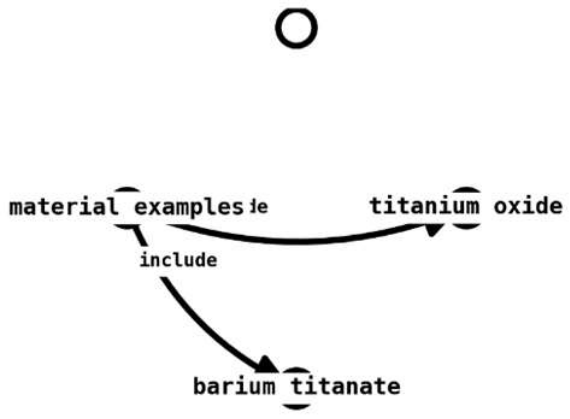

Based on “Dispersion liquid, composition, film, manufacturing method of film, and dispersant” - https://patents.google.com/patent/US10928726/

**Table S-9**: Introductory knowledge bases of coffee grinder and glue gun.

| | **Coffee Grinder** | **Glue Gun** |
|---|---|---|
| Number of Patents | 7 | 7 |
| Number of Sentences | 1,084 | 888 |
| Number of Sentences with Facts | 847 | 675 |
| Number of Facts | 3,618 | 2,660 |
| Number of Unique Entities | 1,616 | 1,252 |
| Number of Unique Relationships | 991 | 707 |
| Number of Facts with Freq >= 2 | 320 | 506 |

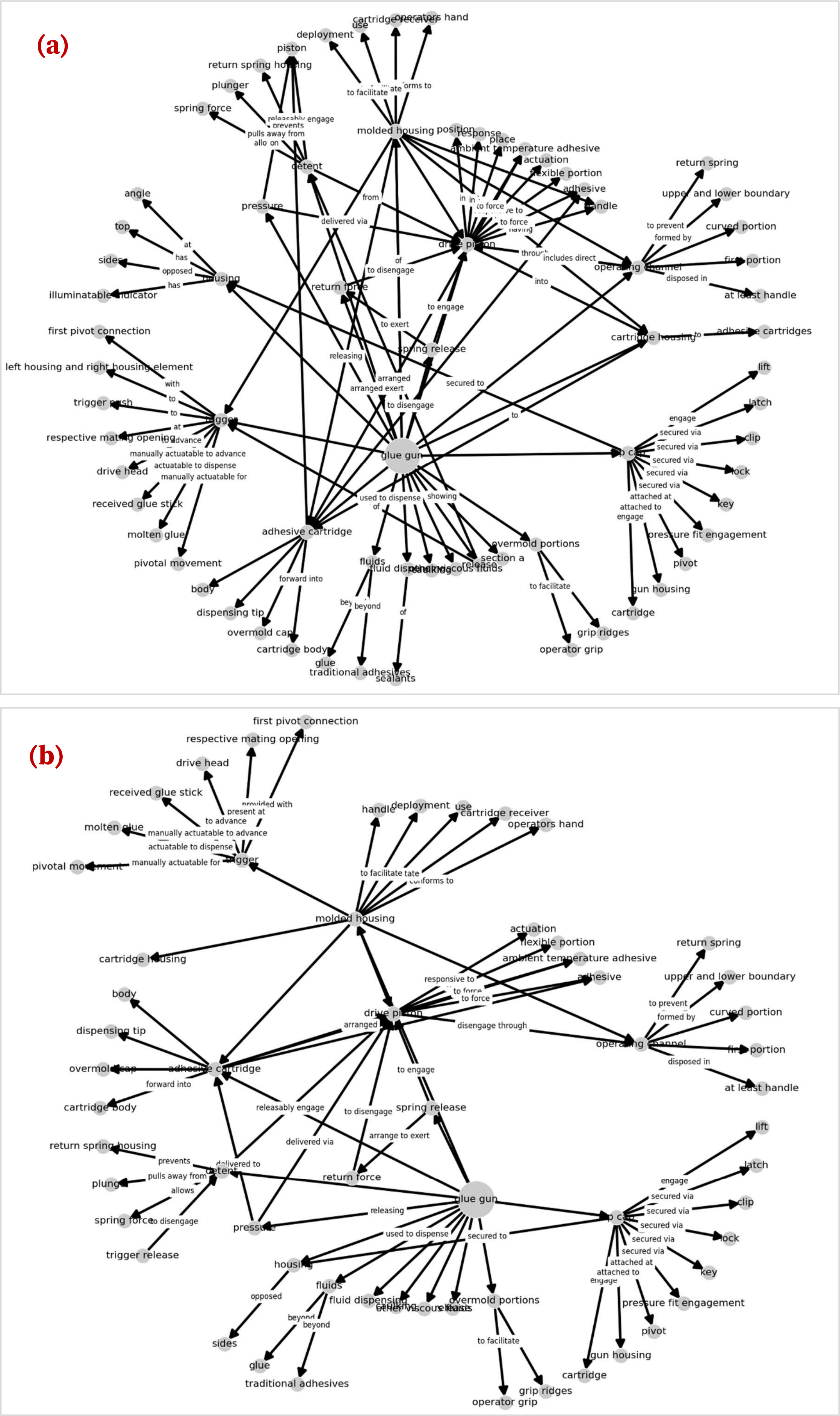

**Figure S-5**: a) extended neighbourhood of “glue gun” entity; b) transformed neighbourhood thereof. The edges without a label indicate a hierarchical relationship – “include”.